\documentclass{article}
\usepackage{arxiv}

\usepackage{amsmath,amsfonts,bm}

\def\eqref#1{equation~\ref{#1}}

\def\1{\bm{1}}

\DeclareMathAlphabet{\mathsfit}{\encodingdefault}{\sfdefault}{m}{sl}
\SetMathAlphabet{\mathsfit}{bold}{\encodingdefault}{\sfdefault}{bx}{n}

\def\gA{{\mathcal{A}}}

\def\gG{{\mathcal{G}}}
\def\gH{{\mathcal{H}}}

\def\gN{{\mathcal{N}}}

\def\gS{{\mathcal{S}}}

\def\gX{{\mathcal{X}}}
\def\gY{{\mathcal{Y}}}
\def\gZ{{\mathcal{Z}}}

\newcommand{\E}{\mathbb{E}}

\newcommand{\R}{\mathbb{R}}

\DeclareMathOperator*{\argmin}{arg\,min}

\usepackage{authblk}
\usepackage{booktabs}       
\usepackage{nicefrac}       
\usepackage{microtype}      
\usepackage{graphicx}
\usepackage{subcaption}
\usepackage{wrapfig}
\usepackage{enumitem}
\usepackage{amsthm}
\usepackage{amssymb}
\usepackage{algorithm,algorithmic}
\usepackage{placeins}
\usepackage{comment}

\usepackage{multirow}
\usepackage{mathtools}
\usepackage{dsfont}
\usepackage{adjustbox}
\usepackage{array}
\usepackage{titletoc}
\usepackage{tikz}
\usetikzlibrary{arrows.meta,positioning}
\usepackage[colorlinks,citecolor=dgreen,urlcolor=dblue,linkcolor=dblue]{hyperref}

\newcommand{\HS}{\mathrm{HS}}
\newcommand{\norm}[1]{\left\lVert #1\right\rVert}

\allowdisplaybreaks
\newtheorem{theorem}{Theorem}[section]
\newtheorem{lemma}[theorem]{Lemma}
\newtheorem{definition}[theorem]{Definition}
\newtheorem{proposition}[theorem]{Proposition}
\newtheorem{corollary}[theorem]{Corollary}
\newtheorem{remark}[theorem]{Remark}

\newtheorem{assumption}[theorem]{Assumption}

\AtBeginDocument{
  \setlength{\abovedisplayskip}{6pt plus 2pt minus 2pt}
  \setlength{\belowdisplayskip}{6pt plus 2pt minus 2pt}
  \setlength{\abovedisplayshortskip}{3pt plus 1pt minus 1pt}
  \setlength{\belowdisplayshortskip}{3pt plus 1pt minus 1pt}
  \setlength{\jot}{2pt}
}

\newcommand{\bbrevise}[1]{\textcolor{black}{#1}}

\title{Sampling Interventional Distributions with Orthogonal MMD Flow}

\author{Bariscan Bozkurt}
\author{Hugh Dance}
\author{Zonghao Chen}
\author{Alexandre Galashov}
\author{Arthur Gretton}
\author{Houssam Zenati}
\affil{Gatsby Computational Neuroscience Unit, University College London}
\date{}

\begin{document}

\maketitle

\begin{abstract}
Many causal inference problems require sampling from interventional outcome distributions rather than estimating low-dimensional summaries. Existing approaches often rely on fitting generative models of the target distribution together with additional nuisance components, which can be computationally demanding for high-dimensional outcomes. We propose Interventional Maximum Mean Discrepancy Flow (IMMD-Flow), a particle-based method that instead represents the target interventional distribution through its kernel mean embedding. This kernel embedding admits a doubly robust, Neyman-orthogonal estimator
with finite-sample guarantees and can be efficiently reused to sample from
the interventional distribution via MMD gradient flow. IMMD-Flow accommodates both hard and soft interventions, admits theoretical guarantees for estimation and particle approximation, and extends to high-dimensional image outcomes through frozen encoders. Across image-based benchmarks, IMMD-Flow matches or improves upon state-of-the-art generative approaches while showing significant speed advantages.
\end{abstract}

\section{Introduction}

Interventional sampling aims to generate outcomes from the distribution
induced by a specified intervention rather than from the observational
distribution. This is particularly useful for high-dimensional outcomes. For example, one
may wish to generate images with a given demographic attribute $A$ while
adjusting contextual features, such as clothing or background, so that their
distribution does not inherit spurious associations with $A$ from the
observational data. Such tasks arise in representational fairness and bias mitigation
\citep{xu2018fairgan,choi2020fair,bianchi2023easily}.

Recent approaches construct interventional samplers by specifying a
parameterized generative or density-based model, identifying from observational
data a training objective that targets the interventional law, and optimizing
this objective over the model parameters
\citep{kennedy2023semiparametric,melnychuk2023normalizing,martinez2024counterfactual,luedtke2025doublegen,dance2026debiased}.
This yields flexible samplers, but requires first learning a generative model in addition to the nuisance functions required for causal
adjustment. While recent work has simplified the generative learning problem via using flows from data-adaptive source distributions \citep{dance2026debiased}, we show the computational cost remains burdensome.

In this work, we introduce \emph{Interventional MMD Flow} (IMMD-Flow), a method for sampling from interventional distributions which bypasses the need to first learn an explicit generative model. The central idea is to formulate interventional sampling as a \emph{sample-based distributional optimization} problem, in which an empirical distribution of particles is directly optimized to match the target distribution. We instantiate this using Maximum Mean Discrepancy (MMD), for which the target distribution enters the objective only through its kernel mean embedding \citep{gretton2012kernel,sriperumbudur2010hilbert}. Thus, once the interventional kernel mean embedding is estimated from observational data, samples can be generated directly by evolving the particles under the corresponding MMD gradient flow \citep{arbel2019mmdflow}.\looseness=-1

The remaining difficulty is statistical: neither the interventional embedding nor samples from the interventional distribution are observed.  Under standard identification assumptions, the embedding can be
represented through either the treatment-assignment mechanism or a
conditional outcome embedding \citep{muandet2021counterfactual,zenati2025cpme}. However, each representation alone is sensitive to nuisance misspecification. We therefore combine them through a doubly robust, Neyman-orthogonal estimator
\citep{bang2005doubly,chernozhukov2018double,foster2023orthogonal}. This estimator has a simple closed form, can be computed efficiently
using a projected Nystr\"om approximation, and has bias of product order in the two nuisance errors. Since the target-dependent component of the squared MMD is linear in the interventional embedding, this second-order robustness carries directly over to the sampling objective. For smooth kernels, the same estimation error also controls perturbations of the induced transport velocity.\looseness=-1


Our main contributions are as follows: (i) we formulate interventional sampling as MMD-based distributional optimization and derive a doubly robust,
Neyman-orthogonal fixed target whose estimation error perturbs the transport
velocity only at product order; (ii) we develop IMMD-Flow and a projected
Nystr\"om implementation that reduces conditional embedding regression costs
and enables reuse across supported interventions; (iii) for smooth kernels,
we derive finite-particle guarantees that propagate target-estimation error
through the flow without requiring the estimated target to be a probability
embedding, yielding conditions for convergence to the true interventional
law; and (iv) we evaluate sampling accuracy, nuisance robustness, and
computational performance on synthetic and structured-outcome benchmarks.

Section~\ref{sec:problem-statement} defines the sampling problem,
Section~\ref{sec:interventional-mmd-flow} develops IMMD-Flow,
Section~\ref{sec:sampling-procedure} gives the practical procedure, and
Section~\ref{sec:consistency} provides the statistical and particle-level
analysis, and Section ~\ref{sec:numerical-experiments} empirically evaluate our method on several benchmarks.  We provide a more detailed discussion of related work in Appendix~\ref{sec:related_work}.

\section{Problem Statement}
\label{sec:problem-statement}

We study the problem of generating outcomes under a known soft intervention
from observational data. We first identify the interventional outcome law
and then formulate sampling from this law as optimization over probability
measures.

\noindent\textbf{Interventional target and identification.} 
Let $\gX$, $\gA$, and $\gY$ be Polish spaces, and write $\mathcal P(\gY)$ for the set of probability measures on $\gY$. Let $O=(X,A,Y)\sim P_0$, where $X$, $A$, and $Y$ denote pretreatment covariates, treatment, and outcome, and
\begin{equation}
P_0(dy,da,dx)
=
P_{Y\mid A,X}(dy| a,x)\,
\pi_0(da| x)\,
P_X(dx)
\label{eq:observed-law}
\end{equation}
where $P_X$ is the marginal law of $X$, $\pi_0(\cdot\mid x):=P_0(A\in\cdot\mid X=x)$ is the observational assignment mechanism, and $P_{Y\mid A,X}$ is the conditional outcome law.

Let $\{Y(a):a\in\gA\}$ be a jointly measurable potential-outcome process \citep{rubin1974estimating,rosenbaum1983central}. A fixed, known soft intervention $\pi$ is a Markov kernel $x\mapsto\pi(\cdot\mid x)$ from $\gX$ to $\gA$ \citep{Robins1986,diaz2012population}. Its interventional outcome law $P^\pi\in\mathcal P(\gY)$ is defined by
\begin{equation}
P^\pi(B)
=
\E\left[
    \int_{\gA}
    \mathbf 1\{Y(a)\in B\}\,
    \pi(da\mid X)
\right],
\label{eq:causal-target-law}
\end{equation}
where $B\subseteq\gY\ \text{ is Borel measurable}$. Thus, $P^\pi$ is the outcome law induced by replacing the observational assignment mechanism $\pi_0$ with the intervention $\pi$.

\begin{assumption}[Identification]
\label{assum:identification}
The following standard conditions hold
\citep{rosenbaum1983central,Robins1986,diaz2012population}: 
\begin{enumerate}[leftmargin=*,nosep]
    \item \emph{Consistency:} $Y=Y(A)$ almost surely.
    \item \emph{Conditional exchangeability:}
    $\{Y(a)\}_{a\in\gA}\perp A\mid X$.
    \item \emph{Overlap:}
    $\pi(\cdot\mid x)\ll\pi_0(\cdot\mid x)$
    for $P_X$-almost every $x$.\looseness=-1
\end{enumerate}
\end{assumption}

\begin{proposition}[Identification of the interventional law]
\label{prop:identification}
Under Assumption~\ref{assum:identification}, $P^\pi$ is identified from
$P_0$. For every measurable $B\subseteq\gY$,
\begin{equation}
P^\pi(B)
=
\int_{\gX}\int_{\gA}
P_{Y| A,X}(B| a,x)\,
\pi(da| x)\,
P_X(dx).
\label{eq:target-outcome-law}
\end{equation}
\end{proposition}

Since $P_{Y\mid A,X}$ is defined only
$P_X(dx)\pi_0(da\mid x)$-almost everywhere, overlap ensures that Equation \ref{eq:target-outcome-law} is invariant to its choice of version. Hence $P^\pi$ is a well-defined functional of $P_0$ and $\pi$.

\noindent\textbf{Interventional sampling as distributional optimization.}
Sampling from $P^\pi\in\mathcal P(\gY)$ means generating a random outcome $\widetilde Y\in\gY$ such that
\begin{equation}
\Pr(\widetilde Y\in B)
=
P^\pi(B),
\qquad
B\subseteq\gY\ \text{measurable}.
\label{eq:exact-sampling}
\end{equation}
Our goal is therefore to generate from the full interventional outcome
distribution, rather than to estimate a particular functional of it.

Although Proposition~\ref{prop:identification} identifies $P^\pi$,
direct samples from it are unavailable: the observed outcomes arise under
$\pi_0$, rather than under the intervention $\pi$. Transport-based sampling
provides a constructive alternative by progressively transforming samples
from a tractable reference distribution toward the target
\citep{marzouk2017sampling,ambrosio2008gradient}. This can be formalized as optimization over probability measures. The resulting formulation can be called \emph{sampling distributional optimization}
(SDO): choose a functional
\(
\mathcal F_\pi:\mathcal P(\gY)\to\mathbb R\cup\{+\infty\}
\)
whose minimizer is the interventional law,
\begin{equation}
P^\pi
\in
\argmin_{Q\in\mathcal P(\gY)}
\mathcal F_\pi(Q),
\label{eq:sampling-distributional-optimization}
\tag{SDO}
\end{equation}
and construct dynamics that decrease $\mathcal F_\pi$ from a distribution
that can be sampled directly
\citep{jordan1998variational,ambrosio2008gradient}.

Let $Q_0\in\mathcal P(\gY)$ denote such a reference distribution and
$(Q_t)_{t\geq0}$ the laws induced by the resulting dynamics. If the objective
converges to its minimum and $\mathcal F_\pi$ controls convergence of
probability measures, then \(Q_t\Rightarrow P^\pi
\enspace
\text{as }t\to\infty.\)
Hence, if $\widetilde Y_t\sim Q_t$, the evolved outputs become
asymptotically distributed according to $P^\pi$. Wasserstein gradient flows
provide a principled way to construct such measure-valued dynamics, while
particle approximations yield practical sampling algorithms
\citep{ambrosio2008gradient,liu2016stein,arbel2019mmdflow}.

The choice of $\mathcal F_\pi$ is therefore central: it must identify the
interventional target while remaining amenable to estimation and transport
without direct samples from $P^\pi$. In the next section, we show that
maximum mean discrepancy has this structure. Its dependence on $P^\pi$
reduces to a kernel mean embedding, which admits an orthogonal
representation under the observational law and induces the transport
dynamics used for sampling.

\section{Interventional MMD Flow}
\label{sec:interventional-mmd-flow}

The sampling distributional optimization formulation in \eqref{eq:sampling-distributional-optimization} formulation requires a distributional objective that identifies the interventional target and is estimable from observational data. We use maximum mean discrepancy (MMD) \citep{gretton2012kernel,sriperumbudur2010hilbert}, whose target dependence is linear in a kernel mean embedding. This structure permits an orthogonal observed-data representation \citep{chernozhukov2018double,foster2023orthogonal} and, subsequently, explicit transport dynamics \citep{arbel2019mmdflow,hertrich2024generative,galashov2024deep,chen2025stationary}.

\subsection{MMD objective for interventional sampling}
\label{subsec:interventional-mmd-risk}

To define the spatial dynamics used below, we work on
$\gZ=\mathbb R^{d_Z}$. Fix a measurable representation map $E:\gY\to\gZ$ and write \(Z:=E(Y), \enspace Q^\pi:=E_\#P^\pi,\)
where $E_\#P^\pi$ is the pushforward of $P^\pi$ through $E$. When $\gY=\mathbb R^{d_Z}$, taking $E=\mathrm{Id}$ recovers the outcome-space construction; otherwise, $E$ provides a latent space representation. The flow targets $Q^\pi$, and recovery of $P^\pi$ after decoding is treated separately. Both the representation and the kernel below are fixed throughout this section.

Let $k:\gZ\times\gZ\to\mathbb R$ be a bounded, continuous, positive-definite, characteristic kernel, with reproducing kernel Hilbert space $\gH$ and feature map $\varphi(z):=k(z,\cdot)$. Boundedness ensures that, for every $Q\in\mathcal P(\gZ)$, \(\mu_Q:=\int_{\gZ}\varphi(z)\,Q(dz),
\enspace
\chi(\pi):=\mu_{Q^\pi}\)
are well-defined elements of $\gH$. The latter is the interventional mean embedding, following the representation of causal distributions by \citet{muandet2021counterfactual,zenati2025cpme}. We take the sampling objective to be
\begin{equation}
\mathcal F_\pi(Q)
:=\frac12\|\mu_Q-\chi(\pi)\|_{\gH}^2
=\frac12\operatorname{MMD}_k^2(Q,Q^\pi).
\label{eq:interventional-mmd-objective}
\end{equation}
Characteristicness gives $\mathcal F_\pi(Q)=0$ if and only if $Q=Q^\pi$. For weak-consistency conclusions, we additionally require that $\operatorname{MMD}_k$ metrizes weak convergence; under this condition, vanishing objective values imply weak convergence to $Q^\pi$ \citep{simon2023metrizing}.

For observational estimation, the useful feature of Equation \ref{eq:interventional-mmd-objective} is its risk representation. Define \(\ell_Q(z)
:=\frac12\|\mu_Q\|_{\gH}^2
-\langle\mu_Q,\varphi(z)\rangle_{\gH}.\)
Its interventional risk
$R_\pi(Q):=\int_{\gZ}\ell_Q(z)\,Q^\pi(dz)$ satisfies
\begin{equation}
R_\pi(Q)-R_\pi(Q^\pi)
=\mathcal F_\pi(Q)
=\frac12\operatorname{MMD}_k^2(Q,Q^\pi).
\label{eq:excess-risk-mmd}
\end{equation}
Thus, excess interventional risk is exactly half the squared sampling
discrepancy. Moreover, the part of the objective that determines
optimization depends linearly on $\chi(\pi)$. Recovering this one
element therefore recovers the entire family of risks indexed by $Q$.

\subsection{Orthogonal representation of the interventional objective}
\label{subsec:orthogonal-interventional-objective}

Identification expresses $\chi(\pi)$ under the observational law, but
introduces unknown nuisance functions. Direct substitution of
approximate nuisances can introduce first-order bias into the sampling
objective. We instead seek an observed-data representation whose
first-order nuisance sensitivity vanishes, as in orthogonal statistical
learning \citep{foster2023orthogonal}.

Write $\mathbb E_0$ for expectation under $P_0$ and let
$P_{A,X}(da,dx)=\pi_0(da\mid x)P_X(dx)$. The relevant nuisances are
\begin{align}
m_0(a,x)
&:=\mathbb E_0[\varphi(Z)\mid A=a,X=x],
\quad
w_0(a,x)
:=\frac{d\pi(\cdot\mid x)}{d\pi_0(\cdot\mid x)}(a).\nonumber
\end{align}
Proposition~\ref{prop:identification} yields two representations,
\begin{align}
\chi(\pi)
&=\mathbb E_0\left[
    \int_{\gA}m_0(a,X)\,\pi(da\mid X)
\right] = \mathbb E_0[w_0(A,X)\varphi(Z)].
\label{eq:embedding-identification}
\end{align}
Either representation in Equation \ref{eq:embedding-identification} could be used to construct the target embedding. A regression plug-in relies entirely on $m_0$, while inverse weighting relies entirely on $w_0$; errors in either representation therefore enter the resulting target at first order. 

Instead, we combine the two through an augmented score, following the
doubly robust RKHS construction
\citep{luedtke2024one,zenati2025cpme}. For candidate nuisance functions
$m:\gA\times\gX\to\gH$ and
$w:\gA\times\gX\to\mathbb R$, write $\eta=(m,w)$ and define
\begin{align}
\Gamma_\eta(O)
&:=
\int_{\gA}m(a,X)\,\pi(da\mid X)
 +
w(A,X)\{\varphi(Z)-m(A,X)\},
\label{eq:orthogonal-score}
\end{align}
with population target \(h_\eta := \E_0[\Gamma_\eta(O)]
\in\gH.\)
The first term is the regression representation of the interventional
embedding; the second uses the observed feature residual to correct it
through intervention-to-observation reweighting.

\begin{assumption}[Nuisance integrability]
\label{assum:nuisance-integrability}
Let $P_{A,X}(da,dx):=\pi_0(da\mid x)P_X(dx)$. The nuisance functions satisfy $m,m_0\in L^2(P_{A,X};\gH)$, and $w,w_0\in L^2(P_{A,X}).$
\end{assumption}

Together with boundedness of $k$, Assumption~\ref{assum:nuisance-integrability} ensures that Equation \ref{eq:orthogonal-score} is Bochner integrable and $h_\eta$ is well defined. The $L^2$ condition on $w_0$ strengthens the absolute-continuity condition in Assumption~\ref{assum:identification} and controls the degree of intervention-to-observation reweighting.\looseness=-1

Importantly, since the interventional MMD risk depends on the target only through $\chi(\pi)$, the observed-data counterpart of the risk can be obtained simply by replacing
$\chi(\pi)$ with $h_\eta$:
\begin{align}
R(Q;\eta)
&:=
\frac12\|\mu_Q\|_{\gH}^2
-
\langle\mu_Q,h_\eta\rangle_{\gH}
=
\frac12\|\mu_Q-h_\eta\|_{\gH}^2
-
\frac12\|h_\eta\|_{\gH}^2.
\label{eq:orthogonal-immd-risk}
\end{align}
Thus, the same nuisance correction simultaneously defines the target for
every candidate distribution $Q$. This is the particular simplification
of MMD that we exploit: unlike a generic nuisance-dependent generative
risk, causal adjustment reduces to a single fixed Hilbert-space element.

Write $r_m:=\|m-m_0\|_{L^2(P_{A,X};\gH)}$, and $r_w:=\|w-w_0\|_{L^2(P_{A,X})}$.

\begin{proposition}[Double robustness and orthogonality]
\label{prop:orthogonal-interventional-target} Under Assumptions~\ref{assum:identification} and~\ref{assum:nuisance-integrability},
\begin{align}
h_\eta-\chi(\pi)
&=
-\E_0\!\left[(w-w_0)(m-m_0)\right],\nonumber
\\
\|h_\eta-\chi(\pi)\|_{\gH} &\le r_w r_m,
\tag{DR}
\label{eq:dr-target-remainder}
\end{align}
where the nuisance functions are evaluated at $(A,X)$. Hence $h_{m,w_0}=h_{m_0,w}=\chi(\pi)$, and therefore $R(Q;m,w_0)=R(Q;m_0,w)=R_\pi(Q)$ for every $Q$.

Moreover, viewing $R(Q;\eta)$ as a function of $u=\mu_Q$, its gradient with respect to \(u\) is $u-h_\eta$ and, for every square-integrable direction $\dot\eta$,
\begin{equation}
D_\eta\!\left(u-h_\eta\right)\big|_{\eta=\eta_0}[\dot\eta]
=0,
\qquad
u\in\gH,
\tag{ORTH}
\label{eq:universal-orthogonality}
\end{equation}
where $\eta_0=(m_0,w_0)$.
\end{proposition}

The remainder in Equation \ref{eq:dr-target-remainder} is bilinear in the two nuisance errors, which yields both double robustness and first-order orthogonality. The proof is given in Appendix~\ref{app:proof-orthogonal-interventional-target}.


Equation~\ref{eq:universal-orthogonality} is the form of Neyman
orthogonality relevant to the sampling problem: nuisance perturbations have
no first-order effect on the gradient of the risk being optimized
\citep{chernozhukov2018double,foster2023orthogonal}. In fact, the property
holds for every candidate embedding $u$, rather than only at the population
minimizer. Therefore the nuisance-induced perturbation remains second order
throughout the subsequent distributional optimization.

\begin{remark}
 This construction is related to doubly robust generative-risk correction \citep{luedtke2025doublegen}, but MMD yields an additional simplification: because the target-dependent term is linear in the kernel mean embedding, orthogonalizing the entire family of sampling risks reduces to constructing the single target element $h_\eta$. The next subsection shows that the same element also determines the force driving the MMD transport.\looseness=-1
\end{remark}

\subsection{Interventional MMD dynamics}
\label{subsec:interventional-mmd-dynamics}

We now turn the orthogonalized MMD objective into a transport sampler.
Because the target enters only through a fixed RKHS element, we derive the
dynamics for a generic \(h\in\gH\), covering both the exact target
\(h=\chi(\pi)\) and its orthogonal representation \(h=h_\eta\). We define
\begin{equation}
\mathcal F_h(Q)
:=
\frac12\|\mu_Q-h\|_{\gH}^2.
\label{eq:generic-mmd-objective}
\end{equation}
When $h=\chi(\pi)$, this is the interventional MMD objective in
Equation \ref{eq:interventional-mmd-objective}. When $h=h_\eta$, it is the
orthogonal population objective in
Equation \ref{eq:orthogonal-immd-risk}, up to a constant independent of $Q$.

To turn Equation \ref{eq:generic-mmd-objective} into a spatial flow, we impose
the following regularity conditions.

\begin{assumption}[Flow regularity]
\label{assum:immd-flow}
Let $\gZ=\mathbb R^{d_Z}$ and let $Q_0\in\mathcal P_2(\gZ)$.
The feature map $\varphi:\gZ\to\gH$ is continuously Fr\'echet
differentiable and, for finite constants $L_\varphi,L_D$,
\begin{align*}
\sup_{z\in\gZ}\|D\varphi(z)\|_{\mathrm{op}}
&\le L_\varphi,
\
\|D\varphi(z)-D\varphi(z')\|_{\mathrm{op}}
\le L_D\|z-z'\|_2.   
\end{align*}
For the exact interventional flow, assume additionally
$Q^\pi\in\mathcal P_2(\gZ)$.
\end{assumption}

These conditions ensure that the velocity fields below are well defined
and Lipschitz, so that the corresponding dynamics admit a
well-defined measure-valued evolution in $\mathcal P_2(\gZ)$
\citep{ambrosio2008gradient,arbel2019mmdflow}. It can also be relaxed to local Lipschitzness~\citep{tian2026sobolev}. 

For fixed $h$, define the target potential \(s_h(z)
:=
\langle h,\varphi(z)\rangle_{\gH}\)
and the associated MMD witness \(f_{Q,h}(z) :=
\langle\mu_Q-h,\varphi(z)\rangle_{\gH}=
\int_{\gZ}k(z,z')\,Q(dz')-s_h(z).\)
The first term depends on the current law $Q$, whereas $s_h$ is a fixed
target potential. Following MMD gradient flow
\citep{arbel2019mmdflow}, samples move along the negative spatial gradient
of the witness,
\begin{align}
v_h(Q,z)
&:=
-\nabla f_{Q,h}(z)
=
-\int_{\gZ}\nabla_1 k(z,z')\,Q(dz')
+\nabla s_h(z),\nonumber
\end{align}
where $\nabla_1$ denotes differentiation in the first argument.

We define the \emph{Interventional MMD Flow (IMMD-F)} as the law-valued
evolution induced by \(v_h\):
\begin{equation}
\partial_t Q_t^h
+
\nabla\cdot
\bigl(
Q_t^h v_h(Q_t^h,\cdot)
\bigr)
=
0,
\
Q_{t=0}^h=Q_0.
\tag{IMMD-F}
\label{eq:immd-flow}
\end{equation}
Under Assumption~\ref{assum:immd-flow}, this evolution admits the
characteristic representation
\begin{equation}
\dot Z_t
=
v_h(Q_t^h,Z_t),
\qquad
Z_0\sim Q_0, \nonumber
\end{equation}
with \(Z_t\sim Q_t^h\) for every \(t\geq0\). Hence the distributional
optimization can be realized by transporting samples from \(Q_0\), which
provides the basis for the interacting-particle procedure.

The generic target \(h\) covers both constructions of interest: \(h=\chi(\pi)\) gives the ideal interventional flow, while \(h=h_\eta\) gives its orthogonal counterpart. By Proposition~\ref{prop:orthogonal-interventional-target}, the two coincide whenever either nuisance is correct.\looseness=-1

We next characterize the descent of this flow and the effect of nuisance
perturbations on its velocity field.

\begin{proposition}[Dissipation and velocity stability]
\label{prop:immd-stability}
Under Assumption~\ref{assum:immd-flow}, any sufficiently regular solution
of \eqref{eq:immd-flow} satisfies
\begin{equation}
\frac{d}{dt}\mathcal F_h(Q_t^h)
=
-\int_{\gZ}
\|v_h(Q_t^h,z)\|_2^2\,Q_t^h(dz)
\le 0.
\label{eq:immd-dissipation}
\end{equation}
Moreover, for every $Q\in\mathcal P_2(\gZ)$,
\begin{align}
\sup_{z\in\gZ}
\|v_{h_\eta}(Q,z)-v_{\chi(\pi)}(Q,z)\|_2
&\le
L_\varphi\|h_\eta-\chi(\pi)\|_{\gH}\nonumber
\le
L_\varphi r_w r_m .
\tag{VEL}
\label{eq:orthogonal-velocity-bound}
\end{align}
\end{proposition}

Equation \ref{eq:immd-dissipation} shows that IMMD Flow decreases the
objective associated with its target $h$. As for standard MMD gradient
flow, this dissipation alone does not guarantee convergence to the global
minimizer \citep{arbel2019mmdflow}.

Equation \ref{eq:orthogonal-velocity-bound} gives the dynamical
consequence of orthogonality: nuisance error perturbs the transport
velocity only at product order. Controlling how this perturbation
propagates to trajectories and the generated distribution is deferred to
the convergence analysis.

\section{Sampling Procedure}
\label{sec:sampling-procedure}
\label{sec:algorithm-development}

We estimate the interventional target from observational data, then
hold it fixed while evolving particles. Note that no nuisance refitting or
additional sampling network is required during generation.

\subsection{Orthogonal target estimation}
\label{subsec:empirical-interventional-target}

Given $O_1,\ldots,O_n\overset{\mathrm{iid}}{\sim}P_0$, set
$Z_i=E(Y_i)$ and fit $\widehat\eta=(\widehat m,\widehat w)$.
We estimate the target by averaging the orthogonal score in
\eqref{eq:orthogonal-score}:
\begin{equation}
\widehat h_\pi
:=\frac1n\sum_{i=1}^n\Gamma_{\widehat\eta}(O_i).
\label{eq:dr-interventional-embedding}
\end{equation}
Our implementation fits and evaluates the nuisances on the same
observations; its statistical conditions are stated in the convergence
analysis.

\noindent\textbf{Projected Nystr\"om construction.}
To reduce regression cost, we fit $\widehat m$ in an input-side Nystr\"om subspace \citep{rudi2015less}, using kernels $k_X,k_A$, $m_L$ uniformly sampled landmarks, and ridge parameter $\lambda>0$. Intervention integrals are approximated with $J$ independent draws from $\pi(\cdot\mid X_i)$ per context. The resulting numerical target
and its potential are
\begin{align}
\widetilde h_\pi
&=\sum_{i=1}^n v_{\mathrm{DR},i}\varphi(Z_i),
\label{eq:numerical-interventional-target}\enspace \widetilde s_\pi(z)
=\sum_{i=1}^n v_{\mathrm{DR},i}k(z,Z_i).
\end{align}
Here $\widehat h_\pi$ uses the projected regression with exact intervention integrals, whereas $\widetilde h_\pi$ denotes its numerical realization. Appendix~\ref{sec:projected_nystrom_computation} gives the coefficient construction; the full-kernel alternative is in Appendix~\ref{sec:full-ridge-computation}. The coefficients may be signed and need not sum to one, so zero residual against $\widetilde h_\pi$ is not assumed.

\subsection{Particle generation}
\label{subsec:particle-generation}

We instantiate IMMD Flow with $h=\widetilde h_\pi$. Draw
$Y_j^{(0)}\overset{\mathrm{iid}}{\sim}P_{\rm init}$ and set
$Z_j^{(0)}=E(Y_j^{(0)})$, choosing $P_{\rm init}$ so that
$Q_0=E_\#P_{\rm init}\in\mathcal P_2(\gZ)$.
At iteration $r$, the current law is approximated by
$\widehat Q_M^{(r)}=M^{-1}\sum_{j=1}^M\delta_{Z_j^{(r)}}$.
Following \citet{arbel2019mmdflow} and \citet{chen2025stationary}, we optionally perturb the gradient
evaluation locations: set
$\widetilde Z_j^{(r)}=Z_j^{(r)}+\beta_rU_j^{(r)}$, with fresh,
independent $U_j^{(r)}\sim\mathcal N(0,I_{d_Z})$ and
$\beta_r=c_\beta(r+1)^{-\alpha}$, where $c_\beta\geq0$ and
$\alpha\in[0,1/2)$. For step sizes $\eta_r>0$, update
\begin{equation}
\begin{aligned}
G_j^{(r)}
&=\frac1M\sum_{\ell=1}^M
  \nabla_1k(\widetilde Z_j^{(r)},Z_\ell^{(r)})
  -\nabla\widetilde s_\pi(\widetilde Z_j^{(r)}),\\
Z_j^{(r+1)}&=Z_j^{(r)}-\eta_rG_j^{(r)}.
\end{aligned}
\label{eq:finite-particle-immd-update}
\end{equation}
All gradients use the same current particle cloud; only their
evaluation locations are perturbed. Setting $c_\beta=0$ recovers
the Euler particle update. The target coefficients remain fixed.

After $R$ updates, decode once with a fixed $D:\gZ\to\gY$ and return $\widehat Q_M^{(R)}$ and $\widehat P_M=D_\#\widehat Q_M^{(R)}$. Taking $E=D=\mathrm{Id}$ recovers outcome-space sampling. The generated particles need not be independent. Algorithm~\ref{alg:mmd-flow-interventional-sampler} specifies the smooth-kernel sampler; the Riesz--AdamW variant uses a distinct energy and update rule (see Appendix \ref{sec:riesz-adamw}).

\begin{algorithm}[t]
\caption{Projected-Nystr\"om IMMD sampler}
\label{alg:mmd-flow-interventional-sampler}
\begin{algorithmic}[1]
\REQUIRE Data $O_{1:n}$; intervention $\pi$; maps $E,D$;
kernels $k_X,k_A,k$; settings $m_L,J,\lambda,M,R$;
initial law $P_{\rm init}$; schedules $(\eta_r,\beta_r)$.
\STATE Set $Z_i\gets E(Y_i)$ for all $i$.
\STATE Fit $\widehat w$ and set
$\widehat{\boldsymbol w}\gets(\widehat w(A_i,X_i))_{i=1}^n$.
\STATE Select $m_L$ input landmarks without replacement.
\STATE Draw $J$ intervention actions per context.
\STATE Compute $b^{\rm proj}$ using the projected regression
in Appendix~\ref{sec:projected_nystrom_computation}.
\STATE Set $v_{\rm DR}\gets(b^{\rm proj}+\widehat{\boldsymbol w})/n$
and store $\widetilde s_\pi$ from Equation \ref{eq:numerical-interventional-target}.
\STATE Draw $Y_j^{(0)}\overset{\mathrm{iid}}{\sim}P_{\rm init}$;
set $Z_j^{(0)}\gets E(Y_j^{(0)})$ for all $j$.
\FOR{$r=0,\ldots,R-1$}
    \STATE Draw independent $U_j^{(r)}\sim\mathcal N(0,I_{d_Z})$.
    \STATE Set $\widetilde Z_j^{(r)}\gets
    Z_j^{(r)}+\beta_rU_j^{(r)}$ for all $j$.
    \STATE Compute all $G_j^{(r)}$ and update simultaneously
    using Equation \ref{eq:finite-particle-immd-update}.
\ENDFOR
\STATE Set $\widehat Y_j\gets D(Z_j^{(R)})$ for all $j$.
\RETURN $\widehat Q_M^{(R)}$ and $\widehat P_M$.
\end{algorithmic}
\end{algorithm}

\section{Statistical Guarantees}
\label{sec:consistency}

In this section, we bound the discrepancy between the generated empirical distribution and
$Q^\pi$, accounting for the error in the target used by the sampler.
Recall that $\widehat h_\pi$ in Equation \ref{eq:dr-interventional-embedding} is the
doubly robust target estimator with intervention integrals evaluated
exactly, whereas $\widetilde h_\pi$ in Equation \ref{eq:numerical-interventional-target} is its numerical approximation. 
Write
\begin{equation}
e_{n,J}:=\|\widetilde h_\pi-\chi(\pi)\|_{\gH}.\nonumber
\end{equation}
We first control sampling error in terms of $e_{n,J}$ in Section \ref{subsec:estimated-target-convergence}, then bound this
quantity using orthogonal target estimation in Section \ref{subsec:statistical-target-accuracy}. 
The representation, decoder, and kernels are fixed independently of the observations analyzed here.

\subsection{IMMD-Flow sampling error}
\label{subsec:estimated-target-convergence}

A growing body of work studies the MMD-flow error, establishing conditions
under which it vanishes as the number of particles $M$ and 
iterations $R$ increase. Existing approaches include entropy
regularization~\citep{suzuki2023convergence}, noise
injection~\citep{arbel2019mmdflow,chen2025stationary}, adaptive kernel
lengthscales~\citep{kang2026gradient}, Fisher--Rao
geometry~\citep{belhadji2026weighted,belhadji2026discretize}, and energy/Riesz
kernels~\citep{duong2026wasserstein,chizat2026quantitative,chodrondecourcel2026wasserstein,rosenzweig2026wasserstein}. 
For finite-particle analysis, we consider the noise-injection framework of
\citet{arbel2019mmdflow,chen2025stationary}, because it gives explicit finite-iteration
and finite-particle bounds for smooth kernels. 

\begin{assumption}[Smooth particle dynamics]
\label{ass:sg-particle-regularity}
The feature map $\varphi$ is twice continuously Fr\'echet differentiable.
Its norm and the operator norms of its first two derivatives are uniformly
bounded by $\kappa\geq1$. The update uses a constant step size
$\eta_r\equiv\tau>0$ and noise levels
$\beta_r=c_\beta(r+1)^{-\alpha}$, where
\[
12\tau\kappa^2d_Z\leq1,
\qquad
0<c_\beta\leq\frac{1}{2\sqrt{d_Z}},
\qquad
0\leq\alpha<\frac12.
\]
\end{assumption}

Smoothness alone does not ensure convergence to the target. As in
noise-injected MMD-flow analysis
\citep{arbel2019mmdflow,chen2025stationary}, we additionally impose a
gradient-dominance condition over the iterations being analyzed.
Let $\mathcal D$ contain all observations and target-construction
randomness, and let $\mathcal F_r$ additionally contain the initial
particles and the Gaussian perturbations before iteration $r$.

\begin{assumption}[Noisy gradient dominance]
\label{ass:sg-noisy-pl}
For $r=0,\ldots,R-1$, the particles generated with target
$\widetilde h_\pi$ satisfy, almost surely,
\begin{equation}
\begin{aligned}
&\frac1M\sum_{j=1}^M
\mathbb E\left[
\left\|v_{\chi(\pi)}\bigl(\widehat Q_M^{(r)},
Z_j^{(r)}+\beta_rU_j^{(r)}\bigr)\right\|_2^2
\,\middle|\,\mathcal F_r\right]
\geq
4\kappa^2d_Z\beta_r^2
\operatorname{MMD}_k^2(\widehat Q_M^{(r)},Q^\pi),
\end{aligned}
\label{eq:true-target-noisy-pl}
\end{equation}
where $U_j^{(r)}$ are independent standard Gaussian vectors. 
\end{assumption}

This condition evaluates the true-target velocity along the estimated-target particle process. It is an additional optimization assumption, rather than a consequence of the prescribed noise schedule or nuisance orthogonality. We impose this condition to establish a favorable convergence rate, rather than because it is necessary for the good performance of MMD flows. Indeed, as shown by \citet{logarajah2026optimisation}, a sufficiently large number of interacting particles can also guarantee good performance without a prescribed noise schedule.

\begin{theorem}[Estimated-target particle bound]
\label{thm:dr_flow}
Under Assumptions~\ref{ass:sg-particle-regularity}
and~\ref{ass:sg-noisy-pl}, there exist constants $C_0,C_1,C_2>0$
depending only on $\kappa,d_Z,\tau,c_\beta,\alpha$ such that, for
$R\geq1$ and $\delta\in(0,1)$,
\begin{align}
&\operatorname{MMD}_k(\widehat Q_M^{(R)},Q^\pi)
\leq C_0 \exp\left(- \frac{\tau \kappa^2 d_Z c_\beta^2}{16(1-2 \alpha)} R^{1-2\alpha} \right)  + (1+C_1R^\alpha)e_{n,J} +C_2R^\alpha
 \sqrt{\frac{\log(R/\delta)}{M}}.
\label{eq:estimated-target-particle-bound}
\end{align}
For each realization of $\mathcal D$ satisfying
$\|\widetilde h_\pi\|_{\gH}\leq\kappa$, this bound holds with
conditional probability at least $1-\delta$ over particle initialization
and update noise.
\end{theorem}

Proof is given in
Appendix~\ref{sec:estimated-target-flow-proof}. 
The three terms describe optimization decay, propagated numerical target error, and finite-particle error.  
Increasing the iteration number $R$ reduces the first optimization error term, but amplifies the second target numerical error. 
The boundedness condition that $\|\tilde h_\pi\|_{\mathcal H}\leq\kappa$ holds with probability tending to one under the assumptions of the next Corollary~\ref{cor:sg-sampling-consistency}, as established in Equation~\ref{eq:tilde_h_bound}.

\subsection{Statistical and numerical target error}
\label{subsec:statistical-target-accuracy}

To apply Theorem~\ref{thm:dr_flow}, we control the numerical target relative to $\chi(\pi)$. Let $\widehat\eta_n=(\widehat m_n,\widehat w_n)$ denote the fitted nuisances, and write $P_ng=n^{-1}\sum_i g(O_i)$ and $P_0g=\mathbb E_0[g(O)]$.\looseness=-1

\begin{assumption}[Nuisance estimation and sample reuse]
\label{ass:sg-statistical-target}
For deterministic rates $r_{m,n},r_{w,n}$,
\begin{align*}
\|\widehat m_n-m_0\|_{L^2(P_{A,X};\gH)}&=O_p(r_{m,n}), \quad \|\widehat w_n-w_0\|_{L^2(P_{A,X})}=O_p(r_{w,n}).
\end{align*}
The oracle score has finite second moment, and
\begin{equation}
\left\|(P_n-P_0)
\{\Gamma_{\widehat\eta_n}-\Gamma_{\eta_0}\}\right\|_{\gH}
=O_p(n^{-1/2}),
\label{eq:same-sample-score-control}
\end{equation}
where $\eta_0=(m_0,w_0)$.
\end{assumption}

Condition~\eqref{eq:same-sample-score-control} accounts for fitting and evaluating the nuisances on the same observations; a regression prediction rate alone does not establish it. For intervention integration, define $\bar m_n(X_i):=\int\widehat m_n(a,X_i)\,\pi(da\mid X_i)$ and
\begin{equation}
V_n:=\frac1n\sum_{i=1}^n
\int\|\widehat m_n(a,X_i)-\bar m_n(X_i)\|_{\gH}^2
\,\pi(da\mid X_i).\nonumber
\end{equation}

\begin{proposition}[Accuracy of the numerical target]
\label{prop:sg-target-rate}
Under Assumptions~\ref{assum:identification},
\ref{assum:nuisance-integrability}, and
\ref{ass:sg-statistical-target}, suppose that, conditional on the
observations, landmarks, and fitted nuisances, intervention integration
uses $J$ independent draws per context, also independent across contexts,
and that $V_n=O_p(1)$. For exact projected regression and its finite-$J$
target,
\begin{equation}
e_{n,J}
=O_p\left(n^{-1/2}+r_{m,n}r_{w,n}+(nJ)^{-1/2}\right).\nonumber
\end{equation}
\end{proposition}

The product term follows from \eqref{eq:dr-target-remainder}; the
Monte Carlo term follows from the conditional mean-square error
$V_n/(nJ)$ in Lemma~\ref{lem:finite-J-policy-integration}.
Thus a fixed $J\geq1$ preserves the statistical rate when $V_n$ is tight.
Under the projected Nystr\"om conditions of
Theorem~\ref{thm:nystrom-cme-rate}, substituting
$r_{m,n}=n^{-1/[2(1+\gamma)]}$ gives
\begin{equation}
e_{n,J}
=O_p\left(n^{-1/2}
+r_{w,n}n^{-\frac1{2(1+\gamma)}}+(nJ)^{-1/2}\right),
\label{eq:sg-nystrom-target-rate}
\end{equation}
where $\gamma$ is the input-capacity exponent; see Assumption \ref{ass:nystrom-input-regularity}. This specialization requires
the stated ridge and growing-landmark conditions. 

\subsection{Consistency}
\label{subsec:sg-sampling-consistency}

Now, we are ready to combine Theorem~\ref{thm:dr_flow} and  Proposition~\ref{prop:sg-target-rate} to obtain the final consistency result. 
To give an explicit choice of computational budgets $R$ and particle number $M$ as a function of the target sample size $n$, suppose that the nuisance-error product satisfies
$r_{m,n}r_{w,n}=O(n^{-s})$ for some $s\in(0,1/2]$. 



\begin{corollary}
\label{cor:sg-sampling-consistency}
Suppose the assumptions of Theorem~\ref{thm:dr_flow} and
Proposition~\ref{prop:sg-target-rate} hold over the chosen iteration
horizons, with constants independent of $n$. \textcolor{black}{We further take
$\kappa>\sup_{z\in\gZ}\|\varphi(z)\|_{\mathcal H}$.}
Assume that $r_{m,n}r_{w,n}=O(n^{-s})$ for some $s\in(0,1/2]$,
and fix $J\geq1$. 
\textcolor{black}{Let $C_R=\frac{16 s(1-2 \alpha)}{\tau \kappa^2 d_Z c_\beta^2}$}. 
Choose the iteration and particle budgets as
\begin{equation}
R_n=\left\lceil( \textcolor{black}{C_R} \log n)^{1/(1-2\alpha)}\right\rceil,
\quad
M_n=\left\lceil n^{2s}\log\log n\right\rceil,
\label{eq:sg-joint-schedule}
\end{equation}
Then, $\operatorname{MMD}_k
\bigl(\widehat Q_{M_n}^{(R_n)},Q^\pi\bigr)
=
O_p\!\left(n^{-s}(\log n)^{\alpha/(1-2\alpha)}\right).$
\end{corollary}

If $\operatorname{MMD}_k$ metrizes weak convergence~\citep{simon2023metrizing}, the generated
empirical distribution also converges weakly to the interventional distribution $Q^\pi$.


%
Finally, for encoded outcomes, the remaining step is to transfer latent accuracy
through the decoder.

\begin{proposition}[Outcome-space transfer]
\label{thm:EncoderPullbackMMDTransfer}
Let $k_E(y,y'):=k(E(y),E(y'))$. If $E,D$ are measurable and $\varphi$
is $L_k$-Lipschitz, then, \(\operatorname{MMD}_{k_E}(D_\#Q,P^\pi)
\leq\operatorname{MMD}_k(Q,Q^\pi)+L_k\mathbb E_{Z\sim Q}\|E(D(Z))-Z\|_2.\)
\end{proposition}

Applying this bound to $Q=\widehat Q_M^{(R)}$ adds only the displayed cycle-reconstruction error to the latent guarantee. When $E$ is non-injective, convergence in $k_E$ need not identify the full outcome law. If $D$ is continuous and $D_\#Q^\pi=P^\pi$, latent weak consistency also yields weak consistency of the decoded outcomes. Taking $E=D=\mathrm{Id}$ removes this transfer step.

\section{Numerical Experiments}
\label{sec:numerical-experiments}

\noindent\textbf{Methods.}
We evaluate our orthogonal variant of our IMMD-Flow method, that we write DR--IMMD (for the Doubly Robust embedding) together with two variants of MMD flow using plug-in estimates of the interventional embedding. Plug-in--IMMD constructs the fixed target using only the fitted conditional mean embedding, whereas IPW--IMMD uses only the estimated intervention-to-logging density ratio. DR--IMMD combines both through the orthogonal score in Section~\ref{sec:interventional-mmd-flow}. These variants differ only in the target coefficients supplied to the flow. Their exact constructions are given in Appendix~\ref{sec:orthogonal-target-computation}.

\begin{table}[!ht]
\centering
\caption{
Comparison on ColorMNIST, CelebA, and dSprites. Lower is better; both \(\mathrm{SW}_2\) metrics are multiplied by \(10^3\). Entries are mean \(\pm\) sample standard deviation, \textemdash\ denotes an unavailable result. Boldface marks the lowest reported mean within each benchmark and metric.
}
\label{tab:main-results}
\scriptsize
\setlength{\tabcolsep}{1.8pt}
\renewcommand{\arraystretch}{1.08}

\begin{adjustbox}{max width=\textwidth}
\begin{tabular}{@{}lccccccccc@{}}
\toprule
&
\multicolumn{3}{c}{\textbf{ColorMNIST}}
&
\multicolumn{3}{c}{\textbf{CelebA}}
&
\multicolumn{3}{c}{\textbf{dSprites}}
\\
\cmidrule(lr){2-4}
\cmidrule(lr){5-7}
\cmidrule(lr){8-10}
Method
& \(\mathrm{SW}_2\)
& Lat.\ \(\mathrm{SW}_2\)
& FID
& \(\mathrm{SW}_2\)
& Lat.\ \(\mathrm{SW}_2\)
& FID
& \(\mathrm{SW}_2\)
& Lat.\ \(\mathrm{SW}_2\)
& FID
\\
\midrule

DR--IMMD
& \(\mathbf{9.30\!\pm\!0.30}\)
& \(80.04\!\pm\!1.14\)
& \(30.644\!\pm\!0.809\)
& \(\mathbf{26.56\!\pm\!1.55}\)
& \(\mathbf{51.29\!\pm\!1.36}\)
& \(11.567\!\pm\!0.085\)
& {\(\mathbf{8.41\!\pm\!0.08}\)}
& {\(39.06\!\pm\!0.47\)}
& {\(36.488\!\pm\!1.648\)}
\\

Plug-in--IMMD
& \(9.31\!\pm\!0.32\)
& \(81.25\!\pm\!1.29\)
& \(\mathbf{30.531\!\pm\!0.843}\)
& \(27.47\!\pm\!1.23\)
& \(66.04\!\pm\!1.39\)
& \(11.861\!\pm\!0.074\)
& {\(8.42\!\pm\!0.03\)}
& {\(\mathbf{39.00\!\pm\!0.43}\)}
& {\(36.409\!\pm\!1.770\)}
\\

IPW--IMMD
& \(9.40\!\pm\!0.14\)
& \(93.05\!\pm\!10.92\)
& \(30.599\!\pm\!1.293\)
& \(\mathbf{26.56\!\pm\!1.55}\)
& \(\mathbf{51.29\!\pm\!1.36}\)
& \(11.567\!\pm\!0.085\)
& {\(8.47\!\pm\!0.09\)}
& {\(39.52\!\pm\!0.56\)}
& {\(\mathbf{36.391\!\pm\!1.774}\)}
\\

\addlinespace[1pt]

DecFM--Plugin
& \(11.37\!\pm\!0.78\)
& \(82.06\!\pm\!7.54\)
& \(32.005\!\pm\!1.075\)
& \(32.72\!\pm\!1.83\)
& \(62.20\!\pm\!0.71\)
& \(\mathbf{11.497\!\pm\!0.156}\)
& \textemdash
& \textemdash
& \textemdash
\\

DecFM--DR-I
& \(13.92\!\pm\!0.90\)
& \(92.17\!\pm\!6.57\)
& \(33.650\!\pm\!1.151\)
& \(33.56\!\pm\!3.18\)
& \(62.41\!\pm\!4.05\)
& \(14.609\!\pm\!0.525\)
& \textemdash
& \textemdash
& \textemdash
\\

DecFM--DR-OT
& \(11.90\!\pm\!1.06\)
& \(\mathbf{77.94\!\pm\!3.84}\)
& \(33.824\!\pm\!1.219\)
& \(36.36\!\pm\!2.45\)
& \(67.46\!\pm\!2.57\)
& \(15.521\!\pm\!1.307\)
& \textemdash
& \textemdash
& \textemdash
\\

\bottomrule
\end{tabular}
\end{adjustbox}
\end{table}

We compare against Deconfounding Flow Matching (DecFM)~\cite{dance2026debiased}, which is formulated for arm-specific
counterfactual distributions corresponding to hard interventions
$A\leftarrow a$. Since hard interventions are a special case of our framework,
with $\pi(\cdot\mid x)=\delta_a$, we specialize IMMD-Flow accordingly for
ColorMNIST and CelebA. For structured outcomes, we run DecFM in the same fixed
latent representation, encoding with $E$ and decoding with $D$, and report its
plug-in, DR-I, and DR-OT variants. We do not report DecFM on dSprites, whose
target is a context-dependent soft intervention $\pi(\cdot\mid X)$.

\begin{figure}[!b]
\centering
\begingroup
\setlength{\tabcolsep}{0.6pt}
\setlength{\fboxsep}{0pt}
\setlength{\fboxrule}{0.25pt}
\begin{minipage}[c]{0.47\linewidth}
\centering
\begin{tabular}{@{}rcccc@{}}
& \bbrevise{\(r=0\)}
& \bbrevise{\(r=250\)}
& \bbrevise{\(r=500\)}
& \bbrevise{\(r=3000\)} \\
\raisebox{0.1025\linewidth}{\bbrevise{\(a=0\)}}
& \fbox{\includegraphics[width=0.195\linewidth,trim={7.200bp 263.160bp 516.961bp 544.320bp},clip]{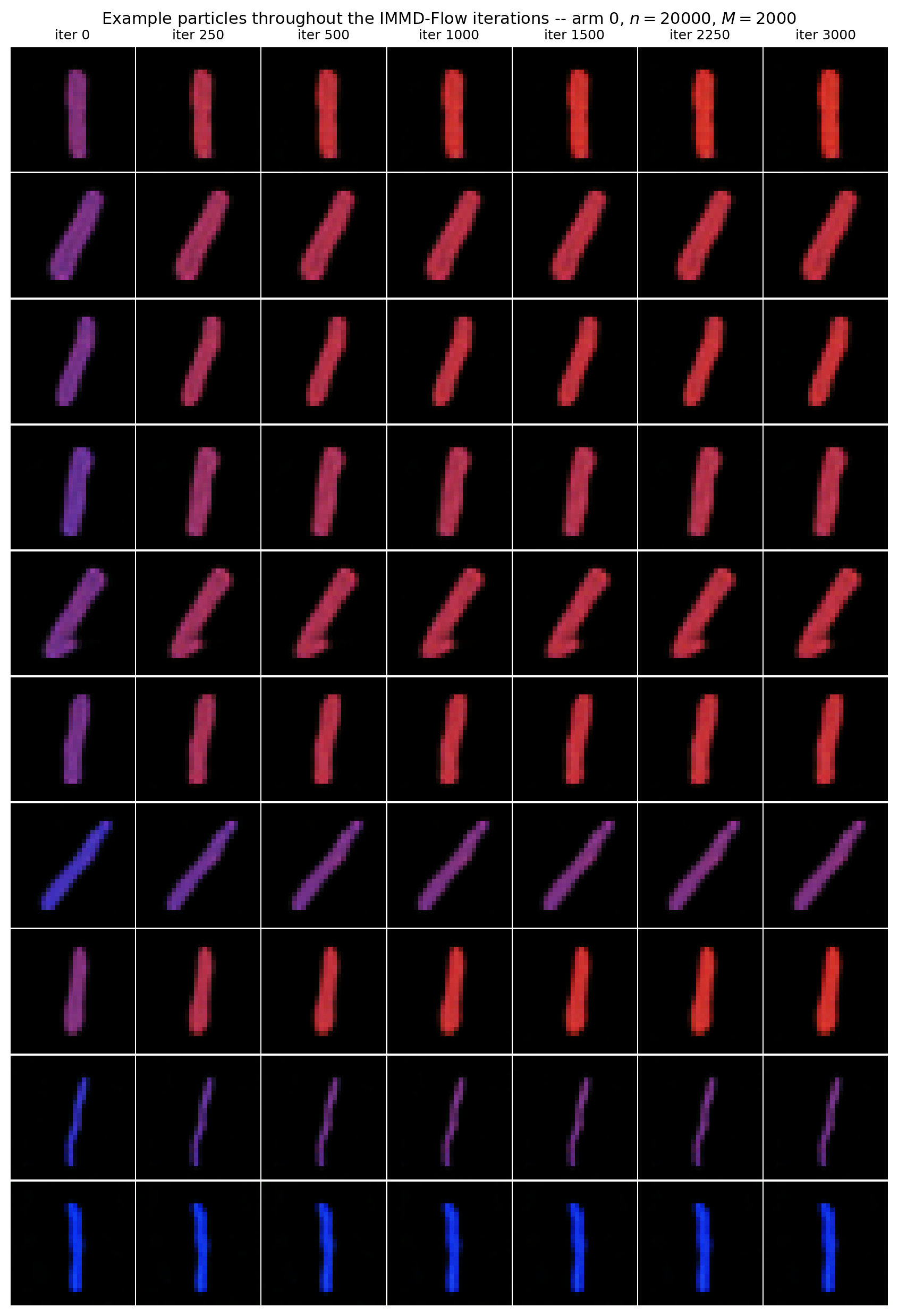}}
& \fbox{\includegraphics[width=0.195\linewidth,trim={92.160bp 263.160bp 432.001bp 544.320bp},clip]{figures/colormnist_flow_arm0_n20000.png}}
& \fbox{\includegraphics[width=0.195\linewidth,trim={177.120bp 263.160bp 347.041bp 544.320bp},clip]{figures/colormnist_flow_arm0_n20000.png}}
& \fbox{\includegraphics[width=0.195\linewidth,trim={517.321bp 263.160bp 6.840bp 544.320bp},clip]{figures/colormnist_flow_arm0_n20000.png}} \\
\raisebox{0.1025\linewidth}{\bbrevise{\(a=1\)}}
& \fbox{\includegraphics[width=0.195\linewidth,trim={7.200bp 92.520bp 516.961bp 715.320bp},clip]{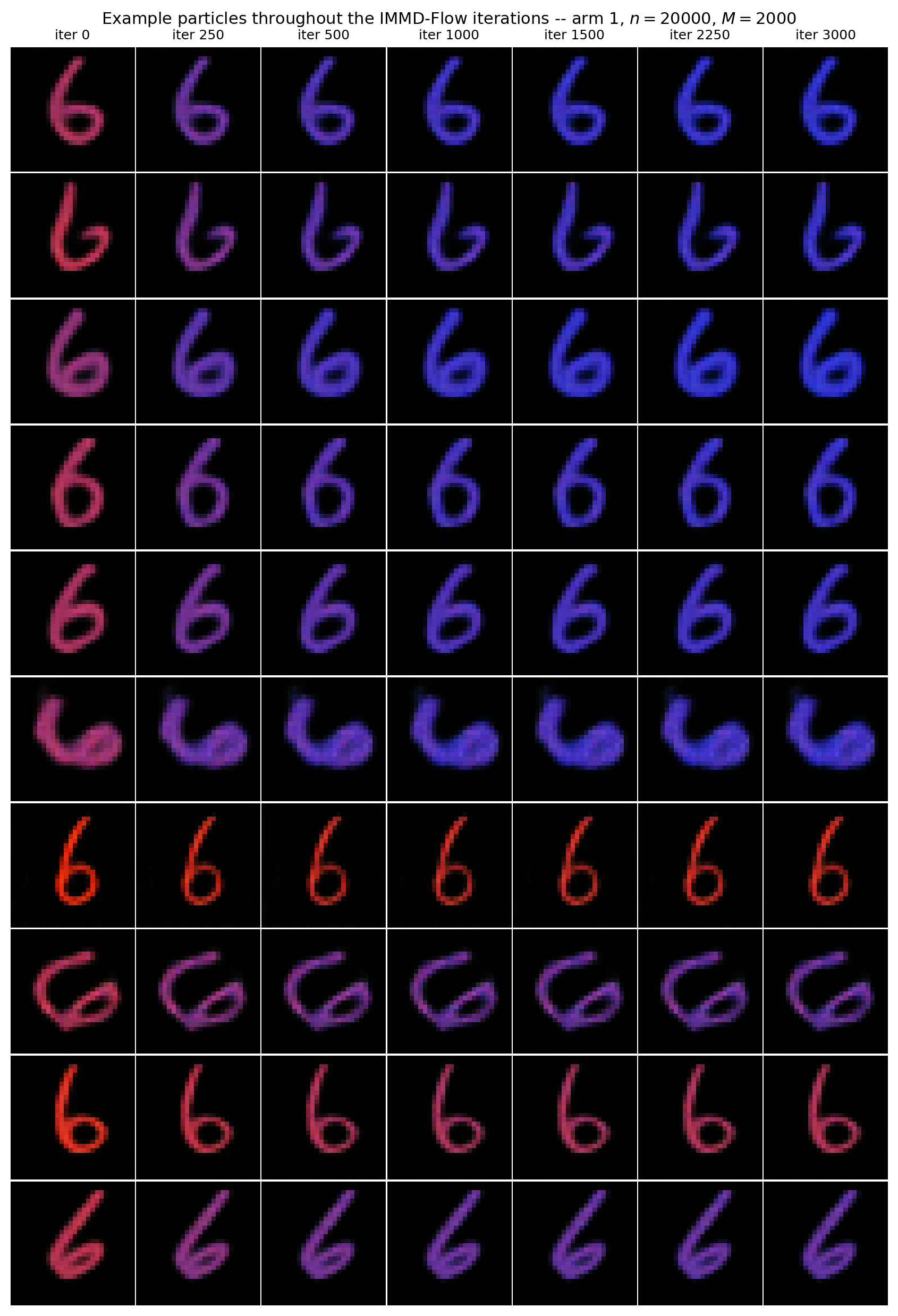}}
& \fbox{\includegraphics[width=0.195\linewidth,trim={92.160bp 92.520bp 432.001bp 715.320bp},clip]{figures/colormnist_flow_arm1_n20000.png}}
& \fbox{\includegraphics[width=0.195\linewidth,trim={177.120bp 92.520bp 347.041bp 715.320bp},clip]{figures/colormnist_flow_arm1_n20000.png}}
& \fbox{\includegraphics[width=0.195\linewidth,trim={517.321bp 92.520bp 6.840bp 715.320bp},clip]{figures/colormnist_flow_arm1_n20000.png}}
\end{tabular}
\end{minipage}\hfill%
\begin{minipage}[c]{0.53\linewidth}
\centering
\includegraphics[width=\linewidth]{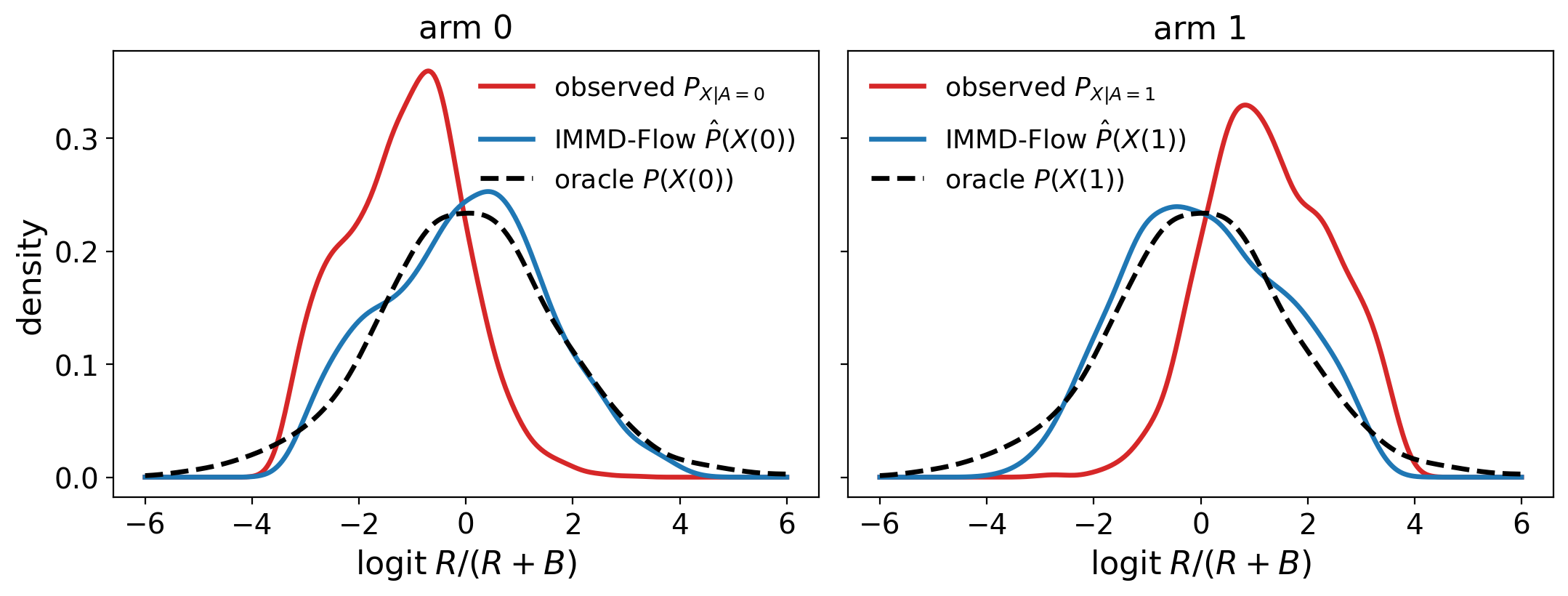}
\end{minipage}
\endgroup
\caption{\bbrevise{ColorMNIST DR--IMMD at \(n=20{,}000\) and \(M=2{,}000\). Left: two illustrative particles initialized from \(P_{Y\mid A=a}\); their foreground colors change during optimization while their digit identities are preserved. Right: the empirical foreground-color laws move away from the confounded observational arms and toward the oracle interventional laws. The displayed particle paths are algorithmic trajectories, not identified individual-level counterfactuals. Additional examples appear in Appendix~\ref{sec:qualitative-results}.}}
\label{fig:colormnist-colour-law}
\end{figure}

\noindent\textbf{Benchmarks.}
We evaluate the methods on three image-valued deconfounding tasks. ColorMNIST, adapted from \citet{dance2026debiased}, couples treatment-specific digit identity with a context variable that affects both treatment assignment and foreground colour. The interventional target fixes the digit class while averaging over the population distribution of the colour-generating context, thereby removing the observational colour imbalance.

The CelebA benchmark of \citet{dance2026debiased} sets \(A=\mathrm{Sex}\) and \(X=\mathrm{HairColor}\), with the observational data constructed so that hair color has different distributions across the two sex groups. For each target value \(a\), the interventional law replaces the group-specific context distribution \(P_{X\mid A=a}\) underlying \(P_{Y\mid A=a}\) by the population distribution \(P_X\). The task is therefore to generate images with attribute \(A=a\) while correcting the observational hair-colour imbalance. Finally, we consider the continuous actions setting  \citet{zenati2025cpme} over soft interventions. The context determines sprite shape and scale and also affects a stochastic action controlling position. The target policy changes the conditional action distribution and hence the joint distribution of visual content and position. Full descriptions of the data-generating processes are given in Appendix~\ref{sec:benchmark-constructions}.

\noindent\textbf{Results.}
We evaluate generated samples using empirical sliced
$2$-Wasserstein distance in decoded pixel space ($\mathrm{SW}_2$)
and encoder latent space (Lat.\ $\mathrm{SW}_2$)
\citep{Bonneel2014SlicedAR}, together with FID on decoded
images \citep{heusel2017gans}; lower is better.
All experiments in this section use $n=20{,}000$,
$M=2{,}000$, and three data seeds, with results for
static interventions averaged over the two arms within
each seed.

\medskip
\noindent\textbf{Sample quality.}
Table~\ref{tab:main-results} shows that all IMMD variants
outperform every DecFM baseline in pixel-space
$\mathrm{SW}_2$ on ColorMNIST and CelebA.
DR--IMMD also achieves the best latent $\mathrm{SW}_2$
on CelebA, lower FID than all DecFM variants on
ColorMNIST, and comparable FID to the best DecFM result
on CelebA.
Across all three benchmarks, DR--IMMD performs
competitively with the better single-nuisance variant,
without the benchmark-specific degradations observed
for Plug-in--IMMD on CelebA and IPW--IMMD on ColorMNIST.

\medskip
\noindent\textbf{Computational efficiency.} A central advantage of IMMD-Flow is that it avoids training an intervention-specific generative model. Once the nuisance functions have been estimated, the target kernel embedding is constructed directly and held fixed, and samples are generated by evolving a finite collection of particles under the resulting MMD flow. In contrast, DecFM learns a neural velocity field through flow matching for the interventional distribution.

Table~\ref{tab:compute-time-comparison} compares one-time fitting and post-fitting sample generation on ColorMNIST and CelebA. For DR--IMMD, fitting comprises learning the nuisance functions---the conditional mean embedding and intervention-to-logging density ratio---and constructing the fixed orthogonal RKHS target. During sampling, these quantities remain fixed while \(M=2{,}000\) particles are iteratively transported under the resulting MMD flow. For DecFM, fitting instead consists of training the neural velocity field through flow matching, while sampling transports particles by integrating the learned flow. Shared representation preprocessing, decoding, and metric evaluation are excluded from all reported timings. At \(n=20{,}000\) and \(M=2{,}000\), DR--IMMD reduces total fitting-and-generation time by \(11.0\)--\(42.2\times\) on ColorMNIST and \(12.6\)--\(41.3\times\) on CelebA relative to the DecFM variants. This advantage is driven by the fitting stage: DR--IMMD constructs its fixed Hilbert-space target in seconds, whereas DecFM must train a neural flow-matching model. Conversely, once fitting is complete, DecFM generates samples faster than the iterative IMMD particle optimizer. The comparison therefore demonstrates a substantial end-to-end computational advantage at the reported sampling scale, rather than an advantage in post-fitting sampling time.

\begin{table}[H]
\centering
\caption{End-to-end wall-clock comparison at \(n=20{,}000\) and \(M=2{,}000\). Within each benchmark, all methods were run on the same hardware: an NVIDIA L40S for ColorMNIST and an NVIDIA A100-SXM4-40GB for CelebA. Each entry covers both intervention arms. Fitting and sampling times are means \(\pm\) sample standard deviations over three data seeds; totals are sums of the reported means. The final column reports total time relative to DR--IMMD.}
\label{tab:compute-time-comparison}
\setlength{\tabcolsep}{6pt}
\renewcommand{\arraystretch}{1.1}
\begin{tabular}{@{}llrrrr@{}}
\toprule
Benchmark & Method & Fitting (s) & Sampling (s) & Total (s) & $\times$ DR--IMMD \\
\midrule
\multirow{4}{*}{ColorMNIST}
& DR--IMMD (ours) & $2.8\!\pm\!0.7$    & $35.6\!\pm\!0.0$  & $38.4$   & $1.0$ \\
& DecFM Plugin    & $422.6\!\pm\!108.1$ & $0.59\!\pm\!0.00$ & $423.2$            & $11.0$ \\
& DecFM DR        & $1560.0\!\pm\!105.1$ & $0.77\!\pm\!0.04$ & $1560.8$          & $40.6$ \\
& DecFM DR--OT    & $1621.0\!\pm\!108.7$ & $0.77\!\pm\!0.04$ & $1621.8$          & $42.2$ \\
\midrule
\multirow{4}{*}{CelebA}
& DR--IMMD (ours) & $1.8\!\pm\!0.6$      & $698.7\!\pm\!0.2$ & $700.5$ & $1.0$ \\
& DecFM Plugin    & $8823.6\!\pm\!216.0$ & $25.98\!\pm\!0.22$ & $8849.5$         & $12.6$ \\
& DecFM DR        & $27104.4\!\pm\!410.4$ & $21.56\!\pm\!0.07$ & $27126.0$       & $38.7$ \\
& DecFM DR--OT    & $28882.8\!\pm\!550.8$ & $21.56\!\pm\!0.07$ & $28904.4$       & $41.3$ \\
\bottomrule
\end{tabular}
\end{table}

\medskip
\noindent\textbf{Particle evolution.}
Following \citet{dance2026debiased}, we initialize particles from the observational distribution $P_{Y\mid A=a}$, which already contains much of the image structure present in the interventional target. Figure~\ref{fig:colormnist-colour-law} illustrates the resulting deconfounding process on ColorMNIST. As the particles evolve, their foreground colours change while their digit identities are largely preserved. At the population level, the colour distribution moves from the confounded observational law towards the oracle interventional distribution $P_X$. Thus, the flow primarily corrects the observational colour imbalance while preserving the remaining image structure.

\begin{figure}[t]
\centering
\begingroup
\setlength{\tabcolsep}{0.7pt}
\setlength{\fboxsep}{0pt}
\setlength{\fboxrule}{0.25pt}
\begin{tabular}{@{}rccccccc@{}}
& \bbrevise{\(r=0\)}
& \bbrevise{\(r=400\)}
& \bbrevise{\(r=800\)}
& \bbrevise{\(r=1600\)}
& \bbrevise{\(r=2400\)}
& \bbrevise{\(r=3200\)}
& \bbrevise{\(r=4000\)} \\
\raisebox{0.060\linewidth}{\bbrevise{\(a=0\)}}
& \fbox{\includegraphics[width=0.120\linewidth]{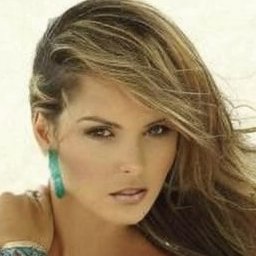}}
& \fbox{\includegraphics[width=0.120\linewidth]{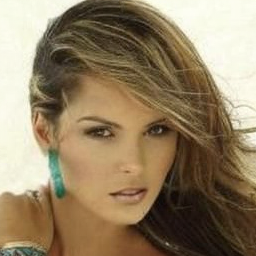}}
& \fbox{\includegraphics[width=0.120\linewidth]{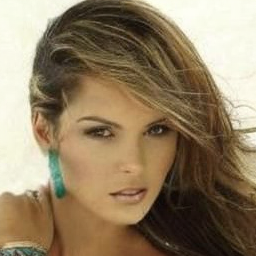}}
& \fbox{\includegraphics[width=0.120\linewidth]{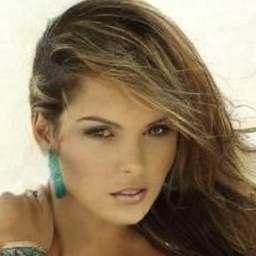}}
& \fbox{\includegraphics[width=0.120\linewidth]{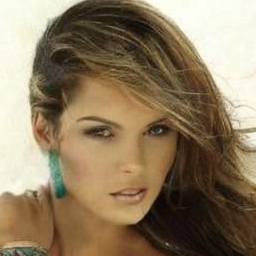}}
& \fbox{\includegraphics[width=0.120\linewidth]{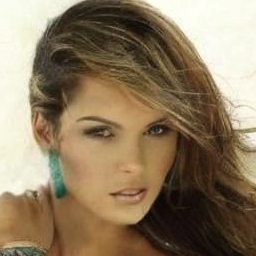}}
& \fbox{\includegraphics[width=0.120\linewidth]{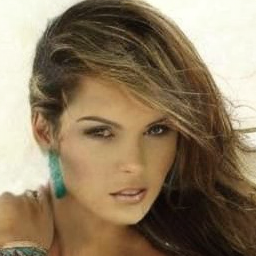}} \\
\raisebox{0.060\linewidth}{\bbrevise{\(a=1\)}}
& \fbox{\includegraphics[width=0.120\linewidth]{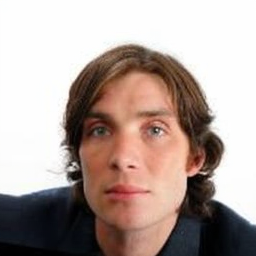}}
& \fbox{\includegraphics[width=0.120\linewidth]{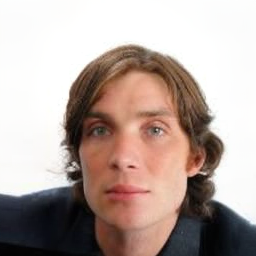}}
& \fbox{\includegraphics[width=0.120\linewidth]{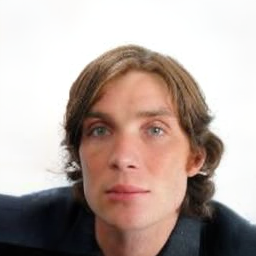}}
& \fbox{\includegraphics[width=0.120\linewidth]{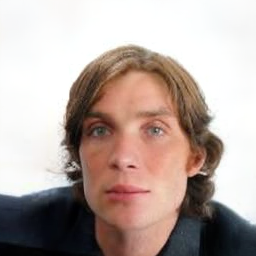}}
& \fbox{\includegraphics[width=0.120\linewidth]{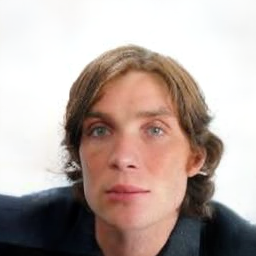}}
& \fbox{\includegraphics[width=0.120\linewidth]{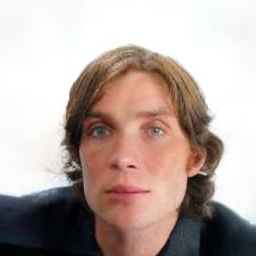}}
& \fbox{\includegraphics[width=0.120\linewidth]{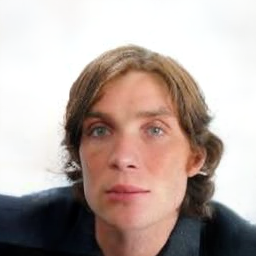}}
\end{tabular}
\endgroup
\caption{\bbrevise{Decoded CelebA particles during DR--IMMD optimization at \(n=20{,}000\) and \(M=2{,}000\). Particles are initialized from the corresponding observational arm \(P_{Y\mid A=a}\), and each row follows one numerical particle across optimization iterations as the empirical particle law is transported toward the interventional law \(P_{Y(a)}\). Note that these are not identified individual-level counterfactual trajectories, but rather represent the distributional change needed for these particles to represent valid samples from the interventional distribution. Additional particle trajectories appear in Appendix~\ref{fig:celeba-particle-evolution}.}}
\label{fig:celeba-particle-evolution-main}
\end{figure}

Figure~\ref{fig:celeba-particle-evolution-main} illustrates a similar process on CelebA, where particles undergo gradual changes in appearance, including hair colour, while much of their original facial structure remains recognisable. In both cases, these changes arise from optimizing the particle locations against a fixed estimate of the interventional kernel embedding, with the decoder held fixed throughout. The resulting trajectories illustrate how initializing from observational samples allows the flow to exploit their existing structure when approximating the interventional distribution.\looseness=-1

\medskip
\noindent\textbf{Ablations.}
Appendices~\ref{sec:qualitative-results}
and~\ref{sec:ablation-studies} provide additional
qualitative samples, particle trajectories, and
ablations over sample size, particle count,
initialization, confounding, and Nystr\"om
approximation. Sampling generally improves with
more logged data and consistently with more
particles. Observational initialization is
particularly beneficial on the harder image
benchmarks, while performance degrades only
gradually as overlap weakens.

\vspace{-0.1in}
\section{Conclusion}
\vspace{-0.1in}
We introduced IMMD-Flow, which turns an estimated interventional RKHS target into samples through particle transport, without training an additional sampler. The framework accommodates both supported hard interventions and context-dependent stochastic policies, including continuous-action policies. Its doubly robust, orthogonal target yields product-order perturbations of the transport dynamics. Projected Nystr\"om enables reuse across interventions. On ColorMNIST and CelebA, IMMD-Flow improves pixel-space \(\mathrm{SW}_2\) over the DecFM baselines and reduces total fitting-and-sampling time by \(11\)--\(42\times\) at the reported sampling scale, while achieving comparable FID. A limitation is reliance on standard observed-confounding and overlap assumptions.\looseness=-1

\subsection*{\bbrevise{Code Availability}}

\bbrevise{Our implementation and code for reproducing the experiments are available at \url{https://github.com/BariscanBozkurt/Interventional-MMD-Flow-Sampling}.}

\subsection*{AI use statement}

Generative AI tools were used as editorial and technical aids in preparing this paper. Their use included refining the manuscript's language and structure, supporting code modifications, generating simulation scripts, generating Bash scripts to launch simulations in SLURM cluster, and drafting and checking mathematical concepts and proofs. The authors examined all AI-assisted outputs and independently validated the final text, implementation, and mathematical arguments. The authors take responsibility for the final content of the paper and the accompanying implementation

\newpage
\FloatBarrier
\bibliography{references}
\bibliographystyle{unsrtnat}

\clearpage
\appendix

\startcontents[appendix]

\section*{Appendix Contents}
\printcontents[appendix]{}{1}{\setcounter{tocdepth}{2}}
\bigskip

\newpage

\section{Discussion of Related work}
\label{sec:related_work}

\paragraph{Debiased estimation of interventional distributions.}
A growing literature extends semiparametric debiasing methods from scalar causal estimands to entire interventional or counterfactual outcome distributions. Early work developed efficient estimators for projections and functionals of counterfactual densities, including truncated-series and exponential-family approximations
\citep{kennedy2023semiparametric}. More recent methods combine these ideas
with flexible generative models. Interventional Normalizing Flows
\citet{melnychuk2023normalizing} parameterize the target interventional density
by a normalizing flow and optimize a one-step corrected likelihood objective,
while \citet{martinez2024counterfactual} use a doubly robust Kernel Stein
Discrepancy to estimate an energy-based counterfactual density.
\citet{luedtke2025doublegen} give a more general doubly robust construction
for generative training objectives, with instantiations including diffusion
models and flow matching. \citet{dance2026debiased} learn a flow from the observational
treatment-conditional distribution to the interventional distribution, thereby
exploiting geometric structure shared by the observed and target laws rather
than generating the latter from a fixed generic base distribution. A common limitation of these approaches is that sampling from each interventional distribution of interest requires fitting a corresponding
generative model, in addition to estimating the causal nuisances. This can be especially costly in high-dimensional settings, when multiple interventional distributions are of interest. IMMD-Flow instead avoids target-model training by formulating sampling as distributional optimization. For MMD, the target distribution is represented by a single kernel mean embedding, which admits a closed-form orthogonal estimator from observational
data. Sampling then reduces to directly optimizing a particle approximation
toward this estimated embedding.

\paragraph{Kernel embeddings of interventional distributions.}
Our target representation builds directly on work representing causal outcome
distributions by kernel mean embeddings. \citet{muandet2021counterfactual}
introduced counterfactual mean embeddings (CMEs), which identify
interventional outcome distributions nonparametrically in an RKHS and permit
distributional comparison as well as sample reconstruction by kernel herding.
Subsequent work developed doubly robust versions of these embeddings for
distributional treatment-effect testing
\citep{fawkes2024doubly}, while \citet{luedtke2024one} developed general
one-step theory for Hilbert-valued causal parameters, including
counterfactual kernel mean embeddings. More recently,
\citet{zenati2025cpme} introduced counterfactual policy mean embeddings for
distributional treatment effects under soft interventions, deriving doubly robust estimators for
general target policies together with tests (see also \citep{zenati2026semiparametric}) and a kernel-herding procedure for
generating representative counterfactual samples \citep{simon2023metrizing}. Our method uses the same doubly robust kernel embedding, but as a fixed target
for distributional optimization. Instead of kernel herding, IMMD-Flow jointly
transports particles under the MMD gradient flow, allowing initialization from
a useful source distribution such as $\mathbb P_{Y\mid A=a}$, while propagating
embedding-estimation error directly through the sampling dynamics.

\paragraph{MMD-based particle transport.}
Methodologically, our sampling stage is related to nonparametric generative
methods based on kernel discrepancies and Wasserstein gradient flows.
MMD provides a metric on probability measures for characteristic kernels
\citep{gretton2012kernel,sriperumbudur2010hilbert}, and
\citet{arbel2019mmdflow} introduced the Wasserstein gradient flow of MMD
together with interacting-particle approximations and noise regularization.
Subsequent work has studied alternative kernels, regularization schemes, and
scalable or deep versions of MMD transport
\citep{hertrich2024generative,galashov2024deep,chen2025stationary}.
Recent analyses have further clarified convergence properties of MMD and
energy-kernel gradient flows. These methods assume that the target distribution, or samples from it, are
available. Our setting introduces an additional causal-estimation layer:
samples from the target interventional law are unobserved, and the force
driving the particle dynamics is instead constructed from an estimated
interventional mean embedding. The main connection between the two parts of
our method is therefore that causal estimation error appears as a perturbation
of the target potential in the MMD flow. This permits us to combine
orthogonal estimation of the interventional distribution with particle-level
transport guarantees in a single error analysis.

\paragraph{Counterfactual sampling.}
Our work is also related to the broader literature on sampling from
counterfactual distributions, including structural causal models and
counterfactual transport methods
\citep{pawlowski2020deep,sanchez2022diffusion,javaloy2023causal,dance2025cocycles}.
A key distinction is that IMMD-Flow targets the \emph{marginal interventional law}
$\mathbb P^\pi$, rather than \emph{unit-level counterfactual distributions} such as
$\mathbb P^\pi(\cdot\mid Y=y)$. Consequently, although particles may be
initialized from $\mathbb P_{Y\mid A=a}$ and transported to $\mathbb P^\pi$,
the resulting particle trajectories do not define unit-level counterfactual
couplings. Such an interpretation requires additional cross-world or
structural assumptions beyond those needed to identify $\mathbb P^\pi$.

\section{Orthogonal Target Computation}
\label{sec:orthogonal-target-computation}

This appendix derives the orthogonal Hilbert-space target used by IMMD-Flow. We first give the full kernel-ridge computation, then derive its input-projected Nystr\"om approximation. We subsequently describe the Monte Carlo approximation of the intervention integrals and compare the computational costs of the two constructions. Throughout, the target-policy integral is treated separately from its finite-\(J\) approximation.

\subsection{Full-ridge computation}
\label{sec:full-ridge-computation}

Write \(S_i=(A_i,X_i)\) and equip \(\gA\times\gX\) with the product kernel
\[
    k_S\bigl((a,x),(a',x')\bigr)
    =
    k_A(a,a')k_X(x,x').
\]
Let \(K\in\mathbb R^{n\times n}\) be the corresponding Gram matrix,
\(K_{ij}=k_S(S_i,S_j)\), and define
\[
    B_\lambda
    :=
    (K+n\lambda I_n)^{-1}.
\]
For \(s=(a,x)\), let
\[
    k_s
    =
    \bigl(
        k_S(S_1,s),\ldots,k_S(S_n,s)
    \bigr)^\top.
\]
The vector-valued kernel-ridge estimator of the conditional mean embedding
is
\begin{equation}
    \widehat m_{\rm full}(s)
    =
    \sum_{j=1}^n
    [B_\lambda k_s]_j\,\varphi(Z_j).
    \label{eq:full-cme}
\end{equation}

For each observed context \(X_i\), define the intervention-integrated kernel
vector
\begin{equation}
    \bar k_{\pi,i}
    :=
    \int_{\gA}
    k_{(a,X_i)}\,
    \pi(da\mid X_i)
    \in\mathbb R^n,
    \label{eq:full-policy-kernel}
\end{equation}
and collect these vectors as
\[
    K_\pi
    :=
    \bigl(
        \bar k_{\pi,1},\ldots,\bar k_{\pi,n}
    \bigr)
    \in\mathbb R^{n\times n}.
\]

For the product kernel above, this matrix can equivalently be written as
\[
    K_\pi
    =
    K_X\odot K_{A,\pi},
\]
where
\begin{equation}
    [K_{A,\pi}]_{ji}
    :=
    \int_{\gA}
    k_A(A_j,a)\,
    \pi(da\mid X_i).
    \label{eq:exact-policy-action-kernel}
\end{equation}

Writing
\[
    \widehat{\boldsymbol w}
    =
    \bigl(
        \widehat w(A_1,X_1),\ldots,
        \widehat w(A_n,X_n)
    \bigr)^\top,
\]
substitution of \eqref{eq:full-cme} into the empirical orthogonal score
gives
\begin{equation}
    \widehat h_{\pi,\rm full}
    =
    \sum_{i=1}^n
    v_{{\rm full},i}\varphi(Z_i),
    \label{eq:full-dr-expansion}
\end{equation}
where
\begin{align}
    b^{\rm full}
    &=
    B_\lambda
    \left(
        K_\pi\mathbf 1_n
        -
        K\widehat{\boldsymbol w}
    \right), \enspace
    v_{\rm full}
    =
    \frac1n
    \left(
        b^{\rm full}
        +
        \widehat{\boldsymbol w}
    \right).
    \label{eq:full-dr-coefficients}
\end{align}

Equivalently, defining the full-ridge plug-in coefficients
\begin{equation}
    v_{\mathrm{PI,full}}
    :=
    B_\lambda K_\pi\frac{\mathbf 1_n}{n},
    \label{eq:full-plugin-coefficients}
\end{equation}
the doubly robust coefficients satisfy
\begin{equation}
    v_{\rm full}
    =
    v_{\mathrm{PI,full}}
    +
    \frac1n
    \left(
        \widehat{\boldsymbol w}
        -
        B_\lambda K\widehat{\boldsymbol w}
    \right).
    \label{eq:full-plugin-correction-form}
\end{equation}
This is the finite-sample full-kernel form of the doubly robust policy mean embedding of \citet{zenati2025cpme}.

Thus, once the regression system is solved, changing the intervention affects the estimator only through the intervention integrals and, when target-specific, the density-ratio nuisance.

\subsection{Projected Nystr\"om computation}
\label{sec:projected_nystrom_computation}

The full construction requires solving an \(n\times n\) kernel system. We instead restrict the conditional mean embedding regression to a Nystr\"om subspace of the input RKHS \citep{rudi2015less,pmlr-v238-el-ahmad24a}.

Select landmark indices
\[
    L
    =
    \{\ell_1,\ldots,\ell_{m_L}\}
    \subseteq
    \{1,\ldots,n\}
\]
uniformly without replacement, and write
\[
    \widetilde S_q
    :=
    S_{\ell_q}.
\]
Define
\begin{align}
    C_{iq}
    &:=
    k_S(S_i,\widetilde S_q),
    &
    W_{qr}
    &:=
    k_S(\widetilde S_q,\widetilde S_r),
    \label{eq:nystrom-matrices}
\end{align}
so that
\[
    C\in\mathbb R^{n\times m_L},
    \qquad
    W\in\mathbb R^{m_L\times m_L}.
\]
Let
\begin{equation}
    G_\lambda
    :=
    \left(
        C^\top C+n\lambda W
    \right)^\dagger,
    \label{eq:nystrom-system}
\end{equation}
where \(\dagger\) denotes the Moore--Penrose inverse when required.

For \(s=(a,x)\), define the landmark kernel vector
\[
    c_s
    =
    \bigl(
        k_S(\widetilde S_1,s),\ldots,
        k_S(\widetilde S_{m_L},s)
    \bigr)^\top.
\]
The projected estimator can then be written directly in the logged-outcome
basis as
\begin{equation}
    \widehat m_{\rm Nys}(s)
    =
    \sum_{i=1}^n
    [CG_\lambda c_s]_i\,
    \varphi(Z_i).
    \label{eq:projected-nystrom-cme}
\end{equation}

For each context \(X_i\), define
\begin{equation}
    \bar c_{\pi,i}
    :=
    \int_{\gA}
    c_{(a,X_i)}\,
    \pi(da\mid X_i)
    \in\mathbb R^{m_L},
    \label{eq:nystrom-policy-integral}
\end{equation}
and let
\[
    C_\pi
    :=
    \bigl(
        \bar c_{\pi,1},\ldots,\bar c_{\pi,n}
    \bigr)
    \in\mathbb R^{m_L\times n}.
\]

Because the landmarks are selected from the logged inputs,
\[
    C_\pi
    =
    (K_\pi)_{L:}.
\]
Thus the projected construction evaluates the exact intervention-integrated regression contrast only at the landmarks.

Substituting \eqref{eq:projected-nystrom-cme} into the orthogonal score
gives
\begin{align}
    b^{\rm proj}
    &=
    CG_\lambda
    \left(
        C_\pi\mathbf 1_n
        -
        C^\top\widehat{\boldsymbol w}
    \right),
     \enspace v_{\rm DR}
    =
    \frac1n
    \left(
        b^{\rm proj}
        +
        \widehat{\boldsymbol w}
    \right).
    \label{eq:projected-dr-coefficients}
\end{align}
Hence the exact-policy-integral projected target is
\begin{equation}
    \widehat h_{\pi,\rm Nys}
    =
    \sum_{i=1}^n
    v_{{\rm DR},i}\varphi(Z_i).
    \label{eq:projected-dr-target}
\end{equation}

The projection acts on the input-side regression problem only: \eqref{eq:projected-dr-target} still expands over all \(n\) logged outcome features. This distinction is relevant to the particle-stage complexity.

\paragraph{Whitened landmark coordinates.}
For numerical stability, let
\[
    W
    =
    V_rD_rV_r^\top
\]
denote the eigendecomposition of \(W\) on its retained positive eigenspace, and define
\[
    R_W
    :=
    V_rD_r^{-1/2},
    \qquad
    \Phi
    :=
    CR_W.
\]
Writing
\begin{equation}
    r_\pi
    :=
    C_\pi\mathbf 1_n
    -
    C^\top\widehat{\boldsymbol w},
    \qquad
    q_\pi
    :=
    R_W^\top r_\pi,
    \label{eq:nystrom-whitened-rhs}
\end{equation}
the projected coefficients can equivalently be computed as
\begin{equation}
    b^{\rm proj}
    =
    \Phi
    \left(
        \Phi^\top\Phi+n\lambda I_r
    \right)^{-1}
    q_\pi.
    \label{eq:nystrom-whitened-computation}
\end{equation}
Using all positive eigenpairs is only a change of coordinates within the landmark subspace. Discarding additional positive eigenvalues through a numerical threshold introduces a further subspace approximation.

\subsection{Intervention integration and computational reuse}
\label{sec:policy-integral-computation}

When the intervention integrals in \eqref{eq:full-policy-kernel} and \eqref{eq:nystrom-policy-integral} are unavailable analytically, we draw
\[
    A_{i1}^{\pi},\ldots,A_{iJ}^{\pi}
    \overset{\mathrm{iid}}{\sim}
    \pi(\cdot\mid X_i)
\]
for every logged context \(X_i\).

For the full-ridge construction, define
\begin{equation}
    \widetilde k_{\pi,i}^{(J)}
    :=
    \frac1J
    \sum_{j=1}^J
    k_{(A_{ij}^{\pi},X_i)}
    \in\mathbb R^n,
    \label{eq:mc-full-policy-integral}
\end{equation}
and collect these vectors as
\(
    \widetilde K_\pi^{(J)}
    :=
    \bigl(
        \widetilde k_{\pi,1}^{(J)},\ldots,
        \widetilde k_{\pi,n}^{(J)}
    \bigr).
\)
The corresponding finite-\(J\) full-ridge coefficients are
\begin{align}
    \widetilde b_J^{\rm full}
    &:=
    B_\lambda
    \left(
        \widetilde K_\pi^{(J)}\mathbf 1_n
        -
        K\widehat{\boldsymbol w}
    \right), \enspace
    \widetilde v_{\mathrm{full},J}
    :=
    \frac1n
    \left(
        \widetilde b_J^{\rm full}
        +
        \widehat{\boldsymbol w}
    \right),
    \label{eq:finite-J-full-coefficients}
\end{align}
with numerical target
\begin{equation}
    \widetilde h_{\pi,\mathrm{full},J}
    :=
    \sum_{i=1}^n
    \widetilde v_{\mathrm{full},J,i}
    \varphi(Z_i).
    \label{eq:finite-J-full-target}
\end{equation}

For the projected construction, use
\begin{equation}
    \widetilde c_{\pi,i}^{(J)}
    :=
    \frac1J
    \sum_{j=1}^J
    c_{(A_{ij}^{\pi},X_i)}.
    \label{eq:mc-policy-integral}
\end{equation}
Let
\[
    \widetilde C_\pi^{(J)}
    :=
    \bigl(
        \widetilde c_{\pi,1}^{(J)},\ldots,
        \widetilde c_{\pi,n}^{(J)}
    \bigr).
\]

Using the same target-policy draws in the full and projected constructions,
\[
    \widetilde C_\pi^{(J)}
    =
    \bigl(
        \widetilde K_\pi^{(J)}
    \bigr)_{L:}.
\]
Replacing \(C_\pi\) by \(\widetilde C_\pi^{(J)}\) in
\eqref{eq:projected-dr-coefficients} gives
\begin{align}
    \widetilde b_J^{\rm proj}
    &:=
    CG_\lambda
    \left(
        \widetilde C_\pi^{(J)}\mathbf 1_n
        -
        C^\top\widehat{\boldsymbol w}
    \right),\enspace
    \widetilde v_{\mathrm{DR},J}
    :=
    \frac1n
    \left(
        \widetilde b_J^{\rm proj}
        +
        \widehat{\boldsymbol w}
    \right).
    \label{eq:finite-J-projected-coefficients}
\end{align}
The resulting numerical target is
\begin{equation}
    \widetilde h_{\pi,\mathrm{Nys},J}
    :=
    \sum_{i=1}^n
    \widetilde v_{\mathrm{DR},J,i}
    \varphi(Z_i).
    \label{eq:numerical-target-appendix}
\end{equation}
This is the finite-\(J\) target denoted by
\(\widetilde h_\pi\) in the main text:
\[
    \widetilde h_\pi
    =
    \widetilde h_{\pi,\mathrm{Nys},J}.
\]
Similarly,
\(
    \widehat h_\pi
    =
    \widehat h_{\pi,\mathrm{Nys}}
\)
denotes the projected estimator with exact intervention integration.

Let \(\mathcal F_n\) contain the logged sample, selected landmarks, and fitted nuisance estimates, prior to drawing the target-policy actions. Conditional on \(\mathcal F_n\),
\begin{align}
    \E\!\left[
        \widetilde K_\pi^{(J)}
        \mid
        \mathcal F_n
    \right]
    &=
    K_\pi,
    &
    \E\!\left[
        \widetilde C_\pi^{(J)}
        \mid
        \mathcal F_n
    \right]
    &=
    C_\pi.
    \label{eq:policy-integral-matrix-unbiasedness}
\end{align}
Because both coefficient maps are linear in the intervention-integrated right-hand side,
\begin{align}
    \E\!\left[
        \widetilde h_{\pi,\mathrm{full},J}
        \mid
        \mathcal F_n
    \right]
    &=
    \widehat h_{\pi,\mathrm{full}}, &
    \E\!\left[
        \widetilde h_{\pi,\mathrm{Nys},J}
        \mid
        \mathcal F_n
    \right]
    &=
    \widehat h_{\pi,\mathrm{Nys}}.
    \label{eq:finite-J-target-unbiasedness-matrix}
\end{align}
The corresponding Hilbert-space Monte Carlo error is analyzed in Appendix~\ref{sec:theoretical-analysis}.

For fixed representation, kernels, and adjustment variables, the landmark set, matrices \(C,W\), and factorization of \(C^\top C+n\lambda W\) are shared across interventions. Intervention integrals and intervention-specific density ratios are recomputed as needed.

Thus, evaluating a new supported intervention does not require refitting the conditional mean embedding. It requires only the intervention-specific policy integrals, density-ratio evaluations, and the resulting linear-system right-hand side.

\subsection{Plug-in and inverse-probability variants}
\label{app:PlugIn_IPW_IMMD}

{ 
Our benchmark tables also compare two single-nuisance IMMD variants. For the
finite-\(J\) projected construction, the plug-in, inverse-probability-weighted
(IPW), and doubly robust coefficient vectors are
\begin{align*}
    \widetilde v_{\mathrm{PI},J}
    &:=
    \frac1n CG_\lambda\widetilde C_\pi^{(J)}\mathbf 1_n,
    &
    \widetilde v_{\mathrm{IPW}}
    &:=
    \frac1n\widehat{\boldsymbol w},
    \\
    \widetilde v_{\mathrm{DR},J}
    &:=
    \widetilde v_{\mathrm{PI},J}
    +
    \frac1n
    \left(
        \widehat{\boldsymbol w}
        -
        CG_\lambda C^\top\widehat{\boldsymbol w}
    \right).
\end{align*}
Thus, Plug-in--IMMD uses only the fitted conditional mean embedding,
IPW--IMMD uses only the estimated policy ratio, and DR--IMMD combines both.
The displayed IPW vector is the Horvitz--Thompson form; any declared
benchmark-specific unit-mass rescaling is applied after these raw coefficients
are constructed. For
\(e\in\{\mathrm{PI},\mathrm{IPW},\mathrm{DR}\}\), let
\(\widetilde v_e\) denote the corresponding vector above, with its
\(J\) subscript suppressed. We freeze
\(\widetilde h_{\pi,e}=\sum_{i=1}^n\widetilde v_{e,i}\varphi(Z_i)\) and
apply the same particle flow. Hence the variants differ only in the target
coefficients supplied to the optimizer.
}

\subsection{Computational complexity}
\label{app:ComputationalComplexity}

We compare the time and memory costs of the full-ridge and input-projected Nystr\"om target constructions. Table~\ref{tab:target-computation-complexity} summarizes the comparison, while the discussion below details the dominant operations and the computations that can be reused across interventions.

\begin{table}[t]
\centering
\caption{
Computational comparison of the full-ridge and input-projected
Nystr\"om target constructions. The per-intervention costs assume that
the corresponding regression system has already been factorized.
}
\label{tab:target-computation-complexity}
\setlength{\tabcolsep}{8pt}
\renewcommand{\arraystretch}{1.20}
\begin{tabular*}{0.94\textwidth}{
    @{\extracolsep{\fill}}
    lcc
    @{}
}
\toprule
& \textbf{Full ridge} & \textbf{Nystr\"om} \\
\midrule
Kernel storage
& \(O(n^2)\)
& \(O(nm_L+m_L^2)\)
\\
One-time regression solve
& \(O(n^3)\)
& \(O(nm_L^2+m_L^3)\)
\\
Finite-\(J\) target per intervention
& \(O((J+1)n^2)\)
& \(O((J+1)nm_L+m_L^2)\)
\\
Target force per particle iteration
& \(O(Mn)\)
& \(O(Mn)\)
\\
\bottomrule
\end{tabular*}
\end{table}

For the full-ridge estimator, constructing and factorizing the \(n\times n\) Gram system costs \(O(n^3)\) time and \(O(n^2)\) memory. When intervention integrals are approximated with \(J\) draws per context, forming the intervention-integrated right-hand side costs \(O(Jn^2)\); the remaining matrix--vector operations cost \(O(n^2)\).

For the Nystr\"om estimator, forming \(C^\top C\) costs \(O(nm_L^2)\), factorizing the resulting \(m_L\times m_L\) system costs \(O(m_L^3)\), and storing \(C\) and \(W\) requires \(O(nm_L+m_L^2)\) memory. Finite-\(J\) intervention integration costs \(O(Jm_Ln)\), while constructing the remaining intervention-specific right-hand side and coefficients costs \(O(nm_L+m_L^2)\). For fixed
\(J\), the total target-construction cost is therefore
\(
    O(nm_L^2+m_L^3)
\)
time and
\(
    O(nm_L+m_L^2)
\)
memory, assuming lazy or blockwise kernel evaluation. Precomputing the full Gram matrix would restore quadratic storage.

Input-side Nystr\"om approximation reduces the cost of conditional-embedding regression and target construction, but it does not reduce the number of logged outcome features in the final target expansion. Consequently, direct evaluation of the target contribution for \(M\) particles still costs \(O(Mn)\) kernel gradients per particle iteration. Reducing this term would require an additional output-side approximation.

If the intervention integrals are available analytically, the factors proportional to \(J\) disappear. Moreover, the regression factorizations are reusable across supported interventions, so subsequent targets require only the intervention-specific integration and density-ratio computations.

The statistical guarantees in Appendix~\ref{sec:theoretical-analysis} require the landmark count \(m_L\) to increase with \(n\) according to the leverage-based schedule stated there. A fixed landmark budget is therefore a computational approximation, rather than a general asymptotic consistency guarantee.

\section{Proofs for the Interventional MMD Construction}
\label{sec:ProofsForIMMDConstruction}

\subsection{Double robustness and orthogonality}
\label{app:proof-orthogonal-interventional-target}

\begin{proof}[Proof of Proposition~\ref{prop:orthogonal-interventional-target}]
By the change-of-measure identity,
\begin{equation}
    \E_0\!\left[
        \int_{\gA}
        m(a,X)\,
        \pi(da\mid X)
    \right]
    =
    \E_0\!\left[
        w_0(A,X)m(A,X)
    \right].
    \label{eq:proof-policy-change-of-measure}
\end{equation}
Moreover,
\(
    \E_0[
        \varphi(Z)
        \mid
        A,X
    ]
    =
    m_0(A,X).
\)
It follows that
\begin{align*}
    h_\eta
    &=
    \E_0\!\left[
        w_0m+w(m_0-m)
    \right].
\end{align*}
Since
\(
    \chi(\pi)
    =
    \E_0[w_0m_0],
\)
we obtain
\begin{align*}
    h_\eta-\chi(\pi)
    &=
    \E_0\!\left[
        w_0m+w(m_0-m)-w_0m_0
    \right]
    \\
    &=
    -\E_0\!\left[
        (w-w_0)(m-m_0)
    \right].
\end{align*}
Taking norms and applying the Hilbert-valued Cauchy--Schwarz inequality gives
\begin{align*}
    \left\|
        h_\eta-\chi(\pi)
    \right\|_{\gH}
    &\leq
    \left\|
        w-w_0
    \right\|_{L_2(P_{A,X})}
    \left\|
        m-m_0
    \right\|_{L_2(P_{A,X};\gH)}
    =
    r_wr_m.
\end{align*}
The double-robustness statements follow immediately.

For a direction
\(\dot\eta=(\dot m,\dot w)\), consider the path
\[
    \eta_\epsilon
    =
    \eta_0+\epsilon\dot\eta
    =
    \left(
        m_0+\epsilon\dot m,
        w_0+\epsilon\dot w
    \right).
\]
The remainder identity gives
\[
    h_{\eta_\epsilon}
    -
    \chi(\pi)
    =
    -\epsilon^2
    \E_0[
        \dot w\,\dot m
    ].
\]
Hence
\[
    D_\eta h_{\eta_0}[\dot\eta]
    =
    0.
\]
Since \(u\) is fixed with respect to \(\eta\),
\[
    D_\eta
    \left(
        u-h_\eta
    \right)
    \big|_{\eta=\eta_0}
    [\dot\eta]
    =
    -D_\eta h_{\eta_0}[\dot\eta]
    =
    0,
\]
which proves \eqref{eq:universal-orthogonality}.
\end{proof}

\subsection{Dissipation and velocity stability}
\label{app:proof-immd-dissipation}

\begin{proof}[Proof of Proposition~\ref{prop:immd-stability}]
We first record the spatial gradient of the witness. For fixed \(Q\in\mathcal P_2(\gZ)\), \(h\in\gH\), and \(u\in\mathbb R^{d_Z}\), Fr\'echet differentiability of \(\varphi\) gives
\begin{align*}
    D f_{Q,h}(z)[u]
    &=
    \left\langle
        \mu_Q-h,
        D\varphi(z)[u]
    \right\rangle_{\gH}
    =
    \left\langle
        D\varphi(z)^*(\mu_Q-h),
        u
    \right\rangle_2.
\end{align*}
Consequently,
\begin{equation}
    \nabla f_{Q,h}(z)
    =
    D\varphi(z)^*(\mu_Q-h),
    \qquad
    v_h(Q,z)
    =
    -D\varphi(z)^*(\mu_Q-h).
    \label{eq:proof-witness-gradient}
\end{equation}

Let
\[
    \mu_t
    :=
    \mu_{Q_t^h}
    =
    \int_{\gZ}
    \varphi(z)\,Q_t^h(dz),
    \qquad
    v_t(z)
    :=
    v_h(Q_t^h,z).
\]
To identify the derivative of \(\mu_t\), fix \(g\in\gH\) and define
the scalar test function
\[
    \psi_g(z)
    :=
    \langle g,\varphi(z)\rangle_{\gH}.
\]
Its spatial gradient is
\[
    \nabla\psi_g(z)
    =
    D\varphi(z)^*g.
\]
Applying the weak formulation of the continuity equation to
\(\psi_g\) yields, for almost every \(t\),
\begin{align*}
    \frac{d}{dt}
    \langle g,\mu_t\rangle_{\gH}
    &=
    \frac{d}{dt}
    \int_{\gZ}
    \psi_g(z)\,Q_t^h(dz)
    \\
    &=
    \int_{\gZ}
    \left\langle
        D\varphi(z)^*g,
        v_t(z)
    \right\rangle_2
    Q_t^h(dz)
    \\
    &=
    \left\langle
        g,
        \int_{\gZ}
        D\varphi(z)[v_t(z)]
        \,Q_t^h(dz)
    \right\rangle_{\gH}.
\end{align*}
Since this holds for every \(g\in\gH\),
\begin{equation}
    \dot\mu_t
    =
    \int_{\gZ}
    D\varphi(z)[v_t(z)]
    \,Q_t^h(dz)
    \label{eq:mean-embedding-evolution}
\end{equation}
for almost every \(t\). The Bochner integral is well defined because
\(D\varphi\) is uniformly bounded and
\(v_t\in L_2(Q_t^h;\mathbb R^{d_Z})\).

Using the Hilbert-space chain rule and
\eqref{eq:mean-embedding-evolution},
\begin{align*}
    \frac{d}{dt}
    \mathcal F_h(Q_t^h)
    &=
    \frac{d}{dt}
    \frac12
    \left\|
        \mu_t-h
    \right\|_{\gH}^2
    \\
    &=
    \left\langle
        \mu_t-h,
        \dot\mu_t
    \right\rangle_{\gH}
    \\
    &=
    \int_{\gZ}
    \left\langle
        D\varphi(z)^*(\mu_t-h),
        v_t(z)
    \right\rangle_2
    Q_t^h(dz).
\end{align*}
By \eqref{eq:proof-witness-gradient},
\[
    D\varphi(z)^*(\mu_t-h)
    =
    -v_t(z).
\]
Therefore,
\begin{equation}
    \frac{d}{dt}
    \mathcal F_h(Q_t^h)
    =
    -
    \int_{\gZ}
    \left\|
        v_h(Q_t^h,z)
    \right\|_2^2
    Q_t^h(dz)
    \leq0,
\end{equation}
which proves the dissipation identity.

For velocity stability, let \(h_1,h_2\in\gH\). From
\eqref{eq:proof-witness-gradient},
\begin{align*}
    v_{h_1}(Q,z)-v_{h_2}(Q,z)
    &=
    D\varphi(z)^*(h_1-h_2).
\end{align*}
Hence
\begin{align*}
    \left\|
        v_{h_1}(Q,z)-v_{h_2}(Q,z)
    \right\|_2
    &\leq
    \left\|
        D\varphi(z)
    \right\|_{\mathrm{op}}
    \left\|
        h_1-h_2
    \right\|_{\gH}
    \\
    &\leq
    L_\varphi
    \left\|
        h_1-h_2
    \right\|_{\gH}.
\end{align*}
Taking the supremum over \(z\in\gZ\), and then setting
\[
    h_1=h_\eta,
    \qquad
    h_2=\chi(\pi),
\]
gives
\begin{align*}
    \sup_{z\in\gZ}
    \left\|
        v_{h_\eta}(Q,z)
        -
        v_{\chi(\pi)}(Q,z)
    \right\|_2
    &\leq
    L_\varphi
    \left\|
        h_\eta-\chi(\pi)
    \right\|_{\gH}
    \\
    &\leq
    L_\varphi r_wr_m,
\end{align*}
where the final inequality follows from Proposition~\ref{prop:orthogonal-interventional-target}.
\end{proof}

\section{Theoretical Analysis}
\label{sec:theoretical-analysis}

\subsection{Projected Nystr\"om approximation of the conditional mean embedding}

This section controls the error introduced when the conditional mean embedding nuisance is estimated by input-projected Nystr\"om kernel ridge regression. The argument specializes the vector-valued regression analysis of \citet{pmlr-v238-el-ahmad24a} to our setting. In contrast to their general construction, we sketch only the input RKHS. The latent-output RKHS remains unchanged, so no output-projection approximation term appears.

\noindent\textbf{Statistical setup.}
Let
\(
    \mathcal S:=\mathcal A\times\mathcal X,
    \enspace
    S:=(A,X),
    \enspace
    Z:=E(Y),
\)
where the encoder \(E\) is fixed and trained independently of the observations used in this section. Let
\(
    (S_i,Z_i)_{i=1}^n
\)
be i.i.d. observations from the logged joint law \(P_0\).

Let \(k_S:\mathcal S\times\mathcal S\to\mathbb R\) be a measurable positive-definite input kernel with RKHS \(\mathcal H_S\) and canonical feature map
\[
    \varphi_S(s):=k_S(s,\cdot).
\]
Let \(k_Z:\mathcal Z\times\mathcal Z\to\mathbb R\) be the
positive-definite latent kernel used to define the interventional embedding, with RKHS \(\mathcal H_Z\) and feature map
\[
    \varphi_Z(z):=k_Z(z,\cdot).
\]

\begin{assumption}[Measurability and kernel regularity]
\label{ass:nystrom-measurability}
We assume that
\begin{itemize}
    \item The context, action, and latent spaces \(\mathcal X\), \(\mathcal A\), and \(\mathcal Z\) are Polish spaces, equipped with their Borel \(\sigma\)-fields.

    \item The RKHSs \(\mathcal H_S\) and \(\mathcal H_Z\) are separable,
    their feature maps are strongly measurable, and the fitted functions,
    operators, and displayed norms are measurable.

    \item There exist finite constants \(\kappa_S,\kappa_Z>0\) such that
\[
    \sup_{s\in\mathcal S} k_S(s,s)\le \kappa_S^2,
    \qquad
    \sup_{z\in\mathcal Z} k_Z(z,z)\le \kappa_Z^2.
\]
\end{itemize}

\end{assumption}

Define the latent conditional mean embedding
\[
    m_0(s)
    :=
    \mathbb E_{P_0}\!\left[
        \varphi_Z(Z)\mid S=s
    \right]
    \in\mathcal H_Z.
\]
We use the identity-decomposable operator-valued kernel (OVK)
\[
    \mathbf K(s,s')
    =
    k_S(s,s')I_{\mathcal H_Z}.
\]
Its vector-valued RKHS is isometrically identified with the space \(\mathcal S_2(\mathcal H_S,\mathcal H_Z)\) of Hilbert--Schmidt operators: an operator \(H:\mathcal H_S\to\mathcal H_Z\) represents the function
\[
    m_H(s)=H\varphi_S(s),
    \qquad
    \|m_H\|_{\mathcal G}=\|H\|_{\mathrm{HS}}.
\]

\begin{assumption}[Attainability; {\citet[Assumption~1]{pmlr-v238-el-ahmad24a}}]
\label{ass:nystrom-attainability}
There exists \(H_0\in\mathcal S_2(\mathcal H_S,\mathcal H_Z)\) such that
\[
    m_0(s)=H_0\varphi_S(s)
    \qquad
    P_{S,0}\text{-almost surely}.
\]
\end{assumption}

We now define the covariance operators associated with conditional mean embedding learning.

\begin{definition}[Population covariance operators]
\label{def:population-covariance-operators}

Let \(P_{S,0}\) denote the marginal distribution of
\(S=(A,X)\) under the logging law. We define the population input
covariance operator
\[
    C_S
    :=
    \mathbb E_{P_0}
    \left[
        \varphi_S(S)\otimes\varphi_S(S)
    \right]
    :
    \mathcal H_S\to\mathcal H_S,
\]
and the population output--input cross-covariance operator
\[
    C_{ZS}
    :=
    \mathbb E_{P_0}
    \left[
        \varphi_Z(Z)\otimes\varphi_S(S)
    \right]
    :
    \mathcal H_S\to\mathcal H_Z.
\]
\end{definition}

Under Assumption~\ref{ass:nystrom-measurability}, these Bochner expectations are well defined. The operator \(C_S\) is positive, self-adjoint, and trace class, with
\[
    \operatorname{Tr}(C_S)
    =
    \mathbb E\|\varphi_S(S)\|_{\mathcal H_S}^2
    \leq \kappa_S^2.
\]
Moreover, \(C_{ZS}\) is Hilbert--Schmidt, since
\[
    \|\varphi_Z(Z)\otimes\varphi_S(S)\|_{\mathrm{HS}}
    =
    \|\varphi_Z(Z)\|_{\mathcal H_Z}
    \|\varphi_S(S)\|_{\mathcal H_S}
    \leq \kappa_Z\kappa_S.
\]

\begin{lemma}[Consequences of attainability]
\label{lem:attainability-consequences}
Under Assumption~\ref{ass:nystrom-attainability},
\[
    C_{ZS}=H_0C_S.
\]
Moreover, for every
\(H\in\mathcal S_2(\mathcal H_S,\mathcal H_Z)\),
\[
    \|m_H-m_0\|_{L_2(P_{S,0};\mathcal H_Z)}
    =
    \|(H-H_0)C_S^{1/2}\|_{\mathrm{HS}}.
\]
\end{lemma}

{
\begin{proof}
For every \(f\in\mathcal H_S\), conditional expectation gives
\begin{align*}
    C_{ZS}f
    &=
    \mathbb E\left[
        \varphi_Z(Z)
        \langle\varphi_S(S),f\rangle_{\mathcal H_S}
    \right]
    =
    \mathbb E\left[
        m_0(S)
        \langle\varphi_S(S),f\rangle_{\mathcal H_S}
    \right]
    =
    H_0
    \mathbb E\left[
        \varphi_S(S)
        \langle\varphi_S(S),f\rangle_{\mathcal H_S}
    \right]
    =
    H_0C_Sf.
\end{align*}
This proves \(C_{ZS}=H_0C_S\). For the second statement,
\begin{align*}
    \|m_H-m_0\|_{L_2(P_{S,0};\mathcal H_Z)}^2
    &=
    \mathbb E
    \left[
        \|(H-H_0)\varphi_S(S)\|_{\mathcal H_Z}^2
    \right]
    =
    \operatorname{Tr}
    \left[
        (H-H_0)C_S(H-H_0)^*
    \right]
    =
    \|(H-H_0)C_S^{1/2}\|_{\mathrm{HS}}^2.
\end{align*}
Taking nonnegative square roots completes the proof.
\end{proof}
}

The next assumption characterizes the eigendecay of the input covariance operator and the maximal leverage of the input features. These quantities determine how many uniformly sampled landmarks are sufficient for the projected Nystr\"om estimator to retain the statistical accuracy of full kernel ridge regression.

\begin{assumption}[Input spectral regularity; {\citet[Assumption~3-4]{pmlr-v238-el-ahmad24a}}]
\label{ass:nystrom-input-regularity}
For some \(\gamma\in(0,1]\),
\[
    Q_S:=\operatorname{Tr}(C_S^\gamma)<\infty.
\]
Moreover, for some \(\mu\in[0,1)\) and \(B_S<\infty\), we have 
\(
    \varphi_S(s)\otimes\varphi_S(s)
    \preceq
    B_SC_S^{1-\mu}
    \enspace
    P_{S,0}\text{-almost surely}.
\)
\end{assumption}

\begin{definition}[Empirical sampling and covariance operators]
\label{def:empirical-operators}
Given observations
\(
    (S_i,Z_i)_{i=1}^n,
\)
define the normalized sampling operators
\begin{equation}
\begin{aligned}
    \mathsf T_S &: \mathcal H_S \to \mathbb R^n,
    &
    (\mathsf T_S f)_i
    &:=
    \frac{1}{\sqrt n} f(S_i),\quad
    \mathsf T_Z : \mathcal H_Z \to \mathbb R^n,
    &(\mathsf T_Z g)_i
    :=
    \frac{1}{\sqrt n} g(Z_i).
\end{aligned}
\label{eq:nystrom-sampling-operators}
\end{equation}
With \(\mathbb R^n\) equipped with its standard Euclidean inner product,
their adjoints are
\begin{equation}
\begin{aligned}
    \mathsf T_S^* a
    &=
    \frac{1}{\sqrt n}
    \sum_{i=1}^n
    a_i\varphi_S(S_i), \quad
    \mathsf T_Z^* a
    =
    \frac{1}{\sqrt n}
    \sum_{i=1}^n
    a_i\varphi_Z(Z_i),
    \quad
    a\in\mathbb R^n.
\end{aligned}
\label{eq:nystrom-sampling-adjoints}
\end{equation}

The empirical input covariance and output--input cross-covariance
operators are respectively
\begin{equation}
\begin{aligned}
    \widehat C_S
    &:=
    \mathsf T_S^*\mathsf T_S
    =
    \frac{1}{n}
    \sum_{i=1}^n
    \varphi_S(S_i)\otimes\varphi_S(S_i),
    \quad
    \widehat C_{ZS}
    :=
    \mathsf T_Z^*\mathsf T_S
    =
    \frac{1}{n}
    \sum_{i=1}^n
    \varphi_Z(Z_i)\otimes\varphi_S(S_i).
\end{aligned}
\label{eq:nystrom-empirical-covariances}
\end{equation}
\end{definition}

\noindent\textbf{Dense Hilbert-valued ridge regression.} By Definition~\ref{def:empirical-operators}, for every
\(H\in\mathcal S_2(\mathcal H_S,\mathcal H_Z)\),
\begin{align}
\norm{\mathsf T_Z^*-H\mathsf T_S^*}_{\mathrm{HS}}^2
&=
\frac1n\sum_{i=1}^n
\norm{
    \varphi_Z(Z_i)-H\varphi_S(S_i)
}_{\mathcal H_Z}^2.
\label{eq:dense-operator-loss-equivalence}
\end{align}
Consequently, for \(\lambda>0\), the dense empirical ridge estimator is
defined by
\begin{align}
\widehat H_\lambda
\in
\operatorname*{arg\,min}_{H\in\mathcal S_2(\mathcal H_S,\mathcal H_Z)}
\left\{
    \norm{\mathsf T_Z^*-H\mathsf T_S^*}_{\mathrm{HS}}^2
    +
    \lambda\norm H_{\mathrm{HS}}^2
\right\}.
\label{eq:nystrom-dense-operator-objective}
\end{align}

Let
\(
    K_S
    :=
    \bigl(k_S(S_i,S_j)\bigr)_{i,j=1}^n
    \in\mathbb R^{n\times n},
\)
and, for \(s\in\mathcal S\), define
\(
    k_S(s)
    :=
    \bigl(k_S(S_i,s)\bigr)_{i=1}^n
    \in\mathbb R^n.
\)
For the product kernel introduced in the algorithmic section,
\(K_S=K_{AX}\).

\begin{proposition}[Dense Hilbert-valued ridge solution; {\citep[Theorem (4.1)]{Micchelli_OnLearningVectorFunctions}}, {\citep[Equation (3)-(4)]{pmlr-v238-el-ahmad24a}}]
\label{prop:dense-operator-solution}
The optimization problem
\eqref{eq:nystrom-dense-operator-objective} admits a unique minimizer,
given by
\begin{align}
\widehat H_\lambda
&=
\widehat C_{ZS}
\left(
    \widehat C_S+\lambda I_{\mathcal H_S}
\right)^{-1}.
\label{eq:nystrom-dense-operator-solution}
\end{align}
Consequently, writing
\(
    \widehat m_\lambda(s)
    :=
    \widehat H_\lambda\varphi_S(s),
\)
we have the equivalent representer formula
\begin{align}
\widehat m_\lambda(s)
&=
\sum_{i=1}^n
\left[
    \left(
        K_S+n\lambda I_n
    \right)^{-1}
    k_S(s)
\right]_i
\varphi_Z(Z_i).
\label{eq:nystrom-dense-representer}
\end{align}
\end{proposition}

\paragraph{Generic input sketch and Nystr\"om specialization.}
Following \citet{pmlr-v238-el-ahmad24a}, let $\mathsf R\in\R^{m\times n}$ be a a possibly random input-sketch matrix. \citet{pmlr-v238-el-ahmad24a} formulate their method for general input and output sketches; here we sketch only the input RKHS and leave the latent-output RKHS unchanged. 

Define the input-sketch operator and its adjoint
\begin{align}
\mathsf B_R
&:=\mathsf R\mathsf T_S:\gH_S\to\R^m, \quad
\mathsf B_R^*
=\mathsf T_S^* \mathsf R^\top:\R^m\to\gH_S,
\label{eq:nystrom-sketch-operator}
\end{align}

The input-sketch operator induces the finite dimensional subspace
\[
\gH_{S,R} \coloneqq \operatorname{Range}(\mathsf B_R^*) \subseteq \gH_\gS.
\]
To make this explicit, let $e_j$ denote the $j$-th canonical basis vector of $\R^m$. Then,
\begin{align*}
    \mathsf B_R^* e_j = \frac{1}{\sqrt{n}} \sum_{i = 1}^n \mathsf R_{ji} \varphi_\gS(S_i).
\end{align*}
Consequently,
\begin{align*}
\gH_{S,R}
&=\operatorname{span}\left\{
\sum_{i=1}^n R_{ji}\varphi_S(S_i):j=1,\ldots,m
\right\}.
\end{align*}

The orthogonal projector onto this subspace is
\begin{align}
P_R
:=\mathsf B_R^*
(\mathsf B_R\mathsf B_R^*)^\dagger
\mathsf B_R,
\label{eq:nystrom-generic-sketch-projector}
\end{align}
where \(\dagger\) denotes Moore-Penrose pseudoinverse. To verify the projector formula, let $r_R=\operatorname{rank}(\mathsf B_R)$ and consider the reduced singular value decomposition
\begin{align*}
\mathsf B_R=U_R\Sigma_RV_R^*,
\end{align*}
where $U_R:\R^{r_R}\to\R^{m}$ and $V_R:\R^{r_R}\to\gH_S$ are isometries, and $\Sigma_R$ is diagonal with strictly positive entries. Then
$P_R=V_RV_R^*$.

For the Nystr\"om estimator used here, choose $1\leq m\leq n$, draw distinct landmark indices
\begin{align*}
L=\{\ell_1,\ldots,\ell_m\}
\subseteq\{1,\ldots,n\}
\end{align*}
uniformly without replacement \citep{rudi2015less}, and define the coordinate-selection matrix $\mathsf R_L\in\R^{m\times n}$ defined, for $a\in\R^n$, by
\begin{align*}
(\mathsf R_L a)_j = a_{\ell_j},
\qquad j=1,\ldots,m.
\end{align*}

Next, define the landmark synthesis operator as
\begin{align*}
\Phi_L:\R^m\to\gH_S,
\quad
\Phi_Lc=\sum_{j=1}^m c_j\varphi_S(S_{\ell_j}).
\end{align*}
Its adjoint is the landmark evaluation operator given by
\begin{align*}
    \Phi_L^* f = \begin{bmatrix}
        f(S_{\ell_1}) \\ \vdots \\ f(S_{\ell_m})
    \end{bmatrix} \in \R^m
\end{align*}

The normalized sampling operator and the landmark synthesis operator are related by
\begin{align*}
    \mathsf R_L\mathsf T_S
&=\frac{1}{\sqrt{n}} \Phi_L^*, \quad
\mathsf T_S^* \mathsf R_L^\top
=\frac{1}{\sqrt{n}}\Phi_L,
\end{align*}
Consequently, the subspace induced by the Nystr\"om sketch is 
\begin{align*}
\gH_{S,R_L} =\operatorname{Range}(\mathsf T_S^* \mathsf R_L^\top)
=\operatorname{Range}(\Phi_L)
=\operatorname{span}\{\varphi_S(S_\ell):\ell\in L\}.
\end{align*}
Thus the Nyström approximation restricts the dependence of the conditional mean embedding to the span of the landmark input features. Define the landmark Gram matrix \(W :=\Phi_L^*\Phi_L\) with entries given by \(W_{jr} = \langle \varphi_\gS(S_{\ell_j}), \varphi(S_{\ell_r})\rangle_{\gH_\gS} = k_\gS(S_{\ell_j}, S_{\ell_r}).\) Hence,
\[
W = \mathsf R_L K_{S} \mathsf R_L^\top
=K_{S}[L,L] \in \R^{m \times m}.
\]

The generic sketch projector now becomes
\begin{align}
P_L=P_{R_L}=\Phi_LW^\dagger\Phi_L^*.
\label{eq:nystrom-landmark-projector}
\end{align}

\noindent\textbf{Projected regression problem.} The sketch projector $P_R$ decomposes the input RKHS as
\begin{align*}
    \gH_\gS = \gH_{\gS, R} \oplus \gH_{\gS, R}^\perp, \quad \gH_{\gS, R} = \operatorname{Range}(P_R).
\end{align*}
We restrict the regression operator to use only the component of the input feature lying in \(\gH_{\gS, R}.\) Define the restricted operator class
\begin{align*}
\gS_{2, R} :=\left\{  H \in \gS(\gH_\gS, \gH_\gZ) : H = H P_R   \right\}.
\end{align*}
The condition \(H=HP_R\) is equivalent to \(H(I_{\gH_\gS} - P_R) = 0\). Thus, \(H\) annihilates all the input directions orthogonal to the sketch-induced subspace. For the associated regression function, we have 
\[
m_H(s)=H\varphi_S(s)=H P_R \varphi_S(s).
\]
Therefore, the estimator depends on $s$ only through the projected feature $P_R \varphi_\gS(s)$. The corresponding input-sketched vector-valued function class is 
\begin{align*}
\gG_R
:=\left\{
m_H:m_H(s)=H\varphi_S(s),\
H\in \gS_{2, R}
\right\}.
\end{align*}

Next, we define the input-sketched ridge estimator by 
\begin{align}
\widehat H_{R,\lambda}
\in\operatorname*{arg\,min}_{\substack{
H\in\mathcal S_{2, R}}}
\left\{
\norm{\mathsf T_Z^*-H\mathsf T_S^*}_{\HS}^2
+\lambda\norm H_{\HS}^2
\right\}.
\label{eq:nystrom-projected-operator-objective}
\end{align}
Equivalently, 
\begin{align}
\widehat H_{R,\lambda}
\in\operatorname*{arg\,min}_{\substack{
H\in\mathcal S_2(\gH_S,\gH_Z)\\ H=HP_R}}
\left\{ \frac{1}{n} \sum_{i = 1}^n 
\norm{ \varphi_\gZ(Z_i) - H P_R \varphi_\gS(S_i)}_{\gH_\gZ}^2
+\lambda\norm H_{\HS}^2
\right\}.
\label{eq:nystrom-projected-operator-objective-equivalent}
\end{align}
It is ordinary Hilbert-valued ridge regression after replacing each input feature \(\varphi_\gS(S_i)\) by its projection onto the sketched subspace. This estimator is the input-only sketching specialization of the low-rank Hilbert-valued estimator of \citet[Section~3 and Appendix~C, Lemma~1, Equations~(29)--(34)]{pmlr-v238-el-ahmad24a}. For Nystr\"om row selection, this is
the Hilbert-valued analogue of the landmark-subspace estimator of \citet[Section~2, Eqs.~(3)--(5)]{rudi2015less}.

Recall that \(r_R=\operatorname{rank}(\mathsf B_R)\) and that the
reduced singular value decomposition of the sketching operator is
\(
    \mathsf B_R
    =
    U_R\Sigma_RV_R^*,
\)
where
\(
    V_R:\mathbb R^{r_R}\to\mathcal H_S
\)
is an isometry satisfying
\(
    V_R^*V_R=I_{r_R},
     \text{ and } 
    V_RV_R^*=P_R.
\)
Hence the columns of \(V_R\) form an orthonormal basis of the sketch-induced subspace
\(
    \mathcal H_{S,R}
    =
    \operatorname{Range}(P_R).
\)
In particular, every \(f\in\mathcal H_{S,R}\) admits the unique
representation
\[
    f=V_Ra,
    \qquad
    a=V_R^*f\in\mathbb R^{r_R}.
\]

We define the reduced empirical covariance matrix by
\begin{equation}
    \widehat C_{S,R}^{\mathrm{red}}
    :=
    V_R^*\widehat C_SV_R
    \in\mathbb R^{r_R\times r_R}.
    \label{eq:nystrom-reduced-covariance}
\end{equation}
This matrix is the coordinate representation of the empirical covariance operator \(\widehat C_S\) on the sketched subspace. Indeed,
for every \(a\in\mathbb R^{r_R}\),
\begin{align*}
    a^\top
    \widehat C_{S,R}^{\mathrm{red}}
    a
    &=
    \left\langle
        V_Ra,
        \widehat C_SV_Ra
    \right\rangle_{\mathcal H_S} =
    \frac1n
    \sum_{i=1}^n
    \left|
        \left\langle
            V_Ra,
            \varphi_S(S_i)
        \right\rangle_{\mathcal H_S}
    \right|^2
    \geq 0.
\end{align*}
Thus \(\widehat C_{S,R}^{\mathrm{red}}\) is symmetric and positive semi-definite. Consequently, for every \(\lambda>0\),
\(
    \widehat C_{S,R}^{\mathrm{red}}
    +
    \lambda I_{r_R}
    \succeq
    \lambda I_{r_R},
\)
and the regularized reduced covariance is invertible.

\begin{definition}[Restricted regularized inverse]
\label{def:nystrom-restricted-inverse}
Following {\citet[Lemma~1, Equation (29)]{pmlr-v238-el-ahmad24a}}, define the restricted regularized inverse for \(\lambda>0\)
\begin{equation}
    \eta_{R,\lambda}(\widehat C_S)
    :=
    V_R
    \left(
        \widehat C_{S,R}^{\mathrm{red}}
        +
        \lambda I_{r_R}
    \right)^{-1}
    V_R^*
    :
    \mathcal H_S\to\mathcal H_S.
    \label{eq:nystrom-restricted-inverse}
\end{equation}
\end{definition}


The following result is the input-only specialization of \citet[Appendix~C, Lemma~1, Eq.~(29), with the derivation in Eqs.~(31)--(34)]{pmlr-v238-el-ahmad24a}.

\begin{proposition}[Projected Hilbert-valued ridge solution]
\label{prop:nystrom-projected-operator-solution}
The optimization problem
\eqref{eq:nystrom-projected-operator-objective} admits a unique minimizer, given by
\begin{equation}
    \widehat H_{R,\lambda}
    =
    \widehat C_{ZS}
    \eta_{R,\lambda}(\widehat C_S).
    \label{eq:nystrom-projected-operator-solution}
\end{equation}
Moreover, the restricted inverse satisfies
\begin{align}
    P_R
    \left(
        \widehat C_S+\lambda I_{\mathcal H_S}
    \right)
    \eta_{R,\lambda}(\widehat C_S)
    &=
    P_R,
    \label{eq:nystrom-left-inverse-identity}
    \\
    \eta_{R,\lambda}(\widehat C_S)
    \left(
        \widehat C_S+\lambda I_{\mathcal H_S}
    \right)
    P_R
    &=
    P_R.
    \label{eq:nystrom-right-inverse-identity}
\end{align}
\end{proposition}

The purpose of \(\eta_{R,\lambda}(\widehat C_S)\) is not only to express the projected estimator compactly. It will also isolate the effect of restricting the regression to the sketch-induced input subspace. In the subsequent error decomposition, the estimator is compared with the population regression operator through a sample term, a projection term, and a population ridge-bias term. The inverse identities above are the main algebraic tool used to control the projection contribution.

\noindent\textbf{Population ridge comparison.}
From this point onward, we specialize the generic sketch to the Nystr\"om landmark sketch \(R_L\), and write
\[
    P_L:=P_{R_L},
    \qquad
    \eta_{L,\lambda}
    :=
    \eta_{R_L,\lambda}(\widehat C_S).
\]
Recall that the corresponding projected estimator is
\[
    \widehat H_{L,\lambda}
    =
    \widehat C_{ZS}\eta_{L,\lambda},
    \qquad
    \widehat m_{L,\lambda}(s)
    :=
    \widehat H_{L,\lambda}\varphi_S(s).
\]

For \(\lambda>0\), define the regularized population covariance operator \(C_{S,\lambda}
    :=
    C_S+\lambda I_{\mathcal H_S}.\) Since \(C_S\) is positive and self-adjoint,
\(
    C_{S,\lambda}
    \succeq
    \lambda I_{\mathcal H_S}.
\)
It is therefore boundedly invertible, with
\(
    \|C_{S,\lambda}^{-1}\|_{\mathrm{op}}
    \leq
    \lambda^{-1}.
\)

Following the input-only specialization of
\citet[Appendix~D, proof of Theorem~1,
Eqs.~(43)--(48)]{pmlr-v238-el-ahmad24a},
we introduce the following comparison operator and error decomposition.

\begin{lemma}[Projected Nystr\"om error decomposition]
\label{lem:nystrom-error-decomposition}
Suppose Assumptions~\ref{ass:nystrom-measurability} and
\ref{ass:nystrom-attainability} hold. For \(\lambda>0\), let
\(
    C_{S,\lambda}
    :=
    C_S+\lambda I_{\mathcal H_S},
\)
and define \(H_\lambda
    :=
    H_0C_SC_{S,\lambda}^{-1}.\)
Then
\begin{equation}
    H_\lambda-H_0
    =
    -\lambda H_0C_{S,\lambda}^{-1}.
    \label{eq:nystrom-ridge-bias-identity}
\end{equation}

For every observed sample and every realized landmark set,
\begin{align}
    \widehat H_{L,\lambda}-H_0
    &=
    \left(
        \widehat C_{ZS}
        -
        H_\lambda\widehat C_S
    \right)
    \eta_{L,\lambda}
     +
    H_\lambda
    \left(
        \widehat C_S\eta_{L,\lambda}
        -
        I_{\mathcal H_S}
    \right)
    +
    \left(
        H_\lambda-H_0
    \right).
    \label{eq:nystrom-exact-operator-decomposition}
\end{align}

Consequently,
\begin{equation}
    r_m(L,\lambda)
    :=
    \left\|
        \widehat m_{L,\lambda}-m_0
    \right\|_{L_2(P_{S,0};\mathcal H_Z)}
    \leq
    (A)_{L,\lambda}
    +
    (B)_{L,\lambda}
    +
    (C)_\lambda,
    \label{eq:nystrom-three-term-error-bound}
\end{equation}
where
\begin{align}
    (A)_{L,\lambda}
    &:=
    \left\|
        \left(
            \widehat C_{ZS}
            -
            H_\lambda\widehat C_S
        \right)
        \eta_{L,\lambda}C_S^{1/2}
    \right\|_{\mathrm{HS}},
    \notag\\
    (B)_{L,\lambda}
    &:=
    \left\|
        H_\lambda
        \left(
            \widehat C_S\eta_{L,\lambda}
            -
            I_{\mathcal H_S}
        \right)
        C_S^{1/2}
    \right\|_{\mathrm{HS}},
    \notag\\
    (C)_\lambda
    &:=
    \left\|
        \left(
            H_\lambda-H_0
        \right)
        C_S^{1/2}
    \right\|_{\mathrm{HS}}.
    \label{eq:nystrom-three-error-terms}
\end{align}
\end{lemma}

\begin{proof}
Since
\(
    C_SC_{S,\lambda}^{-1}
    =
    I_{\mathcal H_S}
    -
    \lambda C_{S,\lambda}^{-1},
\)
Equation~\ref{eq:nystrom-ridge-bias-identity} follows immediately.
Using
\(
    \widehat H_{L,\lambda}
    =
    \widehat C_{ZS}\eta_{L,\lambda},
\)
add and subtract \(H_\lambda\widehat C_S\eta_{L,\lambda}\) and \(H_\lambda\) to obtain Equation \ref{eq:nystrom-exact-operator-decomposition}. The prediction-norm identity in Lemma~\ref{lem:attainability-consequences}, followed by the triangle inequality, yields Equation \ref{eq:nystrom-three-term-error-bound}.
\end{proof}

Lemma~\ref{lem:nystrom-error-decomposition} gives an exact algebraic decomposition of the projected Nystr\"om estimation error into three components, which will be controlled separately. Because the decomposition
holds pathwise for every realized sample and landmark set, it does not yet
require the input-capacity or embedding assumptions, uniform landmark
sampling, or concentration arguments. The sample term
\((A)_{L,\lambda}\) captures fluctuations in the empirical covariance and
cross-covariance operators. The projection/resolvent term
\((B)_{L,\lambda}\) quantifies the error induced by restricting the
regression to the landmark subspace and using the associated restricted
regularized inverse. Finally, \((C)_\lambda\) is the deterministic bias
introduced by population ridge regularization. The additional assumptions
enter only when deriving bounds for these three terms.

\paragraph{Auxiliary operator bounds.}
For \(\lambda>0\), define the regularized empirical covariance
operator
\[
    \widehat C_{S,\lambda}
    :=
    \widehat C_S+\lambda I_{\mathcal H_S}.
\]

We use the covariance-comparison event
\begin{align}
    \mathcal E_{\mathrm{cmp}}(\lambda)
    :=
    \Bigg\{
    &
    \left\|
        \widehat C_{S,\lambda}^{-1/2}
        C_{S,\lambda}^{1/2}
    \right\|_{\mathrm{op}}
    \leq\sqrt2,
    \left\|
        C_{S,\lambda}^{-1/2}
        \widehat C_{S,\lambda}^{1/2}
    \right\|_{\mathrm{op}}
    \leq\sqrt{\frac32}
    \Bigg\}.
    \label{eq:nystrom-covariance-comparison-event}
\end{align}
These are the input-space specializations of
\citet[Lemmas~9--10, Eqs.~(106)--(111)]
{pmlr-v238-el-ahmad24a}.
Applying the two lemmas with failure probability \(\delta/2\) shows
that, under the corresponding ridge condition,
\(\mathcal E_{\mathrm{cmp}}(\lambda)\) holds with probability at least
\(1-\delta\).

We also use the deterministic restricted-resolvent contraction
\begin{equation}
    \left\|
        \widehat C_{S,\lambda}^{1/2}
        \eta_{L,\lambda}
        \widehat C_{S,\lambda}^{1/2}
    \right\|_{\mathrm{op}}
    \leq1,
    \label{eq:nystrom-restricted-resolvent-contraction}
\end{equation}
which is the input-only specialization of
\citet[Lemma~13, Eq.~(124)]
{pmlr-v238-el-ahmad24a}; see also
\citet[Lemma~8]{rudi2015less}.

Finally, define the input-subspace approximation error
\begin{equation}
    \varepsilon_L
    :=
    \left\|
        \left(
            I_{\mathcal H_S}-P_L
        \right)
        C_S^{1/2}
    \right\|_{\mathrm{op}}.
    \label{eq:nystrom-subspace-error}
\end{equation}

The following bound is the input-only specialization of \citet[Lemma~3, Eq.~(55), with the proof in Eqs.~(56)--(59)]{pmlr-v238-el-ahmad24a}.

\begin{lemma}[Bound for the sample term]
\label{lem:nystrom-bound-sample-term}
Suppose the observations are i.i.d., the input and latent kernels are
bounded, Assumption~\ref{ass:nystrom-attainability} and ~\ref{ass:nystrom-input-regularity} hold. Set
\[
    \lambda_n
    :=
    n^{-1/(1+\gamma)}.
\]
Let \(\delta\in(0,1]\), and suppose that \(n\) is sufficiently large
that
\begin{equation}
    \frac{9\kappa_S^2}{n}
    \log\left(\frac{n}{\delta}\right)
    \leq
    \lambda_n
    \leq
    \|C_S\|_{\mathrm{op}}.
    \label{eq:nystrom-finite-ridge-condition-A}
\end{equation}
Then there exists a finite problem-dependent constant \(C_A>0\) such
that, for every realized landmark set \(L\), with probability at least
\(1-\delta\),
\begin{equation}
    (A)_{L,\lambda_n}
    \leq
    C_A
    \log\left(\frac{4}{\delta}\right)
    n^{-\frac{1}{2(1+\gamma)}}.
    \label{eq:nystrom-sample-term-rate}
\end{equation}
\end{lemma}

\begin{proof}
Under the identifications
\(
    X\equiv S,
    \enspace
    Y\equiv Z,
    \enspace
    \widetilde\eta(\widehat C_X)
    \equiv
    \eta_{L,\lambda}(\widehat C_S),
\)
and with the output projector equal to
\(I_{\mathcal H_Z}\), the term
\((A)_{L,\lambda}\) coincides with term \((A)\) in the proof of
\citet[Theorem~1]{pmlr-v238-el-ahmad24a}. The result therefore follows
directly from
\citet[Lemma~3, Eqs.~(55)--(59)]
{pmlr-v238-el-ahmad24a}.
\end{proof}

The following result is the input-only specialization of
\citet[Lemma~4, Eq.~(60), with the proof in
Eqs.~(61)--(73)]{pmlr-v238-el-ahmad24a}.
We use the sharper contraction
\(\|H_\lambda\|_{\mathrm{HS}}\leq\|H_0\|_{\mathrm{HS}}\).

\begin{lemma}[Bound for the projection/resolvent term]
\label{lem:nystrom-bound-projection-term}
For every \(\lambda>0\) and every orthogonal input-sketch projector
\(P_L\), on the covariance-comparison event
\(\mathcal E_{\mathrm{cmp}}(\lambda)\),
\begin{equation}
    (B)_{L,\lambda}
    \leq
    2\sqrt 3\,
    \|H_0\|_{\mathrm{HS}}
    \left(
        \varepsilon_L+\sqrt\lambda
    \right).
    \label{eq:nystrom-projection-term-bound}
\end{equation}
\end{lemma}

\begin{proof}

Set
\(
    \widehat C_{S,\lambda}
    :=
    \widehat C_S+\lambda I_{\mathcal H_S}.
\)
The projected inverse identity
\(
    P_L\widehat C_{S,\lambda}\eta_{L,\lambda}
    =
    P_L
\)
implies that 
\begin{align}
    \widehat C_S\eta_{L,\lambda}
    -
    I_{\mathcal H_S}
    &=
    \left(
        I_{\mathcal H_S}-P_L
    \right)
    \widehat C_{S,\lambda}\eta_{L,\lambda}
    -
    \lambda\eta_{L,\lambda}
    -
    \left(
        I_{\mathcal H_S}-P_L
    \right).
    \label{eq:nystrom-projection-resolvent-decomposition}
\end{align}
Consequently,
\[
    (B)_{L,\lambda}
    \leq
    \|H_\lambda\|_{\mathrm{HS}}
    \left(
        B_1+B_2+B_3
    \right),
\]
where
\begin{align*}
    B_1
    &:=
    \left\|
        \left(
            I_{\mathcal H_S}-P_L
        \right)
        \widehat C_{S,\lambda}
        \eta_{L,\lambda}
        C_S^{1/2}
    \right\|_{\mathrm{op}},
    \enspace
    B_2
    :=
    \lambda
    \left\|
        \eta_{L,\lambda}C_S^{1/2}
    \right\|_{\mathrm{op}},
    \enspace
    B_3
    :=
    \left\|
        \left(
            I_{\mathcal H_S}-P_L
        \right)
        C_S^{1/2}
    \right\|_{\mathrm{op}}
    =
    \varepsilon_L.
\end{align*}

Using the covariance-comparison inequalities and the deterministic
restricted-resolvent contraction, the operator factorizations in
\citet[proof of Lemma~4, Eqs.~(70)--(72)]
{pmlr-v238-el-ahmad24a} give
\[
    B_1
    \leq
    \sqrt3
    \left(
        \varepsilon_L+\sqrt\lambda
    \right),
    \qquad
    B_2
    \leq
    \sqrt{2\lambda}.
\]
Therefore,
\begin{align*}
    B_1+B_2+B_3
    &\leq
    \sqrt3
    \left(
        \varepsilon_L+\sqrt\lambda
    \right)
    +
    \sqrt{2\lambda}
    +
    \varepsilon_L
    \leq
    2\sqrt3
    \left(
        \varepsilon_L+\sqrt\lambda
    \right).
\end{align*}

Finally,
\[
    H_\lambda
    =
    H_0C_S(C_S+\lambda I_{\mathcal H_S})^{-1},
\]
and spectral calculus gives
\[
    \left\|
        C_S(C_S+\lambda I_{\mathcal H_S})^{-1}
    \right\|_{\mathrm{op}}
    \leq1.
\]
Hence
\(
    \|H_\lambda\|_{\mathrm{HS}}
    \leq
    \|H_0\|_{\mathrm{HS}},
\)
which proves
\eqref{eq:nystrom-projection-term-bound}.

\end{proof}

The following lemma is the input-only counterpart of
\citet[Lemma~5, Eqs.~(74)--(81)]
{pmlr-v238-el-ahmad24a}. Because the output RKHS is not sketched,
their output-projection term vanishes. We use a direct spectral
calculation to obtain a slightly sharper constant.

\begin{lemma}[Population ridge bias]
\label{lem:nystrom-bound-ridge-bias}
For every \(\lambda>0\),
\begin{equation}
    (C)_\lambda
    \leq
    \frac{\sqrt{\lambda}}{2}
    \|H_0\|_{\mathrm{HS}}.
    \label{eq:nystrom-ridge-bias-bound}
\end{equation}
\end{lemma}

\begin{proof}
By
\eqref{eq:nystrom-ridge-bias-identity},
\(
    H_\lambda-H_0
    =
    -\lambda H_0C_{S,\lambda}^{-1}.
\)
Hence
\begin{align*}
    (C)_\lambda
    &=
    \lambda
    \left\|
        H_0C_{S,\lambda}^{-1}C_S^{1/2}
    \right\|_{\mathrm{HS}}
    \leq
    \|H_0\|_{\mathrm{HS}}
    \left\|
        \lambda C_{S,\lambda}^{-1}C_S^{1/2}
    \right\|_{\mathrm{op}}.
\end{align*}
Notice that
\begin{align*}
    \left\|
        \lambda C_{S,\lambda}^{-1}C_S^{1/2}
    \right\|_{\mathrm{op}}
    &=
    \sup_{\sigma\in\operatorname{spec}(C_S)}
    \frac{\lambda\sqrt{\sigma}}{\sigma+\lambda}.
\end{align*}
Since
\(
    \sigma+\lambda
    \geq
    2\sqrt{\sigma\lambda},
\)
we have
\(
    \frac{\lambda\sqrt{\sigma}}{\sigma+\lambda}
    \leq
    \frac{\sqrt{\lambda}}{2}
\)
for every \(\sigma\geq0\). Therefore,
\(
    (C)_\lambda
    \leq
    \frac{\sqrt{\lambda}}{2}
    \|H_0\|_{\mathrm{HS}}.
\)
\end{proof}

\noindent\textbf{Reduction to the Nystr\"om projection error.} Combining Lemmas~\ref{lem:nystrom-bound-sample-term}, \ref{lem:nystrom-bound-projection-term}, and \ref{lem:nystrom-bound-ridge-bias}, we obtain, on \(\mathcal E_{\mathrm{cmp}}(\lambda)\),
\begin{align}
    r_m(L,\lambda)
    &\leq
    (A)_{L,\lambda}
    +
    2\sqrt 3\,
    \|H_0\|_{\mathrm{HS}}\,
    \varepsilon_L
    +
    \left(
        2\sqrt 3+\frac12
    \right)
    \|H_0\|_{\mathrm{HS}}
    \sqrt\lambda .
    \label{eq:nystrom-pre-projection-bound}
\end{align}
It remains to control the random subspace error \(\varepsilon_L\) defined in \eqref{eq:nystrom-subspace-error}.

For \(\lambda>0\), define the effective dimension
\begin{equation}
    \mathcal N(\lambda)
    :=
    \operatorname{Tr}
    \left[
        C_SC_{S,\lambda}^{-1}
    \right],
    \label{eq:nystrom-effective-dimension}
\end{equation}
and the maximal ridge leverage
\begin{equation}
    \mathcal N_\infty(\lambda)
    :=
    \operatorname*{ess\,sup}_{s\sim P_{S,0}}
    \left\langle
        \varphi_S(s),
        C_{S,\lambda}^{-1}
        \varphi_S(s)
    \right\rangle_{\mathcal H_S}.
    \label{eq:nystrom-maximal-leverage}
\end{equation}

The effective dimension controls the average regularized complexity of the input RKHS, whereas \(\mathcal N_\infty(\lambda)\) controls the largest regularized leverage among input features and therefore determines the number of uniformly sampled landmarks required below.

\begin{lemma}[Uniform Nystr\"om projection control]
\label{lem:nystrom-uniform-projection}
Let \(L\subseteq\{1,\ldots,n\}\) contain \(m\) indices sampled
uniformly without replacement, and let \(P_L\) be the orthogonal
projector onto
\[
    \operatorname{span}
    \left\{
        \varphi_S(S_\ell):\ell\in L
    \right\}.
\]
If
\begin{equation}
    m
    \geq
    \left\{
        67
        \vee
        5\mathcal N_\infty(\lambda)
    \right\}
    \log
    \left(
        \frac{4\kappa_S^2}
             {\lambda\delta}
    \right),
    \label{eq:nystrom-landmark-condition}
\end{equation}
then, with probability at least \(1-\delta\),
\begin{equation}
    \left\|
        \left(
            I_{\mathcal H_S}-P_L
        \right)
        C_{S,\lambda}^{1/2}
    \right\|_{\mathrm{op}}^2
    \leq
    3\lambda.
    \label{eq:nystrom-regularized-projection-event}
\end{equation}
Consequently,
\begin{equation}
    \varepsilon_L
    \leq
    \sqrt{3\lambda}.
    \label{eq:nystrom-subspace-error-bound}
\end{equation}
\end{lemma}

\begin{proof}
The regularized projection bound
\eqref{eq:nystrom-regularized-projection-event}
is the plain Nystr\"om approximation result of
\citet[Lemma~6]{rudi2015less}, applied to the covariance operator
\(C_S\) and the subspace generated by the uniformly sampled
landmarks. Since \(\mathcal N_\infty(\lambda)\) is defined through an
essential supremum, we apply the result on a
\(P_{S,0}\)-full-measure subset on which the leverage bound holds.

For the second claim, write
\[
    C_S^{1/2}
    =
    C_{S,\lambda}^{1/2}
    C_{S,\lambda}^{-1/2}
    C_S^{1/2}.
\]
Hence
\begin{align*}
    \varepsilon_L
    &\leq
    \left\|
        \left(
            I_{\mathcal H_S}-P_L
        \right)
        C_{S,\lambda}^{1/2}
    \right\|_{\mathrm{op}}
    \left\|
        C_{S,\lambda}^{-1/2}
        C_S^{1/2}
    \right\|_{\mathrm{op}}.
\end{align*}
Spectral calculus gives
\[
    \left\|
        C_{S,\lambda}^{-1/2}
        C_S^{1/2}
    \right\|_{\mathrm{op}}
    =
    \sup_{\sigma\in\operatorname{spec}(C_S)}
    \sqrt{\frac{\sigma}{\sigma+\lambda}}
    \leq1.
\]
The conclusion follows from
\eqref{eq:nystrom-regularized-projection-event}.
\end{proof}

Consequently, on the intersection of \(\mathcal E_{\mathrm{cmp}}(\lambda)\) and the Nystr\"om projection event in \eqref{eq:nystrom-regularized-projection-event},
\begin{align}
    r_m(L,\lambda)
    &\leq
    (A)_{L,\lambda}
    +
    C_H\sqrt\lambda,
    \label{eq:nystrom-post-projection-bound}
\end{align}
where
\[
    C_H
    :=
    \left[
        2\sqrt3(\sqrt3+1)
        +
        \frac12
    \right]
    \|H_0\|_{\mathrm{HS}}.
\]

\begin{lemma}[Capacity bounds for the leverage quantity]
\label{lem:nystrom-leverage-bounds}
Under the capacity condition in Assumption~\ref{ass:nystrom-input-regularity}, we have, for every \(\lambda>0\),
\begin{equation}
    \mathcal N(\lambda)
    \leq
    Q_S\lambda^{-\gamma}.
    \label{eq:nystrom-effective-dimension-bound}
\end{equation}
Moreover, boundedness of \(k_S\) always gives
\begin{equation}
    \mathcal N_\infty(\lambda)
    \leq
    \frac{\kappa_S^2}{\lambda}.
    \label{eq:nystrom-universal-leverage-bound}
\end{equation}
If, in addition, the embedding condition holds and
\(\gamma+\mu\leq1\), then
\begin{equation}
    \mathcal N_\infty(\lambda)
    \leq
    B_SQ_S\lambda^{-(\gamma+\mu)}.
    \label{eq:nystrom-refined-leverage-bound}
\end{equation}
\end{lemma}

\begin{proof}
The effective-dimension bound is the input-side specialization of \citet[Lemma~11, Eqs.~(115)--(118)] {pmlr-v238-el-ahmad24a}.
The universal bound follows from
\(C_{S,\lambda}^{-1}\preceq\lambda^{-1}I\) and
\(\|\varphi_S(s)\|_{\mathcal H_S}^2\leq\kappa_S^2\).

For the final claim, the embedding condition gives
\[
    \mathcal N_\infty(\lambda)
    \leq
    B_S
    \sum_j
    \frac{\sigma_j^{1-\mu}}
         {\sigma_j+\lambda},
\]
where \((\sigma_j)_j\) are the eigenvalues of \(C_S\).
When \(\gamma+\mu\leq1\),
\[
    \frac{\sigma^{1-\mu}}
         {\sigma+\lambda}
    \leq
    \lambda^{-(\gamma+\mu)}
    \sigma^\gamma
\]
for every \(\sigma\geq0\). Summing over \(j\) proves
\eqref{eq:nystrom-refined-leverage-bound}.
\end{proof}

\begin{theorem}[Projected Nystr\"om CME rate]
\label{thm:nystrom-cme-rate}
Suppose Assumptions~\ref{ass:nystrom-measurability} and
\ref{ass:nystrom-attainability} hold, and suppose that, for some
\(\gamma\in(0,1]\),
\(
    \operatorname{Tr}(C_S^\gamma)<\infty.
\)
Set
\[
    \lambda_n
    :=
    n^{-1/(1+\gamma)}.
\]
For each \(n\), let \(L_n\subseteq\{1,\ldots,n\}\) contain \(1\leq m_n\leq n\) indices sampled uniformly without replacement.

Fix \(\delta\in(0,1)\). Suppose \(n\) is sufficiently large that
\begin{equation}
    \frac{9\kappa_S^2}{n}
    \log\left(\frac{2n}{\delta}\right)
    \leq
    \lambda_n
    \leq
    \|C_S\|_{\mathrm{op}},
    \label{eq:nystrom-final-ridge-condition}
\end{equation}
and suppose
\begin{equation}
    m_n
    \geq
    \left\{
        67
        \vee
        5\mathcal N_\infty(\lambda_n)
    \right\}
    \log
    \left(
        \frac{8\kappa_S^2}
             {\lambda_n\delta}
    \right).
    \label{eq:nystrom-final-landmark-condition}
\end{equation}
Then there exists a finite problem-dependent constant \(C_A>0\), independent of \(n\), \(m_n\), and \(\delta\), such that, with probability at least \(1-\delta\),
\begin{equation}
    \left\|
        \widehat m_{L_n,\lambda_n}-m_0
    \right\|_{L_2(P_{S,0};\mathcal H_Z)}
    \leq
    \left\{
        C_A\log\left(\frac{8}{\delta}\right)
        +
        C_H
    \right\}
    n^{-\frac{1}{2(1+\gamma)}}.
    \label{eq:nystrom-cme-high-probability-rate}
\end{equation}
\end{theorem}

\begin{proof}
Apply Lemma~\ref{lem:nystrom-bound-sample-term} with confidence
level \(\delta/2\). With probability at least \(1-\delta/2\),
\[
    (A)_{L_n,\lambda_n}
    \leq
    C_A
    \log\left(\frac{8}{\delta}\right)
    n^{-\frac{1}{2(1+\gamma)}},
\]
and the covariance-comparison event required by
Lemma~\ref{lem:nystrom-bound-projection-term} holds.

Condition in Equation \ref{eq:nystrom-final-landmark-condition} and
Lemma~\ref{lem:nystrom-uniform-projection}, applied with confidence
level \(\delta/2\), give an event of probability at least
\(1-\delta/2\) on which
\[
    \varepsilon_{L_n}
    \leq
    \sqrt{3\lambda_n}.
\]
On the intersection of these two events,
\begin{align*}
    (B)_{L_n,\lambda_n}
    &\leq
    2\sqrt3\,
    \|H_0\|_{\mathrm{HS}}
    \left(
        \varepsilon_{L_n}
        +
        \sqrt{\lambda_n}
    \right)
    \leq
    2\sqrt3(\sqrt3+1)
    \|H_0\|_{\mathrm{HS}}
    \sqrt{\lambda_n},
\end{align*}
while Lemma~\ref{lem:nystrom-bound-ridge-bias} gives
\[
    (C)_{\lambda_n}
    \leq
    \frac12
    \|H_0\|_{\mathrm{HS}}
    \sqrt{\lambda_n}.
\]
Combining the three terms and using
\[
    \sqrt{\lambda_n}
    =
    n^{-\frac{1}{2(1+\gamma)}}
\]
proves the claim.
\end{proof}

\begin{corollary}[Explicit landmark schedule]
\label{cor:nystrom-explicit-schedule}
Suppose Assumption~\ref{ass:nystrom-input-regularity} holds, and define
\begin{equation}
    \vartheta
    :=
    \min\{\gamma+\mu,1\},
    \qquad
    C_\star
    :=
    \begin{cases}
        B_SQ_S,
        & \gamma+\mu\leq1,\\[2pt]
        \kappa_S^2,
        & \gamma+\mu>1.
    \end{cases}
    \label{eq:nystrom-landmark-exponent}
\end{equation}
Then
\begin{equation}
    \mathcal N_\infty(\lambda)
    \leq
    C_\star\lambda^{-\vartheta}.
    \label{eq:nystrom-piecewise-leverage-bound}
\end{equation}

Consequently, for
\[
    \lambda_n
    =
    n^{-1/(1+\gamma)},
\]
a sufficient finite-sample landmark condition is
\begin{equation}
    m_n
    \geq
    \left\{
        67
        \vee
        5C_\star
        n^{\frac{\vartheta}{1+\gamma}}
    \right\}
    \log
    \left(
        \frac{8\kappa_S^2}{\delta}
        n^{\frac1{1+\gamma}}
    \right).
    \label{eq:nystrom-explicit-finite-sample-schedule}
\end{equation}
For fixed \(\delta\), it is therefore sufficient that
\begin{equation}
    m_n
    \gtrsim
    n^{\frac{\vartheta}{1+\gamma}}
    \log n.
    \label{eq:nystrom-explicit-landmark-schedule}
\end{equation}
Under this schedule,
\begin{equation}
    \left\|
        \widehat m_{L_n,\lambda_n}
        -
        m_0
    \right\|_{L_2(P_{S,0};\mathcal H_Z)}
    =
    O_{\mathbb P}
    \left(
        n^{-\frac1{2(1+\gamma)}}
    \right).
    \label{eq:nystrom-cme-op-rate}
\end{equation}
\end{corollary}

\begin{proof}
The piecewise leverage bound follows from
Lemma~\ref{lem:nystrom-leverage-bounds}: use
\eqref{eq:nystrom-refined-leverage-bound} when
\(\gamma+\mu\leq1\), and
\eqref{eq:nystrom-universal-leverage-bound} otherwise.
Substituting
\[
    \lambda_n
    =
    n^{-1/(1+\gamma)}
\]
into
\eqref{eq:nystrom-final-landmark-condition}
gives
\eqref{eq:nystrom-explicit-finite-sample-schedule}.
The rate then follows from
Theorem~\ref{thm:nystrom-cme-rate}.
\end{proof}

\paragraph{Same-sample embedding estimation.}
We next verify the outcome-side class control needed when the regression
and the DR score use the same observations. All asymptotic statements below
use the same ridge and landmark sequences, with the finite-sample conditions
of Theorem~\ref{thm:nystrom-cme-rate} holding eventually for each fixed
\(\delta\in(0,1)\).

\begin{lemma}[Tightness of the projected regression operator]
\label{lem:nystrom-operator-tightness}
Under the conditions of Theorem~\ref{thm:nystrom-cme-rate},
\begin{equation}
    \|\widehat H_{L_n,\lambda_n}\|_{\mathrm{HS}}=O_{\mathbb P}(1),
    \qquad
    \|\widehat H_{L_n,\lambda_n}-H_0\|_{\mathrm{op}}=O_{\mathbb P}(1).
    \label{eq:nystrom-operator-tightness}
\end{equation}
\end{lemma}

\begin{proof}
Write \(T=C_S\), \(\widehat T=\widehat C_S\), \(P=P_{L_n}\),
\(\eta=\eta_{L_n,\lambda_n}\), and \(\lambda=\lambda_n\). Define
\[
    r_0(O)=\varphi_Z(Z)-H_0\varphi_S(S),
    \qquad E_n=P_n[r_0\otimes\varphi_S].
\]
Then \(\mathbb E[r_0\mid S]=0\), \(\|r_0\|\leq2\kappa_Z\) almost surely,
and the projected solution satisfies
\(\widehat H_{L_n,\lambda_n}=(H_0\widehat T+E_n)\eta\).
On the covariance-comparison and regularized projection events
\eqref{eq:nystrom-covariance-comparison-event} and
\eqref{eq:nystrom-regularized-projection-event},
\[
    \|(I-P)\widehat T^{1/2}\|_{\mathrm{op}}
    \leq3\sqrt{\lambda/2}.
\]
The restricted inverse identities give
\(\eta(\widehat T+\lambda I)\eta=\eta\),
\(0\preceq\eta\preceq\lambda^{-1}P\), and therefore
\[
    \|\widehat T^{1/2}\eta\|_{\mathrm{op}}
    \leq\frac1{2\sqrt\lambda},
    \qquad
    \|(T+\lambda I)^{1/2}\eta\|_{\mathrm{op}}
    \leq\sqrt{2/\lambda}.
\]
Since \(\widehat T\eta=P-\lambda\eta+(I-P)\widehat T\eta\),
\(\|\widehat T\eta\|_{\mathrm{op}}\leq1+3/(2\sqrt2)\).
For the noise term, independence and conditional centering yield
\[
    \mathbb E\|E_n(T+\lambda I)^{-1/2}\|_{\mathrm{HS}}^2
    \leq\frac{4\kappa_Z^2}{n}\mathcal N(\lambda).
\]
Factoring \(E_n\eta\) through \((T+\lambda I)^{-1/2}\) consequently gives
\[
    \|\widehat H_{L_n,\lambda_n}\|_{\mathrm{HS}}
    =O_{\mathbb P}\!\left(1+
      \sqrt{\frac{\mathcal N(\lambda_n)}{n\lambda_n}}\right)
    =O_{\mathbb P}(1),
\]
because \(\mathcal N(\lambda)\leq Q_S\lambda^{-\gamma}\), where
\(Q_S=\operatorname{Tr}(C_S^\gamma)<\infty\), and
\(n\lambda_n^{1+\gamma}=1\). The event probabilities can be made
arbitrarily close to one by the stated schedule. Finally,
\(\|\widehat H-H_0\|_{\mathrm{op}}
\leq\|\widehat H\|_{\mathrm{HS}}+\|H_0\|_{\mathrm{HS}}\).
\end{proof}

For the exact-policy-integral estimator
\(\widehat h_\pi=P_n\Gamma_{\widehat\eta}\), put
\(\Gamma_0=\Gamma_{m_0,w_0}\) and
\(\mathbb G_n=\sqrt n(P_n-P_0)\). The population DR identity
\eqref{eq:dr-target-remainder} gives the following exact Hilbert-valued
von Mises decomposition, with oracle influence function
\(\Gamma_0-\chi(\pi)\):
\begin{equation}
\begin{split}
    \widehat h_\pi-\chi(\pi)
    &=(P_n-P_0)\Gamma_0
      +(P_n-P_0)(\Gamma_{\widehat\eta}-\Gamma_0)\\
    &\quad-P_0[(\widehat w-w_0)(\widehat m-m_0)].
\end{split}
\label{eq:same-sample-von-mises}
\end{equation}
Here \(P_0\) integrates with the fitted functions held fixed; it is not
conditional expectation given the training data. The product remainder has
norm at most
\(\|\widehat w-w_0\|_{L_2(P_{S,0})}
\|\widehat m-m_0\|_{L_2(P_{S,0};\mathcal H_Z)}\).

\begin{proposition}[Same-sample score control and equicontinuity]
\label{prop:same-sample-equicontinuity}
Under Assumptions~\ref{assum:identification}, \ref{ass:nystrom-measurability}, and
\ref{ass:nystrom-attainability}, fix the target policy, encoder, and kernels,
and suppose \(P_0w_0^2<\infty\). Let
\(\widehat m=\widehat H\varphi_S\), with
\(\|\widehat H-H_0\|_{\mathrm{op}}=O_{\mathbb P}(1)\).
For estimated weights, suppose the error
\(q_n=\widehat w-w_0\) belongs to a fixed separable scalar RKHS
\(\mathcal K_w\), whose strongly measurable feature map \(t\) satisfies
\(P_0\|t(S)\|_{\mathcal K_w}^2<\infty\), and
\(\|q_n\|_{\mathcal K_w}=O_{\mathbb P}(1)\).
Then
\[
    \|\mathbb G_n(\Gamma_{\widehat\eta}-\Gamma_0)\|_{\mathcal H_Z}
    =O_{\mathbb P}(1).
\]
If, additionally, both nuisance errors tend to zero in their respective
\(L_2(P_{S,0})\) norms, the right-hand side improves to
\(o_{\mathbb P}(1)\), and
\begin{equation}
    \widehat h_\pi-\chi(\pi)
    =(P_n-P_0)\Gamma_0
    -P_0[(\widehat w-w_0)(\widehat m-m_0)]
    +o_{\mathbb P}(n^{-1/2}).
    \label{eq:same-sample-oracle-expansion}
\end{equation}
For known weights, take \(q_n=0\) and omit all ratio-class conditions.
\end{proposition}

\begin{proof}
Write \(v=\varphi_S\), \(D=\widehat H-H_0\), and define the fixed functions
\[
    v_\pi(X)=\int v(a,X)\pi(da\mid X),\qquad
    b_\pi=v_\pi(X)-w_0(S)v(S),\qquad
    r_0=\varphi_Z(Z)-H_0v(S).
\]
Then \(P_0b_\pi=0\), \(B_\pi^2:=P_0\|b_\pi\|^2<\infty\),
and \(\|r_0\|\leq2\kappa_Z\). Direct expansion gives, pathwise,
\begin{equation}
    \mathbb G_n(\Gamma_{\widehat\eta}-\Gamma_0)
    =D\mathbb G_nb_\pi+\mathbb G_n[q_nr_0]
      -D\mathbb G_n[q_nv].
    \label{eq:same-sample-score-factorization}
\end{equation}
For either bounded Hilbert-valued function \(c=r_0\) or \(c=v\),
put \(U_{n,c}=\mathbb G_n[c\otimes t]\). The reproducing property gives
\(\mathbb G_n[qc]=U_{n,c}q\), while
\[
    \mathbb E\|\mathbb G_nb_\pi\|^2=B_\pi^2,\qquad
    \mathbb E\|U_{n,c}\|_{\mathrm{HS}}^2
    \leq\|c\|_\infty^2 P_0\|t\|^2.
\]
Thus \eqref{eq:same-sample-score-factorization} proves the
\(O_{\mathbb P}(1)\) assertion without independence of the fitted nuisances.

For the stronger conclusion, fix \(R<\infty\), and let
\(\mathcal A_R(\delta)\) consist of pairs \((D,q)\) satisfying
\[
    \|D\|_{\mathrm{op}},\ \|q\|_{\mathcal K_w}\leq R,
    \qquad
    \|Dv\|_{L_2(P_{S,0};\mathcal H_Z)},\
    \|q\|_{L_2(P_{S,0})}\leq\delta.
\]
Let \((\sigma_j,u_j)\) be the positive eigenpairs of \(C_S\),
\(\Pi_J\) their first-\(J\) projector, and
\(a_J^2=P_0\|(I-\Pi_J)b_\pi\|^2\).
Separability puts \(v(S)\), and hence
\(v_\pi(X)=\mathbb E[w_0v\mid X]\) and \(b_\pi\), in the closed
covariance support on one common probability-one event. Therefore
\(a_J\to0\). On \(\mathcal A_R(\delta)\),
\(\|D\Pi_J\|_{\mathrm{op}}\leq\delta/\sqrt{\sigma_J}\), so
\[
    \mathbb E^*\sup_{\mathcal A_R(\delta)}\|D\mathbb G_nb_\pi\|
    \leq\frac{\delta B_\pi}{\sqrt{\sigma_J}}+Ra_J.
\]
Similarly, let \((\rho_j,e_j)\) be the positive eigenpairs of the
trace-class operator \(T_w=P_0[t\otimes t]\), with first-\(K\)
projector \(\Lambda_K\). The same common-event argument eliminates its
nullspace. Since
\(\|\Lambda_Kq\|\leq\delta/\sqrt{\rho_K}\) and
\(\mathbb E\|U_{n,c}(I-\Lambda_K)\|_{\mathrm{HS}}^2
\leq\|c\|_\infty^2\sum_{j>K}\rho_j\),
\[
    \mathbb E^*\sup_{\mathcal A_R(\delta)}\|\mathbb G_n[qc]\|
    \leq\|c\|_\infty\left\{
      \delta\sqrt{\frac{\operatorname{Tr}(T_w)}{\rho_K}}
      +R\sqrt{\sum_{j>K}\rho_j}\right\}.
\]
All bounds are uniform in \(n\); outer expectations allow possibly
nonmeasurable suprema. For finite-rank covariances retain the full support;
zero-rank terms vanish. Taking \(J,K\) large, then \(\delta\downarrow0\),
and applying Markov's inequality to \eqref{eq:same-sample-score-factorization}
proves local stochastic equicontinuity:
\[
    \lim_{\delta\downarrow0}\limsup_{n\to\infty}
    \mathbb P^*\!\left(
      \sup_{(D,q)\in\mathcal A_R(\delta)}
      \|\mathbb G_n(\Gamma_{m_0+Dv,w_0+q}-\Gamma_0)\|>\epsilon
    \right)=0
    \quad(\epsilon>0).
\]
Norm tightness and \(L_2\) consistency place the fitted pair in these
localized classes with arbitrarily high probability. This gives the
\(o_{\mathbb P}(1)\) assertion; \eqref{eq:same-sample-von-mises} gives
the expansion. For known weights only the outcome term is needed.
\end{proof}

\begin{lemma}[Smooth parametric weights]
\label{lem:parametric-weight-equicontinuity}
Retain the identification, kernel, fixed-policy, weight-moment, and
outcome-operator conditions of Proposition~\ref{prop:same-sample-equicontinuity},
and suppose \(\|\widehat m-m_0\|_{L_2(P_{S,0};\mathcal H_Z)}=o_{\mathbb P}(1)\).
Replace its ratio-RKHS conditions by
\(\widehat w=w_{\widehat\theta}\), \(w_{\theta_0}=w_0\), where
\(\theta\in\mathbb R^p\), \(p\) is fixed, and
\(\|\widehat\theta-\theta_0\|_2=O_{\mathbb P}(n^{-1/2})\).
On a fixed open convex neighbourhood \(U\) of \(\theta_0\), assume
\(\theta\mapsto w_\theta(s)\) is twice continuously differentiable for
\(P_{S,0}\)-almost every \(s\), and that
\[
    \|\dot w_{\theta_0}\|_2\in L_2(P_{S,0}),\qquad
    B(s):=\sup_{\theta\in U}\|\ddot w_\theta(s)\|_{\mathrm{op}}
    \in L_2(P_{S,0}).
\]
Then, even when both nuisances use all observations,
\[
    \|\widehat w-w_0\|_{L_2(P_{S,0})}=O_{\mathbb P}(n^{-1/2}),
    \qquad
    \|\mathbb G_n(\Gamma_{\widehat\eta}-\Gamma_0)\|_{\mathcal H_Z}
    =o_{\mathbb P}(1),
\]
and
\[
    \widehat h_\pi-\chi(\pi)
    =(P_n-P_0)\Gamma_0+o_{\mathbb P}(n^{-1/2}).
\]
\end{lemma}

\begin{proof}
Put \(\Delta_n=\widehat\theta-\theta_0\). On the event
\(\widehat\theta\in U\), whose probability tends to one, Taylor's formula gives
\[
    q_n=\dot w_{\theta_0}^{\top}\Delta_n+R_n,
    \qquad |R_n|\leq\tfrac12 B\|\Delta_n\|_2^2.
\]
The \(L_2\) assumptions imply the stated ratio rate. For any fixed bounded
strongly measurable function \(c\) taking values in a separable Hilbert space,
\begin{align*}
    \|\mathbb G_n[q_nc]\|
    &\leq\|\Delta_n\|_2
      \sum_{j=1}^p\|\mathbb G_n[\dot w_{\theta_0,j}c]\|\\
    &\quad+\tfrac12\sqrt n\|\Delta_n\|_2^2\|c\|_\infty(P_nB+P_0B)
    =O_{\mathbb P}(n^{-1/2}).
\end{align*}
Here the fixed empirical processes have bounded second moments and
\(P_nB+P_0B=O_{\mathbb P}(1)\). These are pathwise bounds and do not
require independence between \(\widehat\theta\) and the observations.
Apply the bound to \(c=r_0\) and \(c=\varphi_S\) in
\eqref{eq:same-sample-score-factorization}. The outcome localization in
Proposition~\ref{prop:same-sample-equicontinuity} handles its first term,
while \(\|\widehat H-H_0\|_{\mathrm{op}}=O_{\mathbb P}(1)\) controls
the multiplier of the last term. Finally the ratio rate and outcome
consistency make the product remainder in \eqref{eq:same-sample-von-mises}
\(o_{\mathbb P}(n^{-1/2})\).
\end{proof}

A concrete sufficient model is Gaussian logging regression
\(A\mid X\sim\mathcal N(B_0z(X),\Sigma)\), with bounded regressors
\(z(X)\), nonsingular \(\mathbb E[z(X)z(X)^\top]\), and fixed known
\(\Sigma\succ0\). For a fixed target
\(\pi(\cdot\mid X)=\mathcal N(\mu_\pi(X),\Sigma)\) with bounded
\(\mu_\pi\), take \(\widehat B\) to be ordinary least squares and
\(w_B=\pi/p_B\). The regression coefficients are root-\(n\) consistent.
On a bounded neighbourhood of \(B_0\), the ratio and its first two
derivatives are bounded by a polynomial in \(\|A\|\) times
\(\exp(C\|A\|+C)\) for a finite \(C\). Gaussian tails therefore verify
the derivative envelopes and \(P_0w_0^2<\infty\) required above.

Lemma~\ref{lem:nystrom-operator-tightness} and
\eqref{eq:nystrom-cme-op-rate} verify the outcome conditions for the
projected estimator, including with landmarks selected from the same data.
For the weights, Proposition~\ref{prop:same-sample-equicontinuity}
gives RKHS sufficient conditions, and
Lemma~\ref{lem:parametric-weight-equicontinuity} gives a smooth parametric
alternative.
Also, \(\|\Gamma_0\|\leq\kappa_Z(1+2w_0)\) verifies the oracle second moment.
Removing the product remainder from \eqref{eq:same-sample-oracle-expansion}
additionally requires the product of nuisance errors to be
\(o_{\mathbb P}(n^{-1/2})\); the rate below needs only the
weaker \(O_{\mathbb P}(1)\) score control.

\begin{corollary}[Projected Nystr\"om DR embedding rate]
\label{cor:nystrom-dr-rate}
Suppose Corollary~\ref{cor:nystrom-explicit-schedule} holds and
\[
    \|\widehat w-w_0\|_{L_2(P_{S,0})}
    =
    O_{\mathbb P}(r_{w,n}).
\]
Assume further that
\[
    \mathbb E
    \|\Gamma_{m_0,w_0}(O)\|_{\mathcal H_Z}^2
    <
    \infty
\]
and that
\[
    \left\|
        (P_n-P_0)
        \left\{
            \Gamma_{\widehat m_{L_n,\lambda_n},\widehat w}
            -
            \Gamma_{m_0,w_0}
        \right\}
    \right\|_{\mathcal H_Z}
    =
    O_{\mathbb P}(n^{-1/2}).
\]
These two score conditions hold under the fixed-policy and ratio conditions
of Proposition~\ref{prop:same-sample-equicontinuity} or
Lemma~\ref{lem:parametric-weight-equicontinuity}, together with
Lemma~\ref{lem:nystrom-operator-tightness}; known weights require only
\(P_0w_0^2<\infty\).
Then
\begin{equation}
    \left\|
        \widehat\chi_{\mathrm{DR,Nys}}(\pi)
        -
        \chi(\pi)
    \right\|_{\mathcal H_Z}
    =
    O_{\mathbb P}
    \left(
        n^{-1/2}
        +
        r_{w,n}
        n^{-\frac{1}{2(1+\gamma)}}
    \right).
    \label{eq:nystrom-dr-rate}
\end{equation}
\end{corollary}

\begin{proof}
Apply \eqref{eq:same-sample-von-mises}, the Hilbert-valued Cauchy--Schwarz bound for the product remainder, and Corollary~\ref{cor:nystrom-explicit-schedule}.
\end{proof}

\paragraph{Finite target-policy Monte Carlo approximation.}
The preceding results concern the projected Nystr\"om doubly robust
estimator defined using the exact target-policy integral. We now control
the additional error induced by approximating this integral with
\(J\geq1\) conditionally independent target-policy draws at each logged
context, as in the continuous-action resampling construction of
\citet{zenati2025cpme}.

Write
\[
    \widehat m_n
    :=
    \widehat m_{L_n,\lambda_n},
    \qquad
    \widehat\chi_n
    :=
    \widehat\chi_{\mathrm{DR,Nys}}(\pi).
\]
Here, \(\widehat\chi_n\) denotes the theoretical projected Nystr\"om
estimator defined using the exact target-policy integral.

For every \(i=1,\ldots,n\), define the exact policy average
\begin{equation}
    \overline m_n(X_i)
    :=
    \int_{\gA}
    \widehat m_n(a,X_i)\,
    \pi(da\mid X_i),
    \label{eq:exact-fitted-policy-average}
\end{equation}
where the integral is a Bochner integral in \(\gH_Z\). Its existence is
ensured by the moment condition imposed in
Lemma~\ref{lem:finite-J-policy-integration}. The exact-integral estimator
can then be written as
\begin{align}
    \widehat\chi_n
    &=
    \frac1n
    \sum_{i=1}^n
    \Bigg[
        \widehat w(A_i,X_i)
        \left\{
            \varphi_Z(Z_i)
            -
            \widehat m_n(A_i,X_i)
        \right\}
        +
        \overline m_n(X_i)
    \Bigg].
    \label{eq:exact-integral-dr-target}
\end{align}

Let \(\mathcal F_n\) denote the sigma-field generated by all randomness
used to construct
\(\widehat m_n\), \(\widehat w\), and
\(\widehat\chi_n\) prior to drawing the target-policy Monte Carlo
actions. In particular, \(\mathcal F_n\) contains the logged sample and
the selected landmark set.

Conditional on \(\mathcal F_n\), draw
\begin{equation}
    A_{iu}^{\pi}
    \mid
    \mathcal F_n
    \sim
    \pi(\cdot\mid X_i),
    \qquad
    i=1,\ldots,n,
    \quad
    u=1,\ldots,J,
    \label{eq:conditional-target-policy-draws}
\end{equation}
independently across all pairs \((i,u)\). Define the Monte Carlo policy
average
\begin{equation}
    \overline m_{n,J}(X_i)
    :=
    \frac1J
    \sum_{u=1}^J
    \widehat m_n(A_{iu}^{\pi},X_i),
    \label{eq:finite-J-fitted-policy-average}
\end{equation}
and the corresponding finite-\(J\) estimator
\begin{align}
    \widetilde\chi_{n,J}
    &:=
    \frac1n
    \sum_{i=1}^n
    \Bigg[
        \widehat w(A_i,X_i)
        \left\{
            \varphi_Z(Z_i)
            -
            \widehat m_n(A_i,X_i)
        \right\}
        +
        \overline m_{n,J}(X_i)
    \Bigg].
    \label{eq:finite-J-dr-target}
\end{align}

The exact and finite-\(J\) estimators differ only through the
target-policy integration term:
\begin{equation}
    \widetilde\chi_{n,J}
    -
    \widehat\chi_n
    =
    \frac1n
    \sum_{i=1}^n
    \left\{
        \overline m_{n,J}(X_i)
        -
        \overline m_n(X_i)
    \right\}.
    \label{eq:finite-J-target-difference}
\end{equation}
Using the same target-policy draws and the same projected regression
subspace, \(\widetilde\chi_{n,J}\) coincides with
\[
    \widetilde\chi_{\mathrm{DR,Nys},J}(\pi)
\]
in the matrix notation of
Appendix~\ref{sec:projected_nystrom_computation}.

{
\begin{lemma}[Finite-\(J\) policy-integration error]
\label{lem:finite-J-policy-integration}
Assume that, almost surely, for every \(i=1,\ldots,n\), the map
\[
    a
    \longmapsto
    \widehat m_n(a,X_i)
\]
is strongly measurable and satisfies
\begin{equation}
    \int_{\gA}
    \left\|
        \widehat m_n(a,X_i)
    \right\|_{\gH_Z}^2
    \pi(da\mid X_i)
    <
    \infty.
    \label{eq:finite-J-square-integrability}
\end{equation}
Define the conditional average policy variance
\begin{equation}
    V_n
    :=
    \frac1n
    \sum_{i=1}^n
    \int_{\gA}
    \left\|
        \widehat m_n(a,X_i)
        -
        \overline m_n(X_i)
    \right\|_{\gH_Z}^2
    \pi(da\mid X_i).
    \label{eq:conditional-policy-variance}
\end{equation}
Then, almost surely,
\begin{equation}
    \E\!\left[
        \widetilde\chi_{n,J}
        \,\middle|\,
        \mathcal F_n
    \right]
    =
    \widehat\chi_n,
    \label{eq:finite-J-target-unbiasedness}
\end{equation}
and
\begin{equation}
    \E\!\left[
        \left\|
            \widetilde\chi_{n,J}
            -
            \widehat\chi_n
        \right\|_{\gH_Z}^2
        \,\middle|\,
        \mathcal F_n
    \right]
    =
    \frac{V_n}{nJ}.
    \label{eq:finite-J-target-mse}
\end{equation}
Consequently, for every \(\delta\in(0,1)\),
\begin{equation}
    \Pr\!\left(
        \left.
        \left\|
            \widetilde\chi_{n,J}
            -
            \widehat\chi_n
        \right\|_{\gH_Z}
        >
        \sqrt{
            \frac{V_n}
                 {nJ\delta}
        }
        \,\right|\,
        \mathcal F_n
    \right)
    \leq
    \delta.
    \label{eq:finite-J-target-high-probability}
\end{equation}
\end{lemma}

\begin{proof}
For \(i=1,\ldots,n\) and \(u=1,\ldots,J\), define
\[
    \xi_{iu}
    :=
    \widehat m_n(A_{iu}^{\pi},X_i)
    -
    \overline m_n(X_i).
\]
By the definition of the exact policy average,
\(
    \E\!\left[
        \xi_{iu}
        \,\middle|\,
        \mathcal F_n
    \right]
    =
    0.
\)
Moreover, conditionally on \(\mathcal F_n\), the collection
\(
    \{\xi_{iu}:i=1,\ldots,n,\ u=1,\ldots,J\}
\)
is mutually independent.

Using
\eqref{eq:finite-J-target-difference}, we have
\begin{equation}
    \widetilde\chi_{n,J}
    -
    \widehat\chi_n
    =
    \frac1{nJ}
    \sum_{i=1}^n
    \sum_{u=1}^J
    \xi_{iu}.
    \label{eq:finite-J-centered-sum}
\end{equation}
Conditional centering immediately gives
\eqref{eq:finite-J-target-unbiasedness}.

For two distinct index pairs
\((i,u)\neq(i',u')\), conditional independence and centering imply
\[
    \E\!\left[
        \left\langle
            \xi_{iu},
            \xi_{i'u'}
        \right\rangle_{\gH_Z}
        \,\middle|\,
        \mathcal F_n
    \right]
    =
    0.
\]
Expanding the squared norm in
\eqref{eq:finite-J-centered-sum} therefore yields
\begin{align*}
\E\!\left[
    \left\|
        \widetilde\chi_{n,J}
        -
        \widehat\chi_n
    \right\|_{\gH_Z}^2
    \,\middle|\,
    \mathcal F_n
\right]
&=
\frac1{n^2J^2}
\sum_{i=1}^n
\sum_{u=1}^J
\E\!\left[
    \left\|
        \xi_{iu}
    \right\|_{\gH_Z}^2
    \,\middle|\,
    \mathcal F_n
\right]
\\
&\quad=
\frac1{n^2J}
\sum_{i=1}^n
\int_{\gA}
\left\|
    \widehat m_n(a,X_i)
    -
    \overline m_n(X_i)
\right\|_{\gH_Z}^2
\pi(da\mid X_i)
\\
&\quad=
\frac{V_n}{nJ}.
\end{align*}
This proves
\eqref{eq:finite-J-target-mse}. Finally,
\eqref{eq:finite-J-target-high-probability} follows by applying
conditional Markov's inequality to
\[
    \left\|
        \widetilde\chi_{n,J}
        -
        \widehat\chi_n
    \right\|_{\gH_Z}^2.
\]
\end{proof}
}

{
\begin{corollary}[Finite-\(J\) projected Nystr\"om DR rate]
\label{cor:finite-J-nystrom-dr-rate}
Suppose the conditions of
Corollary~\ref{cor:nystrom-dr-rate} hold, and let
\((J_n)_{n\geq1}\) be a deterministic sequence of positive integers.

\begin{enumerate}[label=(\roman*)]

\item
Under the bounded-kernel and ridge assumptions already imposed,
\begin{equation}
    \left\|
        \widetilde\chi_{n,J_n}
        -
        \widehat\chi_n
    \right\|_{\gH_Z}
    =
    O_{\mathbb P}
    \left(
        \frac1{\sqrt{nJ_n\lambda_n}}
    \right).
    \label{eq:finite-J-conservative-rate}
\end{equation}
Consequently,
\begin{align}
    \left\|
        \widetilde\chi_{n,J_n}
        -
        \chi(\pi)
    \right\|_{\gH_Z}
    &=
    O_{\mathbb P}
    \Bigg(
        n^{-1/2}
        +
        r_{w,n}
        n^{-\frac1{2(1+\gamma)}}
        +
        \frac1{\sqrt{nJ_n\lambda_n}}
    \Bigg).
    \label{eq:finite-J-conservative-target-rate}
\end{align}
For
\(
    \lambda_n
    =
    n^{-1/(1+\gamma)},
\)
the Monte Carlo term becomes
\begin{equation}
    \frac1{\sqrt{nJ_n\lambda_n}}
    =
    J_n^{-1/2}
    n^{-\frac{\gamma}{2(1+\gamma)}}.
    \label{eq:finite-J-conservative-expanded}
\end{equation}

\item
If, in addition,
\begin{equation}
    V_n
    =
    O_{\mathbb P}(1),
    \label{eq:policy-variance-tightness}
\end{equation}
then
\begin{equation}
    \left\|
        \widetilde\chi_{n,J_n}
        -
        \widehat\chi_n
    \right\|_{\gH_Z}
    =
    O_{\mathbb P}
    \left(
        (nJ_n)^{-1/2}
    \right).
    \label{eq:finite-J-policy-rate}
\end{equation}
Consequently,
\begin{align}
    \left\|
        \widetilde\chi_{n,J_n}
        -
        \chi(\pi)
    \right\|_{\gH_Z}
    &=
    O_{\mathbb P}
    \left(
        n^{-1/2}
        +
        r_{w,n}
        n^{-\frac1{2(1+\gamma)}}
        +
        (nJ_n)^{-1/2}
    \right).
    \label{eq:finite-J-combined-target-rate}
\end{align}
Since \(J_n\geq1\),
\begin{equation}
    \left\|
        \widetilde\chi_{n,J_n}
        -
        \chi(\pi)
    \right\|_{\gH_Z}
    =
    O_{\mathbb P}
    \left(
        n^{-1/2}
        +
        r_{w,n}
        n^{-\frac1{2(1+\gamma)}}
    \right).
    \label{eq:finite-J-absorbed-target-rate}
\end{equation}
Thus, under
\eqref{eq:policy-variance-tightness}, even a fixed number of
target-policy draws per logged context does not change the
exact-policy-integral statistical rate.

\end{enumerate}
\end{corollary}

\begin{proof}
Because the zero operator is feasible in the projected ridge problem,
optimality of
\(\widehat H_{L_n,\lambda_n}\) gives
\begin{align}
    \lambda_n
    \left\|
        \widehat H_{L_n,\lambda_n}
    \right\|_{\mathrm{HS}}^2
    &\leq
    \frac1n
    \sum_{i=1}^n
    \left\|
        \varphi_Z(Z_i)
    \right\|_{\gH_Z}^2
    \leq
    \kappa_Z^2.
    \label{eq:projected-ridge-operator-bound}
\end{align}
Hence, for every \(s\in\gS\),
\begin{align*}
    \left\|
        \widehat m_n(s)
    \right\|_{\gH_Z}
    &=
    \left\|
        \widehat H_{L_n,\lambda_n}
        \varphi_S(s)
    \right\|_{\gH_Z}
    \leq
    \left\|
        \widehat H_{L_n,\lambda_n}
    \right\|_{\mathrm{HS}}
    \left\|
        \varphi_S(s)
    \right\|_{\gH_S}
    \leq
    \frac{
        \kappa_S\kappa_Z
    }{
        \sqrt{\lambda_n}
    }.
\end{align*}
Using the Hilbert-space variance identity,
\[
    \int_{\gA}
    \left\|
        \widehat m_n(a,X_i)
        -
        \overline m_n(X_i)
    \right\|_{\gH_Z}^2
    \pi(da\mid X_i)
    \leq
    \int_{\gA}
    \left\|
        \widehat m_n(a,X_i)
    \right\|_{\gH_Z}^2
    \pi(da\mid X_i),
\]
and therefore
\begin{equation}
    V_n
    \leq
    \frac{
        \kappa_S^2\kappa_Z^2
    }{
        \lambda_n
    }.
    \label{eq:conservative-policy-variance-bound}
\end{equation}
Lemma~\ref{lem:finite-J-policy-integration} now gives
\eqref{eq:finite-J-conservative-rate}. Combining it with
Corollary~\ref{cor:nystrom-dr-rate} and the triangle inequality
gives
\eqref{eq:finite-J-conservative-target-rate}.

Under
\eqref{eq:policy-variance-tightness},
Lemma~\ref{lem:finite-J-policy-integration} instead gives
\eqref{eq:finite-J-policy-rate}. Combining this result with
Corollary~\ref{cor:nystrom-dr-rate} gives
\eqref{eq:finite-J-combined-target-rate}. Finally,
\[
    (nJ_n)^{-1/2}
    \leq
    n^{-1/2}
\]
because \(J_n\geq1\), proving \eqref{eq:finite-J-absorbed-target-rate}.
\end{proof}
}

\begin{remark}[Choice of the policy Monte Carlo budget]
\label{rem:finite-J-budget}
Under the conservative bound \eqref{eq:finite-J-conservative-expanded}, a fixed \(J_n\) still gives a consistent policy-integration approximation because
\[
    nJ_n\lambda_n
    =
    J_n
    n^{\frac{\gamma}{1+\gamma}}
    \longrightarrow
    \infty.
\]
To make the conservative Monte Carlo error no larger than the projected conditional-mean-embedding rate
\(n^{-1/\{2(1+\gamma)\}}\), it is sufficient to choose
\begin{equation}
    J_n
    \gtrsim
    n^{\frac{1-\gamma}{1+\gamma}}.
    \label{eq:finite-J-rate-preserving-budget}
\end{equation}
Under the stronger variance-tightness condition \eqref{eq:policy-variance-tightness}, no growth of \(J_n\) is required to preserve the statistical rate.
\end{remark}

\subsection{Consistency of the generated particles}
\label{sec:estimated-target-flow-proof}

Throughout this section, write $\chi(\pi)=\mu_{Q^\pi}$ for the true interventional mean embedding, $\widehat h_\pi$ for the exact-integral estimator, and $\widetilde h_\pi$ for the numerical target used by the particle update. Recall that 
$e_{n,J}:=\|\widetilde h_\pi-\chi(\pi)\|_{\mathcal H}$ and $\widehat Q_M^{(r)}:=\frac1M\sum_{j=1}^M\delta_{Z_j^{(r)}}$. 

\textcolor{black}{
Throughout this subsection, the particle bounds are
understood conditionally on the target-norm event
\[
\mathcal E_n
:=
\bigl\{\|\widetilde h_\pi\|_{\mathcal H}
\le \kappa\bigr\}.
\]
}

For $h\in\mathcal H$, define
\[
s_h(z):=\langle h,\varphi(z)\rangle_{\mathcal H},
\qquad
v_h(Q,z)=\nabla s_h(z)-\int\nabla_1k(z,z')\,Q(dz').
\]
Let $\mathcal D$ contain the observations and all target-construction randomness, and let $\mathcal F_r$ additionally contain the initial particles and the Gaussian perturbations before iteration $r$. The update is
\[
Z_j^{(r+1)}=Z_j^{(r)}+
\tau v_{\widetilde h_\pi}\!\left(\widehat Q_M^{(r)},Z_j^{(r)}+\beta_rU_j^{(r)}\right),
\qquad j=1,\ldots,M,
\]
where, conditional on $\mathcal F_r$, the vectors $U_j^{(r)}$ are independent standard Gaussian vectors in $\mathbb R^{d_Z}$.

\begin{assumption}[Smooth particle dynamics]\label{asst:dr_kernel}
The feature map $\varphi$ is twice continuously Fr\'echet differentiable into $\mathcal H$. Its norm and the operator norms of its first two derivatives are uniformly bounded by $\kappa\ge1$. The update uses a constant step size $\eta_r\equiv\tau>0$ and noise levels $\beta_r=c_\beta(r+1)^{-\alpha}$, where
\[
12\tau\kappa^2d_Z\le1,
\qquad 0<c_\beta\le\frac1{2\sqrt{d_Z}},
\qquad 0\le\alpha<\frac12.
\]
\end{assumption}

\begin{assumption}[Noisy gradient dominance]\label{asst:dr_noise}
For $r=0,\ldots,R-1$, the particles generated with target $\widetilde h_\pi$ satisfy, almost surely,
\[
\frac1M\sum_{j=1}^M\mathbb E\!\left[
\left\|v_{\chi(\pi)}\!\left(\widehat Q_M^{(r)},Z_j^{(r)}+\beta_rU_j^{(r)}\right)\right\|_2^2
\,\middle|\,\mathcal F_r\right]
\ge4\kappa^2d_Z\beta_r^2\operatorname{MMD}_k^2(\widehat Q_M^{(r)},Q^\pi).
\]
\end{assumption}

\begin{lemma}[Descent lemma in expectation]\label{lem:dr_descent}
Under Assumptions~\ref{asst:dr_kernel} and~\ref{asst:dr_noise},
\begin{align*}
\mathbb E\!\left[\left\|\mu_{\widehat Q_M^{(r+1)}}-\widetilde h_\pi\right\|_{\mathcal H}^2\,\middle|\,\mathcal F_r\right]
&\le\left(1-\frac{\tau\kappa^2d_Z}{8}\beta_r^2\right)
\left\|\mu_{\widehat Q_M^{(r)}}-\widetilde h_\pi\right\|_{\mathcal H}^2
+\tau\kappa^2\left(\frac92d_Z\beta_r^2+\frac34\right)e_{n,J}^2.
\end{align*}
In particular, with
$a:=\frac{\tau\kappa^2d_Zc_\beta^2}{8}$, $B:=\tau\kappa^2\left(\frac92d_Zc_\beta^2+\frac34\right)$, and $ a_r:=a(r+1)^{-2\alpha}$,
the right-hand side is at most $(1-a_r) \left\|\mu_{\widehat Q_M^{(r)}}-\widetilde h_\pi\right\|_{\mathcal H}^2+Be_{n,J}^2$. 
\end{lemma}

\begin{lemma}[Concentration of particles after noise injection]\label{lem:dr_concentration}
Under Assumption~\ref{asst:dr_kernel}, for any $q>0$, conditional on $\mathcal F_r$, with probability at least $1-e^{-q}$,
\begin{align*}
\left\|\mu_{\widehat Q_M^{(r+1)}}-\widetilde h_\pi\right\|_{\mathcal H}
-\mathbb E\!\left[\left\|\mu_{\widehat Q_M^{(r+1)}}-\widetilde h_\pi\right\|_{\mathcal H}\,\middle|\,\mathcal F_r\right]
\le 2\tau\kappa^3\beta_r\sqrt{\frac{2q}{M}}.
\end{align*}
\end{lemma}

\begin{proof}[Proof of Theorem~\ref{thm:dr_flow}]
Condition on $\mathcal D$. The feature bound and
$\|\widetilde h_\pi\|_{\mathcal H}\le\kappa$ imply
$\|\mu_{\widehat Q_M^{(r)}}-\widetilde h_\pi\|_{\mathcal H}\le2\kappa$
for every particle configuration. Let $a$, $B$, and $a_r$ be as in
Lemma~\ref{lem:dr_descent}, so that $0<a_r<1$.

Fix $\delta\in(0,1)$. Applying Lemma~\ref{lem:dr_concentration}
with $q=\log(R/\delta)$, together with Jensen's inequality,
Lemma~\ref{lem:dr_descent}, and a conditional union bound, shows
that, with conditional probability at least $1-\delta$,
simultaneously for $r=0,\ldots,R-1$,
\[
\|\mu_{\widehat Q_M^{(r+1)}}-\widetilde h_\pi\|_{\mathcal H}
\le
\sqrt{(1-a_r)
\|\mu_{\widehat Q_M^{(r)}}-\widetilde h_\pi\|_{\mathcal H}^2
+Be_{n,J}^2}
+2\sqrt2\,\tau\kappa^3c_\beta(r+1)^{-\alpha}
\sqrt{\frac{\log(R/\delta)}{M}}.
\]
Work on this event. The map
$x\mapsto\sqrt{(1-a_r)x^2+Be_{n,J}^2}$
is increasing and $\sqrt{1-a_r}$-Lipschitz on $[0,\infty)$.
Consequently, iterating the recursion and propagating each
additive concentration term through these Lipschitz maps gives
\begin{align*}
\|\mu_{\widehat Q_M^{(R)}}-\widetilde h_\pi\|_{\mathcal H}
&\le
\left[
\|\mu_{\widehat Q_M^{(0)}}-\widetilde h_\pi\|_{\mathcal H}^2
\prod_{r=0}^{R-1}(1-a_r)
+Be_{n,J}^2
\sum_{r=0}^{R-1}\prod_{\ell=r+1}^{R-1}(1-a_\ell)
\right]^{1/2}\\
&\quad+
2\sqrt2\,\tau\kappa^3c_\beta
\sqrt{\frac{\log(R/\delta)}{M}}
\sum_{r=0}^{R-1}(r+1)^{-\alpha}
\prod_{\ell=r+1}^{R-1}\sqrt{1-a_\ell}.
\end{align*}
Since $a_\ell\ge aR^{-2\alpha}$ for $\ell<R$, a geometric-series
bound gives
$\sum_{r=0}^{R-1}\prod_{\ell=r+1}^{R-1}(1-a_\ell)
\le R^{2\alpha}/a$.
Moreover, using $\sqrt{1-x}\le e^{-x/2}$ and
$\sum_{r=0}^{R-1}(r+1)^{-2\alpha}
\ge\int_1^{R+1}x^{-2\alpha}\,dx$, we obtain
$\prod_{r=0}^{R-1}\sqrt{1-a_r}
\le\exp[-a\{(R+1)^{1-2\alpha}-1\}/\{2(1-2\alpha)\}]$.
For the concentration term, the inequalities
$(r+1)^{-\alpha}\le R^\alpha a_r/a$ and
$a_r\le2(1-\sqrt{1-a_r})$ yield, by telescoping,
$\sum_{r=0}^{R-1}(r+1)^{-\alpha}
\prod_{\ell=r+1}^{R-1}\sqrt{1-a_\ell}
\le2R^\alpha/a$.

Combining these estimates with $\sqrt{x+y}\le\sqrt{x}+\sqrt{y}$,
$\|\mu_{\widehat Q_M^{(0)}}-\widetilde h_\pi\|_{\mathcal H}\le2\kappa$,
and the triangle inequality
$\operatorname{MMD}_k(\widehat Q_M^{(R)},Q^\pi)
\le\|\mu_{\widehat Q_M^{(R)}}-\widetilde h_\pi\|_{\mathcal H}+e_{n,J}$
gives
\[
\operatorname{MMD}_k(\widehat Q_M^{(R)},Q^\pi)
\le C_0e^{-cR^{1-2\alpha}}
+(1+C_1R^\alpha)e_{n,J}
+C_2R^\alpha\sqrt{\frac{\log(R/\delta)}{M}},
\]
where one may take $c=a/\{2(1-2\alpha)\}$,
$C_0=2\kappa e^c$, $C_1=\sqrt{B/a}$, and
$C_2=4\sqrt2\,\tau\kappa^3c_\beta/a$.

\end{proof}

\begin{remark}[Relation to the stationary MMD points result]
The auxiliary RKHS Lipschitz argument and discrete recursion techniques from the stationary MMD points analysis~\citep{chen2025stationary} apply here. Its descent lemma cannot be applied unchanged: the particle update uses $\widetilde h_\pi$, whereas the noisy gradient-dominance assumption uses the true target $Q^\pi$. Their difference contributes the $e_{n,J}^2$ term in Lemma~\ref{lem:dr_descent}. The concentration step is stated for the updated particles in Lemma~\ref{lem:dr_concentration}. Solving the squared recursion before taking a square root gives a propagated target error of order $R^\alpha e_{n,J}$ and a finite-particle error of order $R^\alpha\sqrt{\log(R/\delta)/M}$. 
\end{remark}

\begin{proof}[Proof of Lemma~\ref{lem:dr_descent}]
Fix $r$ and condition on $\mathcal F_r$. The derivative bounds imply
\[
\begin{aligned}
&\|v_{\widetilde h_\pi}(\widehat Q_M^{(r)},z)
-v_{\widetilde h_\pi}(\widehat Q_M^{(r)},z')\|_2
\le\kappa
\|\mu_{\widehat Q_M^{(r)}}-\widetilde h_\pi\|_{\mathcal H}
\|z-z'\|_2,\\
&\|v_{\widetilde h_\pi}(\widehat Q_M^{(r)},z)
-v_{\chi(\pi)}(\widehat Q_M^{(r)},z)\|_2
\le\kappa e_{n,J}.
\end{aligned}
\]
These bounds depend on the target only through its RKHS residual
and do not require $\widetilde h_\pi$ to be a probability embedding.

The gradient of
$\|M^{-1}\sum_{j=1}^M\varphi(z_j)-\widetilde h_\pi\|_{\mathcal H}^2$
with respect to its $j$th particle, evaluated at the current
configuration, is
$-2v_{\widetilde h_\pi}(\widehat Q_M^{(r)},Z_j^{(r)})/M$.
Since the discrepancy is bounded by $2\kappa$ for every configuration
and both feature derivatives are bounded by $\kappa$, its second
directional derivative along $u_{1:M}$ is bounded in absolute value
by $6\kappa^2M^{-1}\sum_{j=1}^M\|u_j\|_2^2$.
Taylor's theorem along the particle update therefore gives
\begin{align*}
\|\mu_{\widehat Q_M^{(r+1)}}-\widetilde h_\pi\|_{\mathcal H}^2
&\le
\|\mu_{\widehat Q_M^{(r)}}-\widetilde h_\pi\|_{\mathcal H}^2 - \frac{2\tau}{M}\sum_{j=1}^M
\left\langle
v_{\widetilde h_\pi}(\widehat Q_M^{(r)},Z_j^{(r)}),
v_{\widetilde h_\pi}(\widehat Q_M^{(r)},
Z_j^{(r)}+\beta_rU_j^{(r)})
\right\rangle\\
&\quad+\frac{3\tau^2\kappa^2}{M}\sum_{j=1}^M
\left\|v_{\widetilde h_\pi}(\widehat Q_M^{(r)},
Z_j^{(r)}+\beta_rU_j^{(r)})\right\|_2^2.
\end{align*}
By the first Lipschitz bound and
$\mathbb E[\|U_j^{(r)}\|_2^2\mid\mathcal F_r]=d_Z$,
the conditional mean squared difference between the perturbed
and unperturbed velocities is at most
$\kappa^2d_Z\beta_r^2
\|\mu_{\widehat Q_M^{(r)}}-\widetilde h_\pi\|_{\mathcal H}^2$.
Using $2\langle a,b\rangle=\|a\|_2^2+\|b\|_2^2-\|a-b\|_2^2$
and $3\tau\kappa^2\le1/4$, and discarding the nonpositive
unperturbed-velocity term, yields
\begin{align*}
\mathbb E\!\left[
\|\mu_{\widehat Q_M^{(r+1)}}-\widetilde h_\pi\|_{\mathcal H}^2
\,\middle|\,\mathcal F_r\right]
\le(1+\tau\kappa^2d_Z\beta_r^2)
\|\mu_{\widehat Q_M^{(r)}}-\widetilde h_\pi\|_{\mathcal H}^2 -\frac{3\tau}{4M}\sum_{j=1}^M
\mathbb E\!\left[
\left\|v_{\widetilde h_\pi}(\widehat Q_M^{(r)},
Z_j^{(r)}+\beta_rU_j^{(r)})\right\|_2^2
\,\middle|\,\mathcal F_r\right].
\end{align*}
The velocity perturbation bound, the inequality
$\|a+b\|_2^2\ge\frac12\|a\|_2^2-\|b\|_2^2$,
and Assumption~\ref{asst:dr_noise} imply
\begin{align*}
\frac1M\sum_{j=1}^M
\mathbb E\!\left[
\left\|v_{\widetilde h_\pi}(\widehat Q_M^{(r)},
Z_j^{(r)}+\beta_rU_j^{(r)})\right\|_2^2
\,\middle|\,\mathcal F_r\right]
&\ge2\kappa^2d_Z\beta_r^2
\operatorname{MMD}_k^2(\widehat Q_M^{(r)},Q^\pi)
-\kappa^2e_{n,J}^2\\
&\ge\frac32\kappa^2d_Z\beta_r^2
\|\mu_{\widehat Q_M^{(r)}}-\widetilde h_\pi\|_{\mathcal H}^2
-(6\kappa^2d_Z\beta_r^2+\kappa^2)e_{n,J}^2,
\end{align*}
where the second inequality uses
$\operatorname{MMD}_k^2(\widehat Q_M^{(r)},Q^\pi)
\ge\frac34
\|\mu_{\widehat Q_M^{(r)}}-\widetilde h_\pi\|_{\mathcal H}^2
-3e_{n,J}^2$.
Substituting into the preceding descent inequality gives
\begin{align*}
\mathbb E\!\left[
\|\mu_{\widehat Q_M^{(r+1)}}-\widetilde h_\pi\|_{\mathcal H}^2
\,\middle|\,\mathcal F_r\right]
&\le
\left(1-\frac{\tau\kappa^2d_Z}{8}\beta_r^2\right)
\|\mu_{\widehat Q_M^{(r)}}-\widetilde h_\pi\|_{\mathcal H}^2 + 
\tau\kappa^2\left(\frac92d_Z\beta_r^2+\frac34\right)e_{n,J}^2.
\end{align*}
Finally, $\beta_r^2=c_\beta^2(r+1)^{-2\alpha}\le c_\beta^2$
gives the stated bound with $a_r$ and $B$.
\end{proof}

\begin{proof}[Proof of Lemma~\ref{lem:dr_concentration}]
Condition on $\mathcal F_r$. The velocity field
$v_{\widetilde h_\pi}(\widehat Q_M^{(r)},\cdot)$ is
$\kappa\|\mu_{\widehat Q_M^{(r)}}-\widetilde h_\pi\|_{\mathcal H}$-Lipschitz,
and $\varphi$ is $\kappa$-Lipschitz. Each updated particle depends
only on its own Gaussian perturbation, with the current empirical
law and target held fixed. Applying these Lipschitz bounds,
averaging over particles, and using Cauchy--Schwarz shows that,
as a function of $U_{1:M}^{(r)}$, the next-step discrepancy
$\|\mu_{\widehat Q_M^{(r+1)}}-\widetilde h_\pi\|_{\mathcal H}$
is Lipschitz with constant at most
$\tau\kappa^2\beta_r
\|\mu_{\widehat Q_M^{(r)}}-\widetilde h_\pi\|_{\mathcal H}/\sqrt M
\le2\tau\kappa^3\beta_r/\sqrt M$.
Gaussian concentration therefore implies that, for any $q>0$,
with conditional probability at least $1-e^{-q}$,
\[
\|\mu_{\widehat Q_M^{(r+1)}}-\widetilde h_\pi\|_{\mathcal H}
-\mathbb E\!\left[
\|\mu_{\widehat Q_M^{(r+1)}}-\widetilde h_\pi\|_{\mathcal H}
\,\middle|\,\mathcal F_r\right]
\le2\tau\kappa^3\beta_r\sqrt{\frac{2q}{M}}.
\]
This proves the claim.
\end{proof}

\subsection{Numerical-target rate and consistency}
\begin{proof}[Proof of Proposition~\ref{prop:sg-target-rate}]
Apply the exact expansion \eqref{eq:same-sample-von-mises} to the
exact-integral estimator. Its oracle term is $O_p(n^{-1/2})$ by the finite
second moment, its empirical remainder has that order by
Assumption~\ref{ass:sg-statistical-target}, and its product remainder is
$O_p(r_{m,n}r_{w,n})$.
Conditional on the logged data, landmarks, and fitted nuisances,
Lemma~\ref{lem:finite-J-policy-integration} gives mean-square error
$V_n/(nJ)$ for $\widetilde h_\pi-\widehat h_\pi$.
Conditional Markov's inequality and $V_n=O_p(1)$ imply
$\|\widetilde h_\pi-\widehat h_\pi\|_{\gH}=O_p((nJ)^{-1/2})$.
The triangle inequality proves the claim. This last argument applies to
the same fitted regression in both targets, with no additional numerical
or subspace discrepancy.
\end{proof}


\begin{proof}[Proof of Corollary~\ref{cor:sg-sampling-consistency}]
Define
\[
    \lambda_{\mathrm{opt}}
    :=
    \frac{
        \tau\kappa^2d_Zc_\beta^2
    }{
        16(1-2\alpha)
    }.
\]
By the definition of \(C_R\) in
Corollary~\ref{cor:sg-sampling-consistency},
\(
    \lambda_{\mathrm{opt}}C_R
    =
    s.
\)

We first control the numerical target error. Since \(J\geq1\) is fixed,
\(s\in(0,1/2]\), and
\(
    r_{m,n}r_{w,n}
    =
    O(n^{-s}),
\)
Proposition~\ref{prop:sg-target-rate} gives
\begin{align*}
    e_{n,J}
    &=
    O_p\left(
        n^{-1/2}
        +
        r_{m,n}r_{w,n}
        +
        (nJ)^{-1/2}
    \right)
    =
    O_p(n^{-s})
    =
    o_p(1).
\end{align*}

We next verify the target-norm event required by
Theorem~\ref{thm:dr_flow}. Let
\[
    \Delta_\kappa
    :=
    \kappa
    -
    \sup_{z\in\gZ}
    \|\varphi(z)\|_{\gH}
    >0
\]
and define
\[
    \mathcal E_n
    :=
    \left\{
        \|\widetilde h_\pi\|_{\gH}
        \leq
        \kappa
    \right\}.
\]
Since
\[
    \|\chi(\pi)\|_{\gH}
    =
    \left\|
        \int_{\gZ}
        \varphi(z)\,Q^\pi(dz)
    \right\|_{\gH}
    \leq
    \sup_{z\in\gZ}
    \|\varphi(z)\|_{\gH},
\]
the triangle inequality gives
\begin{align}\label{eq:tilde_h_bound}
    \|\widetilde h_\pi\|_{\gH}
    \leq
    \sup_{z\in\gZ}
    \|\varphi(z)\|_{\gH}
    +
    e_{n,J}.
\end{align}
Consequently,
\[
    \Pr(\mathcal E_n^c)
    \leq
    \Pr(e_{n,J}>\Delta_\kappa)
    \longrightarrow
    0.
\]

Fix \(\delta\in(0,1)\). On \(\mathcal E_n\), apply Theorem~\ref{thm:dr_flow} with \(R=R_n\) and \(M=M_n\). By Equation \ref{eq:estimated-target-particle-bound}, with conditional probability at least \(1-\delta\),
\begin{align}
&
\operatorname{MMD}_k
\left(
    \widehat Q_{M_n}^{(R_n)},
    Q^\pi
\right)
\leq
C_0
\exp\left\{
    -\lambda_{\mathrm{opt}}
     R_n^{1-2\alpha}
\right\}
+
\left(
    1+C_1R_n^\alpha
\right)e_{n,J}
+
C_2R_n^\alpha
\sqrt{
    \frac{\log(R_n/\delta)}
         {M_n}
}.
\label{eq:corollary-particle-bound}
\end{align}

We now control the three terms in Equation \ref{eq:corollary-particle-bound}. By Equation
\ref{eq:sg-joint-schedule},
\[
    R_n
    \geq
    (C_R\log n)^{1/(1-2\alpha)}.
\]
Therefore,
\begin{align*}
\exp\left\{
    -\lambda_{\mathrm{opt}}
     R_n^{1-2\alpha}
\right\}
&\leq
\exp\left\{
    -\lambda_{\mathrm{opt}}
     C_R\log n
\right\}
=
\exp\{-s\log n\}
=
n^{-s}.
\end{align*}

Moreover,
\(
    R_n^\alpha
    =
    O\left(
        (\log n)^{\alpha/(1-2\alpha)}
    \right).
\)
Combining this with \(e_{n,J}=O_p(n^{-s})\) yields
\[
    \left(
        1+C_1R_n^\alpha
    \right)e_{n,J}
    =
    O_p\left(
        n^{-s}
        (\log n)^{\alpha/(1-2\alpha)}
    \right).
\]

For fixed \(\delta\in(0,1)\),
\(
    \log(R_n/\delta)
    =
    O(\log\log n).
\)
Furthermore, for all sufficiently large \(n\),
\[
    M_n
    \asymp
    n^{2s}\log\log n.
\]
It follows that
\begin{align*}
R_n^\alpha
\sqrt{
    \frac{\log(R_n/\delta)}
         {M_n}
}
&=
O\left(
    (\log n)^{\alpha/(1-2\alpha)}
    \sqrt{
        \frac{\log\log n}
             {n^{2s}\log\log n}
    }
\right)
=
O\left(
    n^{-s}
    (\log n)^{\alpha/(1-2\alpha)}
\right).
\end{align*}

To pass from the conditional statement to an unconditional one, let
\(\mathcal G_n(\delta)\) denote the event on which Equation \ref{eq:corollary-particle-bound} holds. Since \(\mathcal E_n\) is measurable with respect to the observations and target-construction randomness,
\begin{align*}
    \Pr\!\left(
        \mathcal G_n(\delta)^c
    \right)
    &\leq
    \Pr(\mathcal E_n^c)
    +
    \E\!\left[
        \mathbf 1_{\mathcal E_n}
        \Pr\!\left(
            \mathcal G_n(\delta)^c
            \mid
            \mathcal D
        \right)
    \right]
    \leq
    o(1)+\delta.
\end{align*}
Since \(\delta\in(0,1)\) is arbitrary, the preceding bounds imply
\[
    \operatorname{MMD}_k
    \left(
        \widehat Q_{M_n}^{(R_n)},
        Q^\pi
    \right)
    =
    O_p\left(
        n^{-s}
        (\log n)^{\alpha/(1-2\alpha)}
    \right).
\]

In particular, the right-hand side converges to zero. If
\(\operatorname{MMD}_k\) metrizes weak convergence, then
\(
    \widehat Q_{M_n}^{(R_n)}
    \Rightarrow
    Q^\pi
\)
in probability with respect to the weak topology.
\end{proof}

\subsection{Outcome-space transfer through the frozen encoder--decoder}

We now prove Proposition~\ref{thm:EncoderPullbackMMDTransfer} to isolate the error induced by the mapping of the structured outcomes into the encoder's latent space. 

\begin{proof}[Proof of Proposition~\ref{thm:EncoderPullbackMMDTransfer}]
First, $k_E$ is positive definite. Indeed, for arbitrary $m\in\mathbb N$, points $y_1,\ldots,y_m\in\gY$, and coefficients $c_1,\ldots,c_m\in\mathbb R$,
\begin{align*}
\sum_{i=1}^m\sum_{\ell=1}^m c_i c_\ell k_E(y_i,y_\ell)
&=
\sum_{i=1}^m\sum_{\ell=1}^m c_i c_\ell
\langle\varphi(E(y_i)),\varphi(E(y_\ell))\rangle_{\gH_k}\\
&=
\left\|\sum_{i=1}^m c_i\varphi(E(y_i))\right\|_{\gH_k}^2
\geq 0.
\end{align*}
Thus, $y\mapsto\varphi(E(y))$ is a feature map of $k_E$.

Let $Q_1,Q_2$ be  probability laws on $\gZ$ satisfying
$\E_{Z\sim Q_j}\|E(D(Z))-Z\|_2<\infty$, $j\in\{1,2\}$
Set $T=E\circ D$. For $j\in\{1,2\}$, define
\[
\mu_j=\E_{Z\sim Q_j}[\varphi(Z)]
\qquad\text{and}\qquad
\widetilde\mu_j=\E_{Z\sim Q_j}[\varphi(T(Z))].
\]

The definition of a pushforward measure implies that, for every integrable function $f$,
$\int f(u)\,(T_\#Q_j)(du)=\int f(T(z))\,Q_j(dz)$.
Applying this identity to the feature map gives
\begin{align}
\operatorname{MMD}_{k_E}(D_\#Q_1,D_\#Q_2)
&=\|\widetilde\mu_1-\widetilde\mu_2\|_{\gH_k}
\notag\\
&=\operatorname{MMD}_{k}(T_\#Q_1,T_\#Q_2).
\label{eq:encoder-pullback-exact-identity}
\end{align}
This is an exact identity, not an inequality.

Insert and subtract $\mu_1$ and $\mu_2$:
\[
\widetilde\mu_1-\widetilde\mu_2
=(\mu_1-\mu_2)
+(\widetilde\mu_1-\mu_1)
-(\widetilde\mu_2-\mu_2).
\]
The triangle inequality in $\gH_k$ therefore yields
\begin{align}
\|\widetilde\mu_1-\widetilde\mu_2\|_{\gH_k}
&\leq
\|\mu_1-\mu_2\|_{\gH_k}
+\sum_{j=1}^2\|\widetilde\mu_j-\mu_j\|_{\gH_k}
\notag\\
&=
\operatorname{MMD}_{k}(Q_1,Q_2)
+\sum_{j=1}^2\|\widetilde\mu_j-\mu_j\|_{\gH_k}.
\label{eq:encoder-pullback-triangle}
\end{align}
For each $j$, linearity of the Bochner integral gives
\[
\widetilde\mu_j-\mu_j
=\E_{Z\sim Q_j}[\varphi(T(Z))-\varphi(Z)].
\]
The norm of a Bochner integral is at most the integral of the norm. Hence,
\begin{align}
\|\widetilde\mu_j-\mu_j\|_{\gH_k}
&\leq
\E_{Z\sim Q_j}
\|\varphi(T(Z))-\varphi(Z)\|_{\gH_k}
\notag\\
&\leq
L_k\E_{Z\sim Q_j}\|T(Z)-Z\|_2
\notag\\
&=
L_k\E_{Z\sim Q_j}\|E(D(Z))-Z\|_2,
\label{eq:encoder-pullback-cycle-bound}
\end{align}
where the second inequality is precisely the assumed Lipschitz continuity of the canonical feature map. Combining Equations \ref{eq:encoder-pullback-exact-identity}--\ref{eq:encoder-pullback-cycle-bound} proves
\begin{align}
&\operatorname{MMD}_{k_E}(D_\#Q_1,D_\#Q_2)
\leq \operatorname{MMD}_{k}(Q_1,Q_2)+L_k\sum_{j=1}^2
\E_{Z\sim Q_j}\|E(D(Z))-Z\|_2.
\label{eq:encoder-pullback-two-law}
\end{align}

It remains to prove the claim involving the actual interventional law. Because
$Q_\pi=E_\#P_\pi$, another application of pushforward integration gives
\begin{align*}
\operatorname{MMD}_{k_E}(D_\#\widehat Q,P_\pi)
&=
\left\|
\E_{Z\sim\widehat Q}[\varphi(T(Z))]
-\E_{Y\sim P_\pi}[\varphi(E(Y))]
\right\|_{\gH_k} =
\left\|
\E_{Z\sim\widehat Q}[\varphi(T(Z))]
-\E_{Z\sim Q_\pi}[\varphi(Z)]
\right\|_{\gH_k}\\
&=
\operatorname{MMD}_{k}(T_\#\widehat Q,Q_\pi).
\end{align*}
Insert and subtract $\E_{Z\sim\widehat Q}[\varphi(Z)]$. The triangle inequality and \eqref{eq:encoder-pullback-cycle-bound}, now with $Q_j=\widehat Q$, give
\begin{align*}
\operatorname{MMD}_{k_E}(D_\#\widehat Q,P_\pi)
&\leq
\operatorname{MMD}_{k}(\widehat Q,Q_\pi)
+L_k\E_{Z\sim\widehat Q}\|E(D(Z))-Z\|_2.
\end{align*}
This proves the claim.

\end{proof}

\section{Supplementary on Numerical Experiments}
\label{sec:SupplementaryExperiments}

We first recall the RKHS objects and kernels used by IMMD-Flow. We then give the exact objective, normalization, optimizer, and learning-rate schedules used in our experiments.

\subsection{RKHS and Kernels}

Let \(U\) take values in a measurable space \(\mathcal U\), and let
\(k_{\mathcal U}:\mathcal U\times\mathcal U\to\mathbb R\) be a
measurable positive-definite kernel with RKHS
\(\mathcal H_{\mathcal U}\). Its canonical feature map is
\(\varphi_{\mathcal U}(u)=k_{\mathcal U}(u,\cdot)\), and the
reproducing property gives
\(f(u)=\langle f,\varphi_{\mathcal U}(u)\rangle_{\mathcal H_{\mathcal U}}\)
for every \(f\in\mathcal H_{\mathcal U}\)
\citep{berlinet2011reproducing,gretton2013introduction}. If
\(\mathbb E[\sqrt{k_{\mathcal U}(U,U)}]<\infty\), the kernel mean
embedding of a law \(P\) is the Bochner expectation
\[
    \mu_P
    :=
    \mathbb E_{U\sim P}\!\left[\varphi_{\mathcal U}(U)\right]
    \in\mathcal H_{\mathcal U},
    \qquad
    \mathbb E_{U\sim P}[f(U)]
    =
    \langle f,\mu_P\rangle_{\mathcal H_{\mathcal U}}.
\]
For two laws \(P,Q\), their maximum mean discrepancy is
\[
    \operatorname{MMD}_{k_{\mathcal U}}(P,Q)
    :=
    \|\mu_P-\mu_Q\|_{\mathcal H_{\mathcal U}}.
\]
A characteristic kernel makes \(P\mapsto\mu_P\) injective, in which
case zero MMD implies \(P=Q\)
\citep{HilbertSpaceEmbeddingforDistributions}. Likewise, whenever the
conditional Bochner expectation exists,
\[
    m_{U\mid V}(v)
    :=
    \mathbb E[\varphi_{\mathcal U}(U)\mid V=v]
    \in\mathcal H_{\mathcal U}
\]
is the conditional mean embedding of \(P_{U\mid V=v}\)
\citep{HilbertSpaceEmbeddingforConditionalDistributions,
grunewalder2012conditional,klebanov2020rigorous}. In our construction,
\(U=Z=E(Y)\), \(V=S=(A,X)\), and \(m_{U\mid V}\) is the nuisance
embedding \(m_0\). We suppress space and kernel subscripts when they
are clear.

\paragraph{Kernels.}
For \(z,z'\in\mathbb R^{d_Z}\), we use the radial basis function (RBF)
and inverse multiquadric (IMQ) kernels
\begin{align*}
k_{\mathrm{RBF}}(z,z';\ell)
&=
\exp\!\left(-\frac{\|z-z'\|_2^2}{2\ell^2}\right),
\qquad \ell>0,\\
k_{\mathrm{IMQ}}(z,z';c,\beta)
&=
\left(c^2+\|z-z'\|_2^2\right)^\beta,
\qquad c>0,\quad \beta<0.
\end{align*}
Both are positive-definite kernels. For the Riesz interaction, define
\[
    \rho_s(z,z'):=\|z-z'\|_2^s,
    \qquad
    \kappa_s(z,z'):=-\rho_s(z,z'),
    \qquad 0<s<2.
\]
The function \(\kappa_s\) is conditionally positive definite rather
than positive definite: for arbitrary points \(z_1,\ldots,z_N\) and
coefficients \(c_1,\ldots,c_N\) satisfying
\(\sum_{a=1}^N c_a=0\),
\[
    \sum_{a=1}^N\sum_{b=1}^N
    c_ac_b\,\kappa_s(z_a,z_b)\geq0.
\]
For any anchor \(z_\star\), the associated anchored kernel
\[
    k_{s,z_\star}(z,z')
    :=
    \frac12\left\{
       \rho_s(z,z_\star)+\rho_s(z',z_\star)-\rho_s(z,z')
    \right\}
\]
is positive definite. If \(X,X'\overset{\mathrm{iid}}{\sim}P\) and
\(Y,Y'\overset{\mathrm{iid}}{\sim}Q\), with the required finite
\(s\)th moments, then
\begin{align*}
\mathcal D_s(P,Q)
&:=
2\mathbb E\rho_s(X,Y)
-\mathbb E\rho_s(X,X')
-\mathbb E\rho_s(Y,Y') =
2\,\operatorname{MMD}_{k_{s,z_\star}}^2(P,Q).
\end{align*}
The anchor cancels from this discrepancy. Thus, following the Riesz
flow literature, we use a \emph{Riesz kernel} for the conditionally
positive-definite interaction \(\kappa_s\), while its probability-law
discrepancy is equivalently an MMD for the anchored kernel
\citep{hertrich2024generative,hertrich2024wasserstein}.

\subsection{Riesz Particle Optimization with AdamW and Warm-up Learning Rate}
\label{sec:riesz-adamw}

All reported Riesz experiments use \(s=1\) and the full-dimensional interaction. To distinguish the moving particles from the fixed logged support, write \(G_1,\ldots,G_M\in\mathbb R^{d_Z}\) for the equal-mass particles and \(\bar Z_i\), \(i=1,\ldots,n\), for the fixed target locations in the coordinates supplied to the optimizer. For CelebA and dSprites, the flow coordinates are standardized
encodings,
\[
    \bar Z_i
    =
    \{E(Y_i)-\widehat\mu_Z\}\oslash\widehat\sigma_Z,
\]
where \(\widehat\mu_Z\) and \(\widehat\sigma_Z\) are the coordinatewise
mean and standard deviation of the logged latents, with standard
deviations floored at \(10^{-3}\), and \(\oslash\) denotes
coordinatewise division. We apply the inverse affine map before
decoding the final particles.
The coefficients \(v_i=v_{{\rm DR},i}\) are constructed as in
Appendix~\ref{sec:orthogonal-target-computation}. The complete
weighted empirical energy is
\begin{align}
\mathcal D_s(G_{1:M};v,\bar Z_{1:n})
={}&
\frac{2}{M}\sum_{j=1}^M\sum_{i=1}^n
v_i\|G_j-\bar Z_i\|_2^s
-\frac{1}{M^2}\sum_{j=1}^M\sum_{\ell=1}^M
\|G_j-G_\ell\|_2^s
-
\sum_{i=1}^n\sum_{q=1}^n
v_iv_q\|\bar Z_i-\bar Z_q\|_2^s.
\label{eq:full-riesz-energy}
\end{align}
The last term depends only on the fixed target. We therefore optimize
\begin{equation}
\mathcal L_s(G_{1:M})
=
\frac{2}{M}\sum_{j=1}^M\sum_{i=1}^n
v_i\|G_j-\bar Z_i\|_2^s
-\frac{1}{M^2}\sum_{j=1}^M\sum_{\ell=1}^M
\|G_j-G_\ell\|_2^s.
\label{eq:optimized-riesz-energy}
\end{equation}

The first term couples the particles to the fixed weighted target, whereas the second induces repulsion among particles. When \(v_i\geq0\) and \(\sum_i v_i=1\), this objective differs from the empirical energy distance only by a target-dependent constant. For generally signed coefficients, we interpret \eqref{eq:optimized-riesz-energy} as a signed Riesz interaction objective rather than as a discrepancy between two probability laws.

\paragraph{Adaptive scale and particle gradient.}
At iteration \(r\), define
\[
    (U_1^{(r)},\ldots,U_{M+n}^{(r)})
    :=
    (G_1^{(r)},\ldots,G_M^{(r)},\bar Z_1,\ldots,\bar Z_n)
\]
and compute
\begin{equation}
a_r
:=
\max\!\left\{
10^{-3},
\frac{1}{M(M+n)\sqrt{d_Z}}
\sum_{j=1}^M\sum_{q=1}^{M+n}
\|G_j^{(r)}-U_q^{(r)}\|_2
\right\}.
\label{eq:riesz-adaptive-scale}
\end{equation}
Both the particles and target are divided by \(a_r\) before evaluating the loss. The scale is recomputed at each iteration but is treated as constant during differentiation. Consequently, for \(s=1\), the particle gradient is
\begin{equation}
g_j^{(r)}
=
\frac{1}{a_r}\left\{
\frac{2}{M}\sum_{i=1}^n
v_i\psi(G_j^{(r)}-\bar Z_i)
-\frac{2}{M^2}\sum_{\ell=1}^M
\psi(G_j^{(r)}-G_\ell^{(r)})
\right\},
\label{eq:riesz-particle-gradient}
\end{equation}
where \(\psi(u)=u/\|u\|_2\) for \(u\neq0\) and \(\psi(0)=0\).
The implementation uses unsmoothed pairwise Euclidean distances. Thus, a coincident pair contributes the zero subgradient selected by automatic differentiation.

\paragraph{Optimizer and learning-rate schedule.}
We optimize the particle locations using AdamW \citep{kingma2015adam,loshchilov2018decoupled}, with zero weight decay, \(\epsilon_{\rm A}=10^{-8}\), and AMSGrad disabled. The benchmark-specific momentum parameters are reported in Table~\ref{tab:riesz-adamw-settings}.

Let \(\eta_{\min}=10^{-6}\), let \(\eta_{\max}\) denote the peak learning rate, let \(W\) be the number of warm-up iterations, and let \(R\) be the total iteration budget. During warm-up,
\begin{equation}
    \eta_r
    =
    \eta_{\min}
    +
    \left(
        \eta_{\max}-\eta_{\min}
    \right)
    \frac{r}{W},
    \qquad
    0\leq r<W.
    \label{eq:riesz-learning-rate-warmup}
\end{equation}
After warm-up, we set
\begin{equation}
    \eta_r
    =
    \eta_{\max}\,\omega_r,
    \qquad
    W\leq r<R,
    \label{eq:riesz-learning-rate-post-warmup}
\end{equation}
where
\begin{equation}
    \omega_r
    =
    \begin{cases}
        1,
        &\text{constant schedule},\\[3pt]
        \dfrac12
        \left[
            1+
            \cos\!\left(
                \pi
                \dfrac{r-W}{R-W}
            \right)
        \right],
        &\text{cosine schedule}.
    \end{cases}
    \label{eq:riesz-post-warmup-factor}
\end{equation}

The algorithm performs exactly \(R\) updates, with no early stopping. The particles are decoded after the final update. A dense iteration costs
\(
    O\!\left(
        M(n+M)d_Z
    \right)
\)
time and stores
\(
    O\!\left(
        M(n+M)
    \right)
\)
pairwise distances.

\begin{table}[H]
\centering
\caption{Riesz--AdamW settings used for the reported experiments. All rows use \(W=100\), \(M=2{,}000\), adaptive scaling, no weight decay, and no injected noise.}
\label{tab:riesz-adamw-settings}
\begin{adjustbox}{max width=\textwidth}
\begin{tabular}{lrrrrcl}
\toprule
Benchmark & \(d_Z\) & \(n\) & \(R\) & \(\eta_{\max}\) &
Post-warm-up & \((\beta_1,\beta_2)\)\\
\midrule
Two-dimensional mixture & 2 & 20,000 & 4,000 & \(5{\times}10^{-4}\) &
cosine & \((0.9,0.999)\)\\
CelebA & 4,096 & 20,000 & 4,000 & \(10^{-3}\) &
cosine & \((0.9,0.95)\)\\
dSprites & 32 & {20,000} & 2,000 & \(10^{-2}\) &
constant & \((0.9,0.95)\)\\
\bottomrule
\end{tabular}
\end{adjustbox}
\end{table}

\subsection{Details on Data-Generating Processes}
\label{sec:benchmark-constructions}

We evaluate IMMD-Flow on three main benchmarks. ColorMNIST and CelebA evaluate structured image outcomes under discrete interventions and are adapted from \citet{dance2026debiased}. dSprites evaluates a stochastic intervention with a continuous action and is adapted from the setting of \citet{zenati2025cpme}. We additionally use a two-dimensional null-effect mixture \citet{dance2026debiased} in our ablation studies (see Appendix~\ref{sec:ablation-studies}). We describe the exact constructions used in our experiments below.

\paragraph{ColorMNIST.}
We adapt the foreground-color confounding benchmark of \citet{dance2026debiased} that uses MNIST images \citep{lecun1998mnist}. We generate
\begin{align*}
    X &\sim \mathrm{Unif}[0,1], \\
    A\mid X=x &\sim \mathrm{Bernoulli}\!\left(\sigma_\ell\!\left(w(x-1/2)\right)\right), \\
    S\mid A=a &\sim \widehat P_{d_a},
    \quad (d_0,d_1)=(1,6).
\end{align*}
Here, $\widehat P_{d_a}$ is the uniform empirical law of grayscale MNIST training images with label $d_a$; images are sampled with replacement. For $S\in[0,1]^{28\times28}$, define
\begin{align*}
    M(S)&=\sigma_\ell\!\left(\kappa(S-\tau)\right),\enspace c_{\mathrm{fg}}(x)=(x,\alpha,1-x)^\top, \enspace
    Y=M(S)\odot\bigl(c_{\mathrm{fg}}(X)\otimes S\bigr).
\end{align*}
The sigmoid and mask operations are pixelwise, and the background is black. Our default configuration uses $(w,\kappa,\tau,\alpha)=(10,20,0.08,0.2)$. The confounding ablation varies $w\in\{1,\ldots,10\}$. To sample from $P_{Y(a)}$, we fix the digit class at $d_a$, independently redraw $X\sim\mathrm{Unif}[0,1]$, and draw another image from $\widehat P_{d_a}$.

\paragraph{CelebA.}
This benchmark evaluates population-level hair-color standardization within each sex group. We use the CelebA data set \citep{liu2015faceattributes} and follow the controlled construction of \citet{dance2026debiased}. Let
\[
    A
    =
    \bm 1\{\mathrm{Male}\},
    \qquad
    X
    =
    \bm 1\{\mathrm{Blond\ Hair}\}.
\]
We combine the official training and validation splits, retain images
with exactly one of the blond-hair and brown-hair labels, and exclude
images labeled black hair, gray hair, bald, or wearing a hat. Let
\(\widehat P_{a,x}\) denote the empirical image distribution in cell
\((A=a,X=x)\).

The logged design is
\begin{align*}
    P_{\mathrm{obs}}(X=1)&=q=0.3, \quad
    P_{\mathrm{obs}}(A=1\mid X=0)=0.6,
    \quad P_{\mathrm{obs}}(A=1\mid X=1)=0.1, \quad
    Y\mid(A=a,X=x)\sim\widehat P_{a,x}.
\end{align*}
Conditional on each cell, images are sampled without replacement. Let $n_{ax}=\#\{i:A_i=a,X_i=x\}$. At $n_{\mathrm{obs}}=20{,}000$, the counts $(n_{00},n_{01},n_{10},n_{11})$ are $(5{,}600,5{,}400,8{,}400,600)$. For each sex group $a$, the target standardizes the hair-color composition to
\begin{align*}
    P_a^\star=(1-q)\widehat P_{a,0}+q\widehat P_{a,1}.
\end{align*}
We use $2{,}000$ held-out reference images per group: $1{,}400$ brown-haired and $600$ blond-haired images. Logged and reference images are disjoint within every cell. This is population-level post-stratification within a sex group; it is not a same-person edit or a literal intervention on sex. We center-crop the aligned images to $178\times178$, resize them to $256\times256$, and scale pixels to $[-1,1]$.

\paragraph{dSprites.}
{ 
We construct a confounded continuous-action benchmark from dSprites
\citep{dsprites17}, extending the setup of \citet{zenati2025cpme}. Draw
$X=(X_1,X_2)\sim\mathrm{Unif}([0,1]^2)$ and set
\begin{align*}
    C(X)&=\min\{2,\lfloor3X_1\rfloor\}, &
    S(X)&=\min\{5,\lfloor6X_2\rfloor\},
\end{align*}
which select one of three shapes and six scales; orientation is fixed at zero.
Let $R_\theta$ denote the two-dimensional rotation matrix and define
$\mu_0(x)=1.8R_{\pi/6}(x-\tfrac12\bm{1}_2)$. The logging and target policies are
\begin{align*}
    \pi_0(\cdot\mid x)
    &=\gN\!\left(\mu_0(x),0.6^2I_2\right),\\
    \pi(\cdot\mid x)
    &=\tfrac12\gN\!\left(\mu_0(x),0.6^2I_2\right)
      +\tfrac12\gN\!\left(\mu_0(x)+\Delta,0.6^2I_2\right),
      \qquad \Delta=(0.45,-0.45)^\top.
\end{align*}
The corresponding oracle importance ratio is
\begin{align*}
    \frac{\pi(a\mid x)}{\pi_0(a\mid x)}
    =\frac12+\frac12\exp\!\left\{
      \frac{\Delta^\top(a-\mu_0(x))}{0.6^2}
      -\frac{\lVert\Delta\rVert_2^2}{2(0.6^2)}\right\}.
\end{align*}
For $j\in\{1,2\}$, the continuous image coordinate is
$p_j(a)=31\sigma_\ell(1.5a_j)$. The deterministic outcome
$Y=g(X,A)\in[0,1]^{64\times64}$ is obtained by bilinearly interpolating the
four neighboring dSprites templates at $p(A)$, using the shape and scale
selected by $X$; no pixel noise is added. Thus $X$ affects both treatment and
visual content, whereas $A$ controls position, so adjustment for the backdoor
path $A\leftarrow X\rightarrow Y$ is necessary. The policy changes the coupling
between content and position while preserving the marginal distribution of
shape and scale. Fresh draws $X^\star\sim P_X$, $A^\star\sim\pi(\cdot\mid
X^\star)$, and $Y^\star=g(X^\star,A^\star)$ provide exact oracle target samples.
}

\paragraph{Two-dimensional Gaussian mixture.}
We use the null-effect deconfounding benchmark of \citet{dance2026debiased}. Define $\sigma_\ell(t)=(1+e^{-t})^{-1}$ and $b=(1,-1)^\top/\sqrt{2}$. We generate
\begin{align*}
    X &\sim \gN(0,I_2), \quad A\mid X=x \sim \mathrm{Bernoulli}\!\left(\sigma_\ell(\gamma b^\top x)\right), \\
    B &\sim \mathrm{Bernoulli}(1/2),
    \quad \epsilon \sim \gN(0,\sigma_Y^2 I_2), \quad
    Y = X+\delta B\bm{1}_2+\epsilon.
\end{align*}
Here, $B$ and $\epsilon$ are mutually independent and independent of $(X,A)$. We set $(\gamma,\delta,\sigma_Y)=(3,2,0.2)$. Treatment does not enter the outcome equation. Hence $P_{Y(0)}=P_{Y(1)}=P_Y$, although $P_{Y\mid A=a}$ generally differs from $P_Y$.

\subsection{ {Autoencoders for Image Experiments}}
\label{sec:image-autoencoders}

{ 
For ColorMNIST and dSprites, we train convolutional autoencoders on all logged images once per experimental repetition and freeze them before constructing the target embedding and running the particle flow. The ColorMNIST encoder maps a $3\times28\times28$ input through three stride-two convolutional blocks with $32$, $64$, and $128$ channels and spatial sizes $14$, $7$, and $4$, respectively. Each block uses group normalization and a SiLU activation, and a linear layer produces a 32-dimensional bottleneck. Its mirrored transposed-convolution decoder ends with a tanh output. The dSprites encoder maps a $1\times64\times64$ input through four analogous blocks with $32$, $64$, $128$, and $128$ channels and spatial sizes $32$, $16$, $8$, and $4$, followed by a 32-dimensional bottleneck; its mirrored decoder ends with a sigmoid output. We minimize reconstruction MSE using Adam \citep{kingma2015adam}, with batch size $128$ and learning rate $10^{-3}$, for $3{,}000$ ColorMNIST iterations and $4{,}000$ dSprites iterations. During dSprites training, we add fresh Gaussian noise with standard deviation $0.05$ to each input and reconstruct the corresponding
logged image. The statistical analysis in Section \ref{sec:consistency} treats \(E\) and \(D\) as fixed independently of the estimation sample. The same-sample autoencoder training used for ColorMNIST and dSprites in our experiments is not covered by the present guarantees.

For CelebA, we instead use the frozen pretrained \texttt{stabilityai/sd-vae-ft-mse} VAE \citep{Rombach_SD_VAE}, without fine-tuning, and encode each $256\times256$ image by the posterior mode into a $4\times32\times32$ latent. In every case, IMMD-Flow operates on the fixed
latent representations and the final particles are decoded to image space.
}

\subsection{Qualitative samples and particle evolution}
\label{sec:qualitative-results}

In this section, we present qualitative examples of the initialized and generated particles. We also visualize selected particles at intermediate iterations to illustrate how the empirical distribution evolves under the fixed estimated target.

\paragraph{ColorMNIST samples.}

For an image \(y\), let \(R(y)\) and \(B(y)\) denote its total red- and blue-channel intensities, and define the foreground-colour statistic
\begin{equation}
    c(y)
    :=
    \frac{R(y)}
         {R(y)+B(y)}.
    \label{eq:colormnist-colour-statistic}
\end{equation}
Values near zero correspond to predominantly blue foregrounds, values near one correspond to predominantly red foregrounds, and \(c(y)=1/2\) represents equal red and blue content. Under the
ColorMNIST renderer,
\(
    c(Y)=X
\)
up to reconstruction and numerical error, because the red and blue foreground intensities are proportional to \(X\) and \(1-X\), respectively. Since the intervention preserves \(X\sim\mathrm{Unif}[0,1]\), the oracle mean of this statistic is \(1/2\).

\begin{figure}[p]
\centering
\begin{subfigure}[t]{\textwidth}
    \centering
    \includegraphics[
        width=0.85\textwidth
    ]{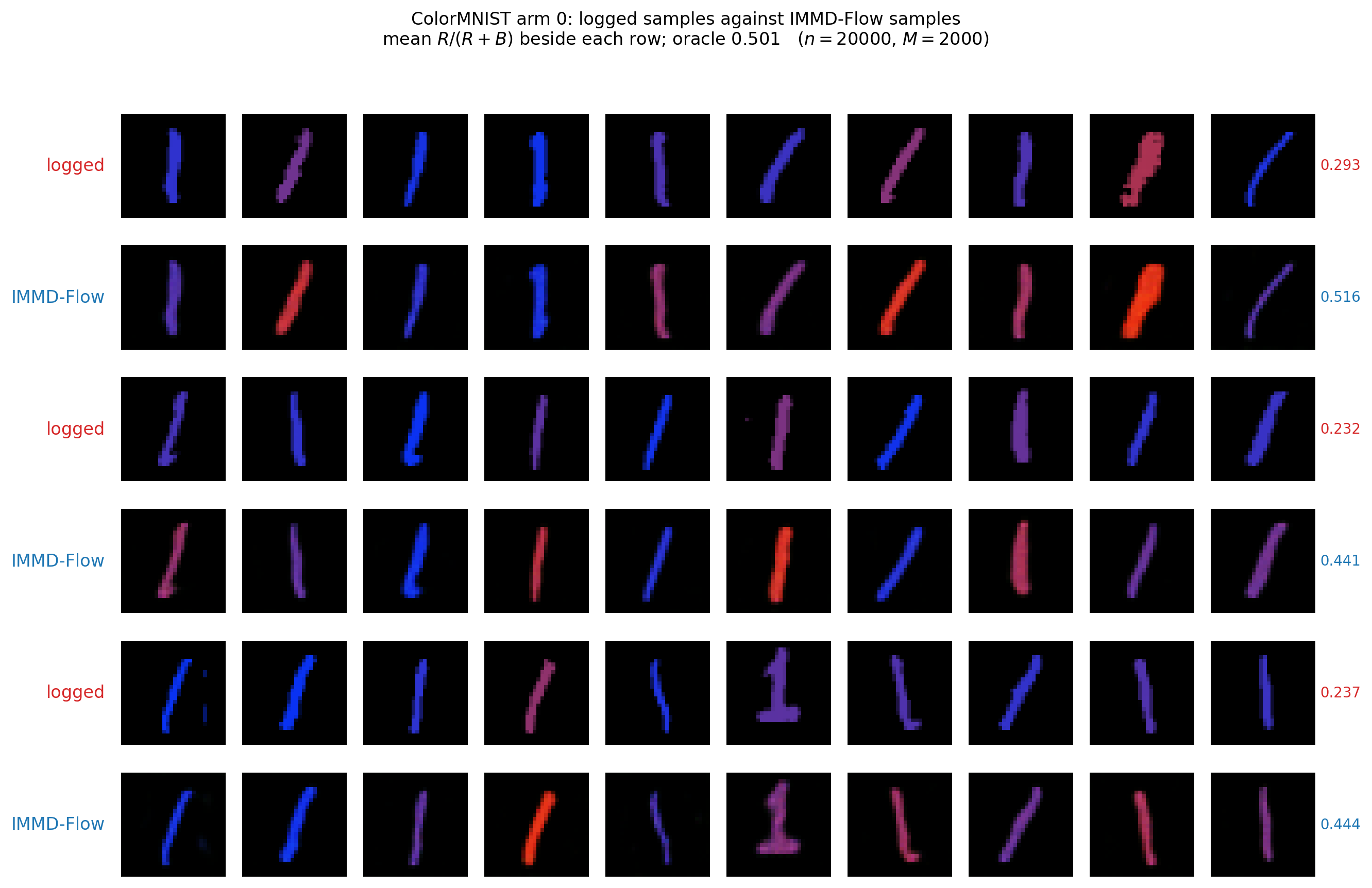}
    \caption{Intervention arm \(a=0\).}
\end{subfigure}

\vspace{0.75em}

\begin{subfigure}[t]{\textwidth}
    \centering
    \includegraphics[
        width=0.85\textwidth
    ]{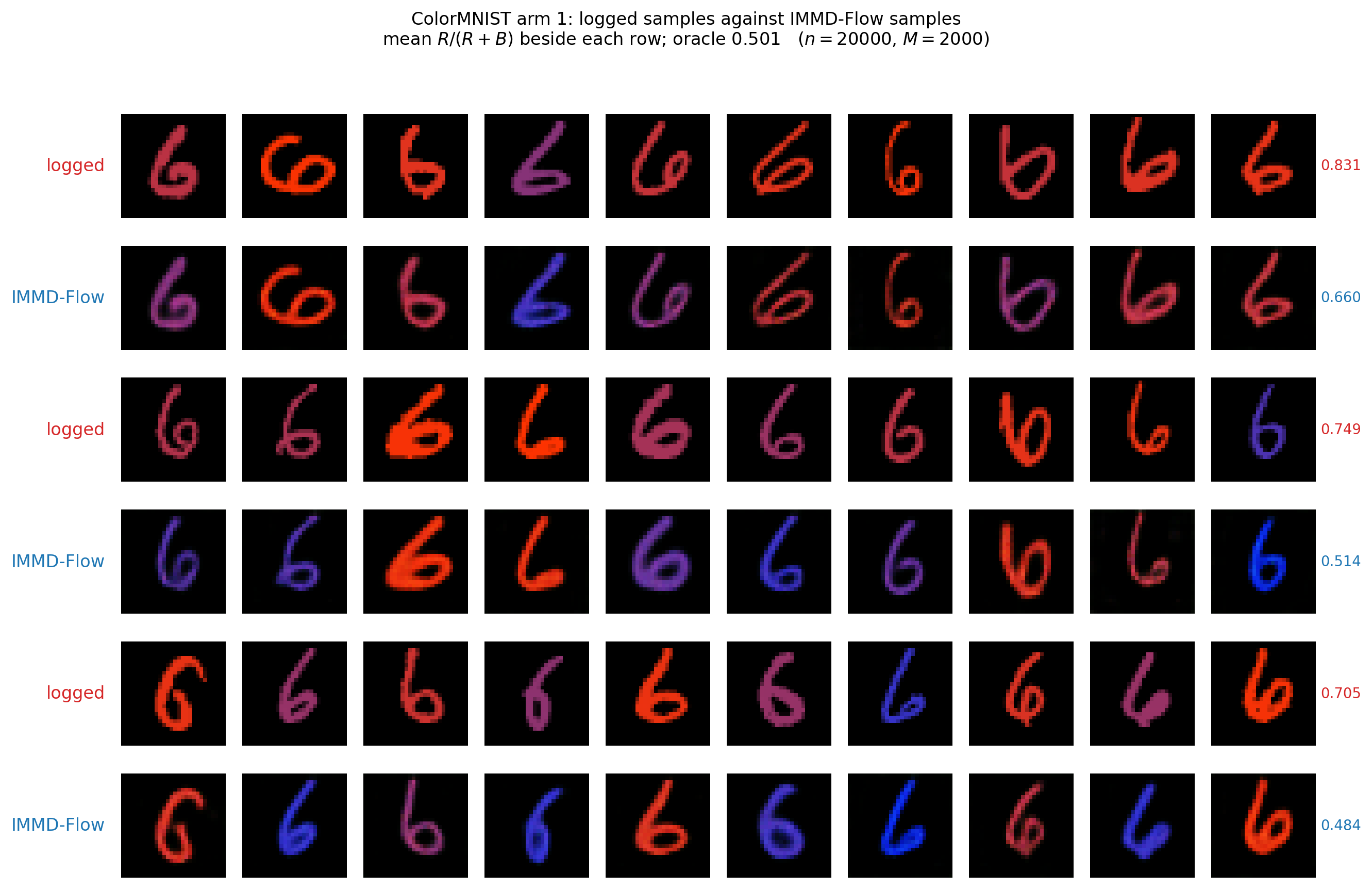}
    \caption{Intervention arm \(a=1\).}
\end{subfigure}

\caption{
Initialized and final ColorMNIST particles. Within each pair of rows, columns correspond to the same numerical particles before and after optimization. The number displayed beside each row is the empirical mean of the foreground-colour statistic \(R/(R+B)\); the oracle mean is \(0.501\). IMMD-Flow moves the colour composition toward the interventional target while preserving the digit class. 
}
\label{fig:colormnist-samples}
\end{figure}

Figure~\ref{fig:colormnist-samples} compares particles before and after IMMD-Flow in both intervention arms. The initialized observational-arm particles have foreground-colour distributions shifted in opposite directions. The final particles move toward the interventional colour law while largely preserving the treatment-specific digit identity.

\begin{figure}[!htbp]
\centering
\includegraphics[
    width=\textwidth
]{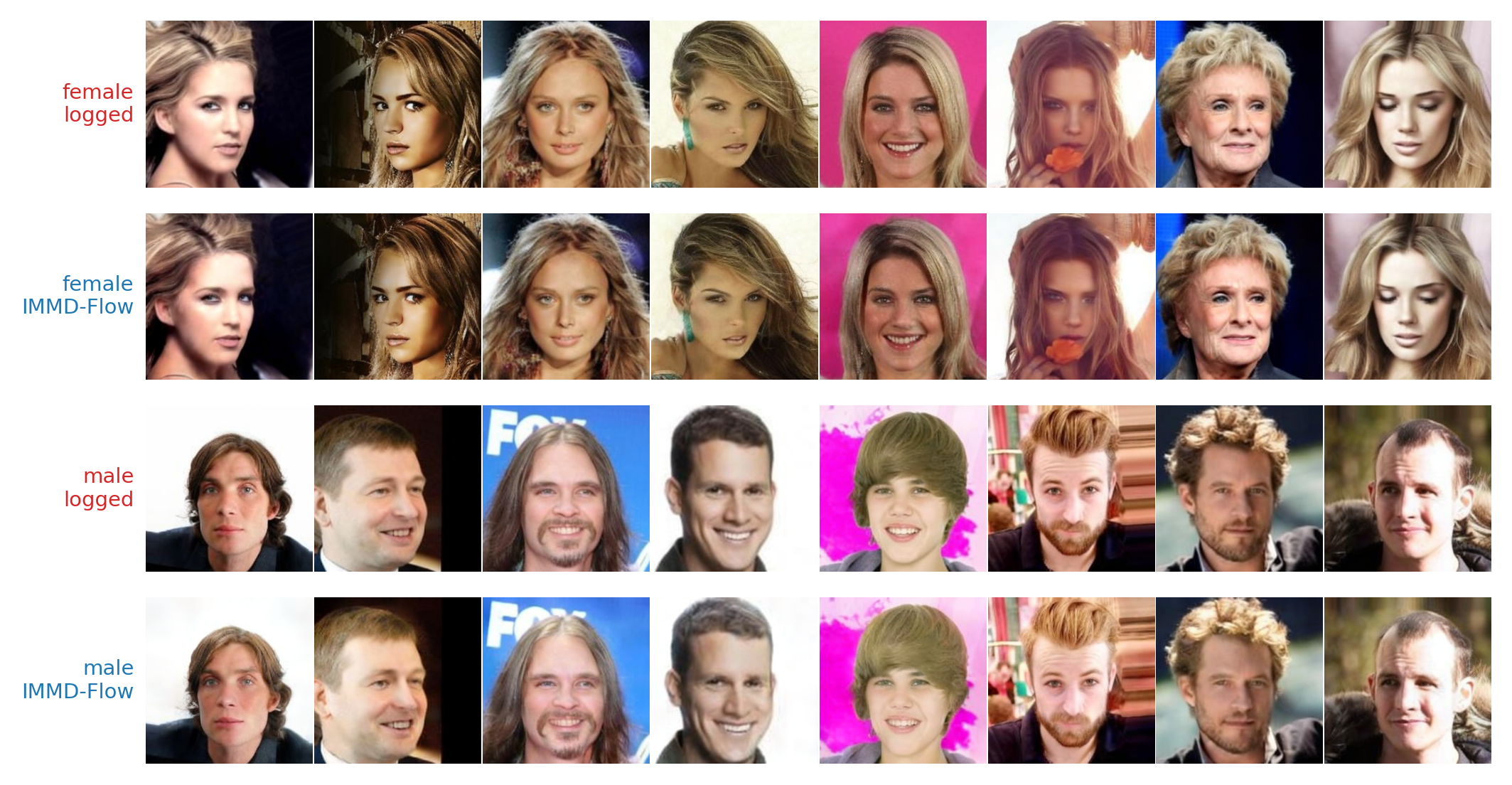}
\caption{
Example CelebA particles before and after IMMD-Flow for the female and male groups. 
}
\label{fig:celeba-samples}
\end{figure}

\paragraph{CelebA samples.}
Figure~\ref{fig:celeba-samples} presents initialized and final particles
for both sex groups. The generated particles remain visually coherent
while changing under the fixed population-level hair-colour target.

\begin{figure}[!htbp]
\centering

\includegraphics[
    width=\textwidth,
    height=0.38\textheight,
    keepaspectratio
]{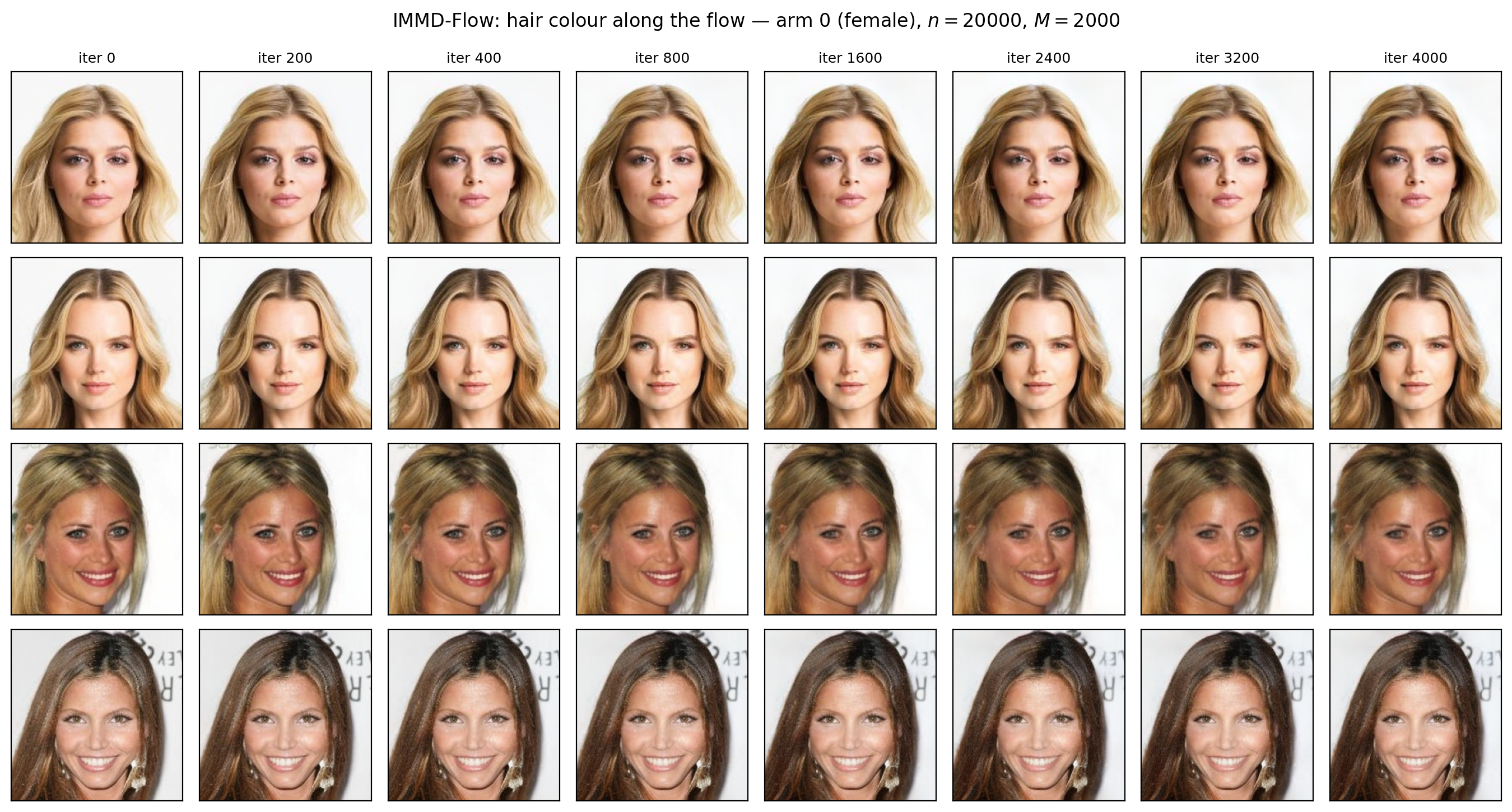}

\vspace{0.5em}

\includegraphics[
    width=\textwidth,
    height=0.38\textheight,
    keepaspectratio
]{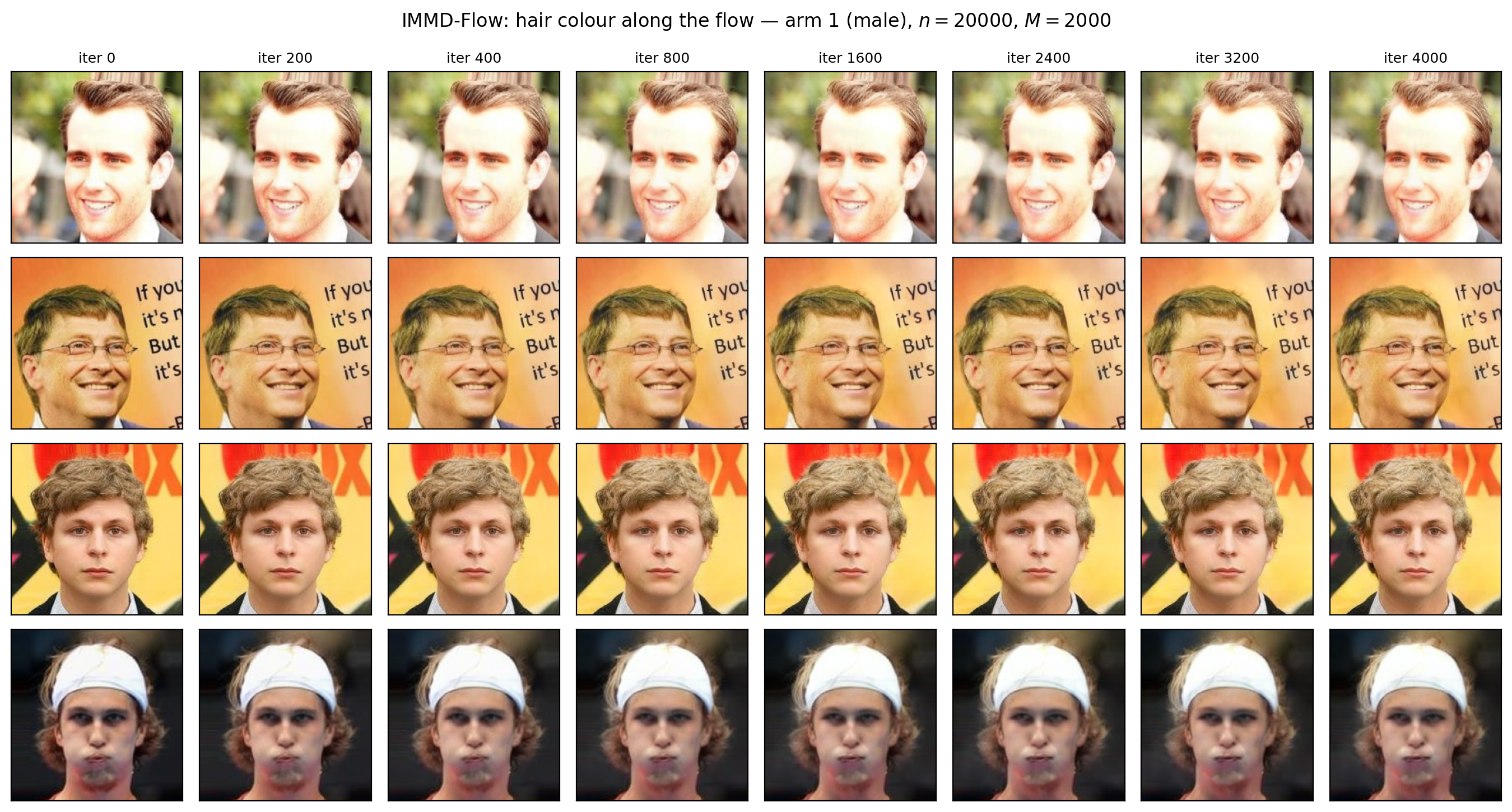}
\caption{
Evolution of selected CelebA particles during IMMD-Flow. The top and bottom panels correspond to the female (\(a=0\)) and male (\(a=1\)) groups, respectively. 
}
\label{fig:celeba-particle-evolution}
\end{figure}

\paragraph{Particle evolution on CelebA.}
Figure~\ref{fig:celeba-particle-evolution} shows selected particles at intermediate iterations. The decoded images change gradually rather than being replaced by unrelated samples, while the target embedding and its coefficients remain fixed throughout optimization.

\clearpage
\subsection{Additional experiments and ablations}
\label{sec:ablation-studies}

We complement the main benchmark comparisons with a low-dimensional illustration and a series of ablations examining the effects of the logged sample size, particle count, particle initialization, projected Nystr\"om approximation, and confounding strength. Unless an experiment varies the stated component, we use the doubly robust target, full-sample nuisance fitting, projected Nystr\"om regression, observed-outcome initialization, and \(M=2{,}000\) particles. ColorMNIST uses the IMQ--Euler update, while CelebA and dSprites use Riesz--AdamW. For the two-dimensional benchmark, we report both RBF--Euler and Riesz--AdamW.

Entries are means and sample standard deviations across three data seeds. For the two static interventions, metrics are first averaged across intervention arms within each seed. For dSprites, they are averaged across ten fresh target-policy draws within each seed. Outcome- and latent-space \(\mathrm{SW}_2\) values are multiplied by \(10^3\) throughout this subsection.

\paragraph{Null-effect deconfounding illustration.}
We first illustrate IMMD-Flow on the two-dimensional null-effect benchmark of \citet{dance2026debiased} which is explained in the previous section. Treatment does not enter the outcome equation, so
\[
    P_{Y(0)}
    =
    P_{Y(1)}
    =
    P_Y.
\]
Nevertheless, treatment assignment depends on \(X\), which also enters the outcome equation. Consequently, the observational arm distributions \(P_{Y\mid A=a}\) generally differ from the common interventional law \(P_Y\).

\begin{figure}[!htbp]
\centering
\includegraphics[width=0.70\textwidth]{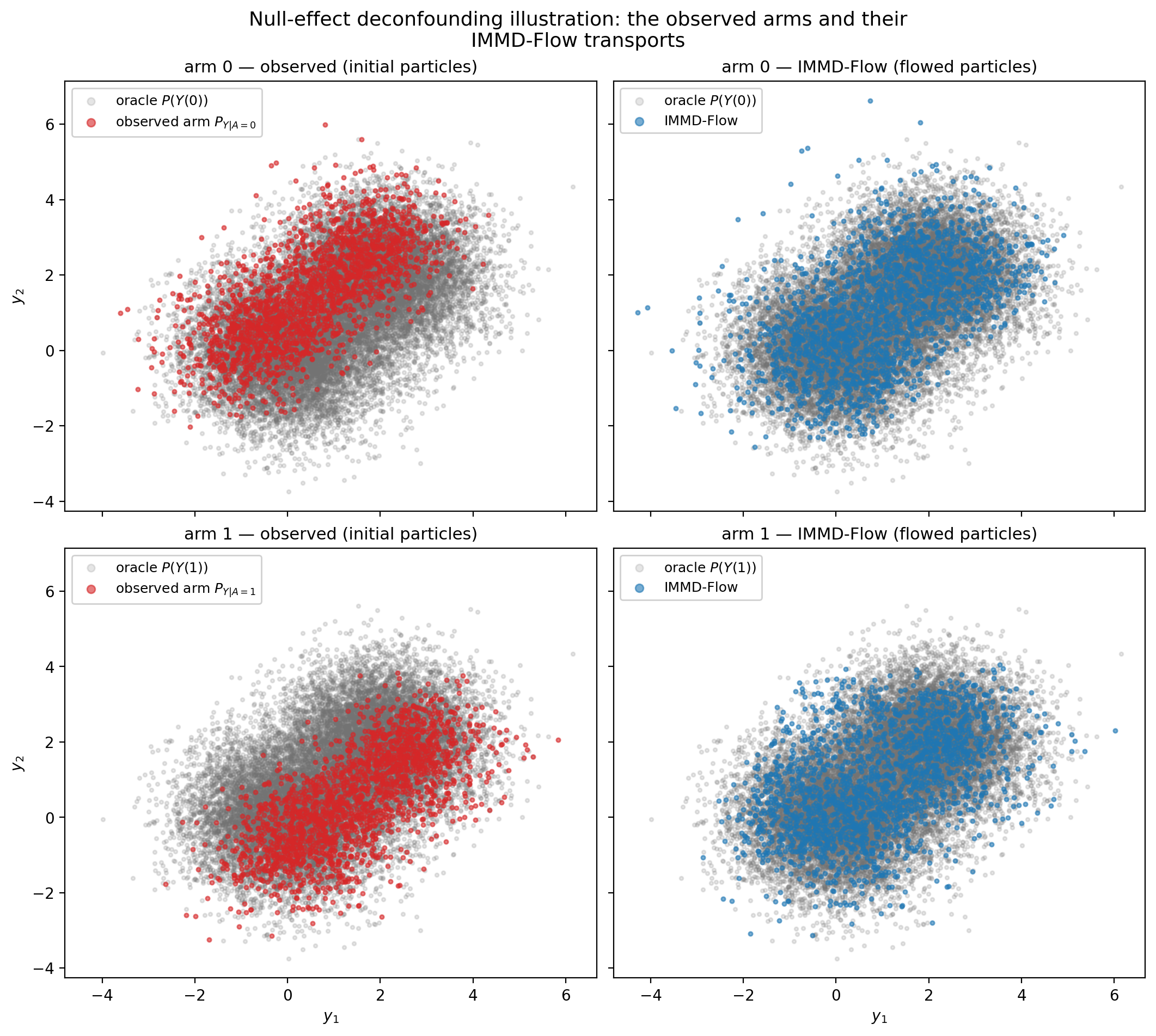}
\caption{ {Initial and flowed particles in the two-dimensional benchmark of \citet{dance2026debiased}. Rows correspond to the intervention arms $a\in\{0,1\}$. Gray points show a fresh oracle draw from the common interventional law $P_{Y(a)}=P_Y$; red points are particles initialized from $P_{Y\mid A=a}$, and blue points are returned by DR--IMMD with the RBF--Euler update. We use $n=20{,}000$ logged observations and $M=2{,}000$ particles.}}
\label{fig:decfm2d-initial-flowed}
\end{figure}

\begin{figure}[!htbp]
\centering
\includegraphics[width=0.90\textwidth]{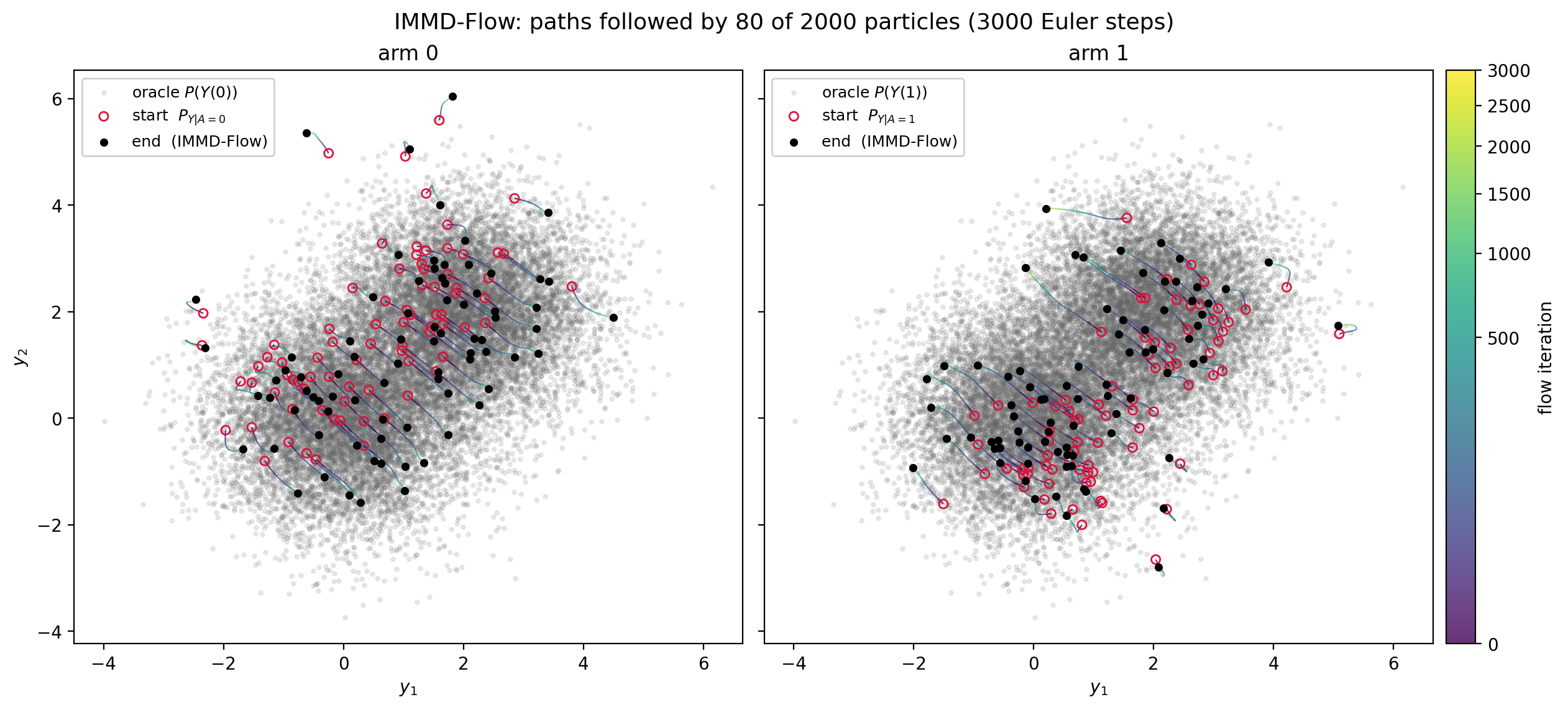}
\caption{ {Paths of 80 representative particles for each
intervention arm in the run from Figure~\ref{fig:decfm2d-initial-flowed} over
3,000 RBF--Euler updates. Path color indicates the iteration; open red markers
and black markers denote initial and returned particles, respectively. Gray
points show a fresh oracle draw from $P_Y$.}}
\label{fig:decfm2d-particle-paths}
\end{figure}

Figure~\ref{fig:decfm2d-initial-flowed} compares particles initialized from each observational arm with the particles returned by DR--IMMD. The initial clouds overrepresent different regions of the common target, whereas the flowed particles recover its overall mixture structure more accurately. Figure~\ref{fig:decfm2d-particle-paths} visualizes representative particle paths during optimization.

The trajectories are generally curved because each particle follows a time-varying MMD witness determined by the evolving empirical particle distribution. They are not intended to approximate straight optimal-transport displacement paths.

\paragraph{Effect of the logged sample size.}
Table~\ref{tab:ablation-nobs} shows that IMMD-Flow substantially reduces outcome-space and, where applicable, latent-space \(\mathrm{SW}_2\) relative to initialization at every sample size. The final discrepancies generally improve as \(n\) increases on the structured-outcome benchmarks, most clearly for CelebA and dSprites, whereas the two-dimensional results fluctuate within their across-seed variability. FID behaves differently across benchmarks: the flow improves it on ColorMNIST, changes it little on dSprites, and slightly increases it on CelebA.

\begin{table}[H]
\centering
\caption{
Ablation over the number of logged observations, before and after
particle optimization. ``Initial'' denotes the particle distribution
before the flow: the corresponding observational arm for the
discrete-treatment benchmarks and the logged outcome distribution for
dSprites. ``Flow'' denotes the DR--IMMD output. Lower is better. A dash
denotes an unreported metric; the two-dimensional experiment has neither
a separate latent-space metric nor image FID. All entries are means
\(\pm\) sample standard deviations across three data seeds. FID is
computed against the same held-out interventional reference of
\(2{,}000\) images at every \(n\).
}
\label{tab:ablation-nobs}
\scriptsize
\setlength{\tabcolsep}{3.5pt}
\renewcommand{\arraystretch}{1.08}
\begin{adjustbox}{max width=\textwidth}
\begin{tabular}{@{}llrcccccc@{}}
\toprule
&
&
&
\multicolumn{2}{c}{\(10^3\!\times\!\mathrm{SW}_2\)}
&
\multicolumn{2}{c}{\(10^3\!\times\!\text{latent }\mathrm{SW}_2\)}
&
\multicolumn{2}{c}{FID}
\\
\cmidrule(lr){4-5}
\cmidrule(lr){6-7}
\cmidrule(lr){8-9}
Benchmark
& Particle update
& \(n\)
& initial
& flow
& initial
& flow
& initial
& flow
\\
\midrule

\multirow{8}{*}{2-D mixture}
& \multirow{4}{*}{RBF--Euler}
& \(5{,}000\)
& \(524.82\!\pm\!10.51\)
& \(137.16\!\pm\!37.77\)
& \textemdash
& \textemdash
& \textemdash
& \textemdash
\\
&
& \(10{,}000\)
& \(519.95\!\pm\!12.99\)
& \(110.62\!\pm\!16.37\)
& \textemdash
& \textemdash
& \textemdash
& \textemdash
\\
&
& \(15{,}000\)
& \(514.42\!\pm\!6.11\)
& \(121.34\!\pm\!18.37\)
& \textemdash
& \textemdash
& \textemdash
& \textemdash
\\
&
& \(20{,}000\)
& \(512.49\!\pm\!3.67\)
& \(111.99\!\pm\!11.61\)
& \textemdash
& \textemdash
& \textemdash
& \textemdash
\\
\cmidrule(lr){2-9}
&
\multirow{4}{*}{Riesz--AdamW}
& \(5{,}000\)
& \(524.82\!\pm\!10.51\)
& \(146.52\!\pm\!35.60\)
& \textemdash
& \textemdash
& \textemdash
& \textemdash
\\
&
& \(10{,}000\)
& \(519.95\!\pm\!12.99\)
& \(125.60\!\pm\!18.00\)
& \textemdash
& \textemdash
& \textemdash
& \textemdash
\\
&
& \(15{,}000\)
& \(514.42\!\pm\!6.11\)
& \(127.85\!\pm\!16.09\)
& \textemdash
& \textemdash
& \textemdash
& \textemdash
\\
&
& \(20{,}000\)
& \(512.49\!\pm\!3.67\)
& \(126.29\!\pm\!16.52\)
& \textemdash
& \textemdash
& \textemdash
& \textemdash
\\

\midrule
\multirow{4}{*}{ColorMNIST}
& \multirow{4}{*}{IMQ--Euler}
& \(5{,}000\)
& \(42.03\!\pm\!0.62\)
& \(10.56\!\pm\!1.13\)
& \(478.82\!\pm\!22.18\)
& \(97.68\!\pm\!5.34\)
& \(40.48\!\pm\!1.18\)
& \(28.32\!\pm\!1.34\)
\\
&
& \(10{,}000\)
& \(42.41\!\pm\!0.80\)
& \(10.54\!\pm\!0.86\)
& \(497.85\!\pm\!13.72\)
& \(91.08\!\pm\!1.99\)
& \(42.91\!\pm\!0.91\)
& \(30.33\!\pm\!0.78\)
\\
&
& \(15{,}000\)
& \(42.00\!\pm\!0.61\)
& \(10.66\!\pm\!0.88\)
& \(466.08\!\pm\!24.57\)
& \(81.15\!\pm\!1.05\)
& \(43.53\!\pm\!0.93\)
& \(30.73\!\pm\!0.74\)
\\
&
& \(20{,}000\)
& \(43.41\!\pm\!0.26\)
& \(9.30\!\pm\!0.30\)
& \(496.75\!\pm\!15.73\)
& \(80.04\!\pm\!1.14\)
& \(43.07\!\pm\!0.50\)
& \(30.64\!\pm\!0.81\)
\\

\midrule
\multirow{4}{*}{CelebA}
& \multirow{4}{*}{Riesz--AdamW}
& \(5{,}000\)
& \(55.57\!\pm\!5.61\)
& \(29.96\!\pm\!2.45\)
& \(80.78\!\pm\!4.56\)
& \(56.74\!\pm\!1.45\)
& \(11.36\!\pm\!0.11\)
& \(11.59\!\pm\!0.18\)
\\
&
& \(10{,}000\)
& \(55.76\!\pm\!5.01\)
& \(29.54\!\pm\!1.28\)
& \(82.69\!\pm\!3.25\)
& \(54.44\!\pm\!0.60\)
& \(11.10\!\pm\!0.11\)
& \(11.45\!\pm\!0.13\)
\\
&
& \(15{,}000\)
& \(55.03\!\pm\!5.04\)
& \(27.05\!\pm\!1.38\)
& \(80.51\!\pm\!5.51\)
& \(51.60\!\pm\!1.13\)
& \(11.30\!\pm\!0.08\)
& \(11.59\!\pm\!0.15\)
\\
&
& \(20{,}000\)
& \(49.60\!\pm\!1.30\)
& \(26.56\!\pm\!1.55\)
& \(77.03\!\pm\!3.22\)
& \(51.29\!\pm\!1.36\)
& \(11.08\!\pm\!0.13\)
& \(11.57\!\pm\!0.08\)
\\

\midrule
\multirow{4}{*}{dSprites}
& \multirow{4}{*}{Riesz--AdamW}
& \(5{,}000\)
& \(27.72\!\pm\!1.01\)
& \(9.51\!\pm\!0.43\)
& \(136.05\!\pm\!15.35\)
& \(44.22\!\pm\!2.49\)
& \(31.62\!\pm\!2.17\)
& \(31.63\!\pm\!2.37\)
\\
&
& \(10{,}000\)
& \(27.56\!\pm\!0.92\)
& \(8.71\!\pm\!0.30\)
& \(138.93\!\pm\!10.76\)
& \(41.02\!\pm\!1.26\)
& \(34.88\!\pm\!1.58\)
& \(34.82\!\pm\!1.57\)
\\
&
& \(15{,}000\)
& \(27.02\!\pm\!0.23\)
& \(8.59\!\pm\!0.14\)
& \(139.40\!\pm\!4.30\)
& \(39.22\!\pm\!1.43\)
& \(35.49\!\pm\!1.12\)
& \(35.35\!\pm\!0.72\)
\\
&
& \(20{,}000\)
& \(27.65\!\pm\!0.40\)
& \(8.41\!\pm\!0.08\)
& \(145.96\!\pm\!7.87\)
& \(39.06\!\pm\!0.47\)
& \(36.78\!\pm\!1.52\)
& \(36.49\!\pm\!1.65\)
\\

\bottomrule
\end{tabular}
\end{adjustbox}
\end{table}

\paragraph{Effect of the particle count.}
Table~\ref{tab:ablation-particles} shows that increasing \(M\)
consistently reduces the final outcome- and latent-space \(\mathrm{SW}_2\) across all benchmarks. Relative to the same particles before optimization, IMMD-Flow gives substantial improvements throughout the two-dimensional, ColorMNIST, and dSprites experiments. On CelebA, small particle clouds remain inaccurate in outcome space, but the flow improves upon initialization from \(M=500\) onward, with increasingly clear gains as \(M\) grows.

The final FID also decreases with \(M\). However, because FID is estimated from \(M\) generated images against a fixed reference set, part of this trend reflects its finite-sample dependence. We therefore use FID primarily for initial-versus-flow comparisons at a fixed particle count, rather than to quantify the effect of \(M\) in isolation.

\begin{table}[H]
\centering
\caption{
Ablation over the number of particles before and after IMMD-Flow.
``Initial'' and ``flow'' denote the metrics before and after particle optimization, respectively. Lower is better. Entries are means \(\pm\) sample standard deviations across three data seeds; a dash denotes an unreported metric. Every initial entry is scored on the same \(M\) samples as the corresponding flow entry, so the two are count-matched; both sliced Wasserstein and FID are biased upward at small sample size, and an initial column scored at a fixed larger count would understate the flow's gain at small \(M\). 
}
\label{tab:ablation-particles}
\scriptsize
\setlength{\tabcolsep}{4pt}
\renewcommand{\arraystretch}{1.08}
\begin{adjustbox}{max width=\textwidth}
\begin{tabular}{@{}llr cc cc cc@{}}
\toprule
& & & \multicolumn{2}{c}{$10^3\!\times\!\mathrm{SW}_2$}
    & \multicolumn{2}{c}{$10^3\!\times\!\text{latent }\mathrm{SW}_2$}
    & \multicolumn{2}{c}{FID} \\
\cmidrule(lr){4-5}\cmidrule(lr){6-7}\cmidrule(lr){8-9}
Benchmark & Particle update & $M$ & initial & flow & initial & flow & initial & flow \\
\midrule
\multirow{8}{*}{2-D mixture}
& \multirow{4}{*}{RBF--Euler}
& $250$   & $540.71\!\pm\!2.58$  & $188.85\!\pm\!48.22$ & \textemdash & \textemdash & \textemdash & \textemdash \\
& & $500$   & $529.39\!\pm\!10.14$ & $162.40\!\pm\!30.71$ & \textemdash & \textemdash & \textemdash & \textemdash \\
& & $1{,}000$ & $530.53\!\pm\!7.71$  & $144.06\!\pm\!35.76$ & \textemdash & \textemdash & \textemdash & \textemdash \\
& & $2{,}000$ & $512.49\!\pm\!3.67$  & $111.99\!\pm\!11.61$ & \textemdash & \textemdash & \textemdash & \textemdash \\
\cmidrule(lr){2-9}
& \multirow{4}{*}{Riesz--AdamW}
& $250$   & $540.71\!\pm\!2.58$  & $194.99\!\pm\!52.64$ & \textemdash & \textemdash & \textemdash & \textemdash \\
& & $500$   & $529.39\!\pm\!10.14$ & $170.19\!\pm\!38.50$ & \textemdash & \textemdash & \textemdash & \textemdash \\
& & $1{,}000$ & $530.53\!\pm\!7.71$  & $155.12\!\pm\!38.55$ & \textemdash & \textemdash & \textemdash & \textemdash \\
& & $2{,}000$ & $512.49\!\pm\!3.67$  & $126.29\!\pm\!16.52$ & \textemdash & \textemdash & \textemdash & \textemdash \\
\midrule
\multirow{4}{*}{ColorMNIST}
& \multirow{4}{*}{IMQ--Euler}
& $250$   & $46.69\!\pm\!1.14$ & $17.35\!\pm\!0.53$ & $541.74\!\pm\!19.39$ & $191.63\!\pm\!8.29$ & $50.46\!\pm\!1.40$ & $38.584\!\pm\!0.632$ \\
& & $500$   & $45.00\!\pm\!1.19$ & $13.01\!\pm\!0.53$ & $511.45\!\pm\!31.94$ & $128.26\!\pm\!5.44$ & $46.64\!\pm\!1.20$ & $34.414\!\pm\!0.666$ \\
& & $1{,}000$ & $43.70\!\pm\!0.72$ & $10.92\!\pm\!0.42$ & $500.47\!\pm\!31.26$ & $101.90\!\pm\!1.42$ & $44.38\!\pm\!1.20$ & $32.211\!\pm\!1.030$ \\
& & $2{,}000$ & $43.41\!\pm\!0.26$ & $9.30\!\pm\!0.30$  & $496.75\!\pm\!15.73$ & $80.04\!\pm\!1.14$  & $43.07\!\pm\!0.50$ & $30.644\!\pm\!0.809$ \\
\midrule
\multirow{4}{*}{CelebA}
& \multirow{4}{*}{Riesz--AdamW}
& $250$   & $87.21\!\pm\!7.97$ & $70.24\!\pm\!1.69$ & $159.33\!\pm\!3.30$ & $131.15\!\pm\!1.03$ & $36.62\!\pm\!3.58$ & $75.714\!\pm\!0.883$ \\
& & $500$   & $70.53\!\pm\!8.60$ & $48.38\!\pm\!1.38$ & $121.68\!\pm\!4.00$ & $94.04\!\pm\!0.87$  & $24.89\!\pm\!2.01$ & $37.813\!\pm\!0.705$ \\
& & $1{,}000$ & $59.83\!\pm\!7.78$ & $35.55\!\pm\!2.22$ & $96.55\!\pm\!1.76$  & $69.50\!\pm\!1.81$  & $16.74\!\pm\!1.41$ & $19.226\!\pm\!0.027$ \\
& & $2{,}000$ & $48.12\!\pm\!5.41$ & $26.56\!\pm\!1.55$ & $77.03\!\pm\!3.22$  & $51.29\!\pm\!1.36$  & $11.61\!\pm\!0.87$ & $11.567\!\pm\!0.085$ \\
\midrule
\multirow{4}{*}{dSprites}
& \multirow{4}{*}{Riesz--AdamW}
& $250$   & $39.17\!\pm\!2.46$ & $22.42\!\pm\!0.39$ & $191.83\!\pm\!3.50$ & $108.60\!\pm\!2.16$ & $39.88\!\pm\!2.02$ & $39.816\!\pm\!2.153$ \\
& & $500$   & $33.46\!\pm\!2.10$ & $16.02\!\pm\!0.35$ & $167.76\!\pm\!9.57$ & $77.76\!\pm\!0.35$  & $38.68\!\pm\!1.24$ & $38.338\!\pm\!1.804$ \\
& & $1{,}000$ & $33.34\!\pm\!1.68$ & $11.41\!\pm\!0.32$ & $150.63\!\pm\!5.24$ & $54.43\!\pm\!2.06$  & $37.51\!\pm\!1.45$ & $37.263\!\pm\!1.458$ \\
& & $2{,}000$ & $30.96\!\pm\!1.10$ & $8.41\!\pm\!0.08$  & $145.96\!\pm\!7.87$ & $39.06\!\pm\!0.47$  & $36.78\!\pm\!1.52$ & $36.488\!\pm\!1.648$ \\
\bottomrule
\end{tabular}
\end{adjustbox}
\end{table}

\paragraph{Effect of particle initialization.}
We examine the sensitivity of IMMD-Flow to its initial particle distribution. We compare initialization from observed outcomes with Gaussian alternatives of increasing distance from the observed outcome geometry. The moment-matched Gaussian uses the empirical mean and full covariance. For CelebA, the standardized normal is \(\mathcal N(0,I)\) in the standardized flow coordinates and therefore matches the empirical coordinatewise location and scale, while the raw-encoder normal draws \(\mathcal N(0,I)\) before applying the standardization map. All other target-estimation and optimization
settings are held fixed.

\begin{table}[H]
\centering
\caption{Particle-initialization ablation at $n=20000$ and $M=2{,}000$. Lower is better.}
\label{tab:ablation-initialization}
\scriptsize
\setlength{\tabcolsep}{4pt}
\renewcommand{\arraystretch}{1.08}
\begin{adjustbox}{max width=\textwidth}
\begin{tabular}{@{}lllccc@{}}
\toprule
Benchmark & Particle update & Initialization
& $10^3\!\times\!\mathrm{SW}_2$
& $10^3\!\times\!\text{latent }\mathrm{SW}_2$ & FID \\
\midrule
\multirow{6}{*}{2-D mixture}
& \multirow{3}{*}{RBF--Euler}
& Observed & $111.99\!\pm\!11.61$ & \textemdash & \textemdash \\
& & Moment-matched & $109.75\!\pm\!9.50$ & \textemdash & \textemdash \\
& & Standard normal & $134.08\!\pm\!20.25$ & \textemdash & \textemdash \\
\cmidrule(lr){2-6}
& \multirow{3}{*}{Riesz--AdamW}
& Observed & $126.29\!\pm\!16.52$ & \textemdash & \textemdash \\
& & Moment-matched & $150.14\!\pm\!11.33$ & \textemdash & \textemdash \\
& & Standard normal & $454.90\!\pm\!48.29$ & \textemdash & \textemdash \\
\midrule
\multirow{3}{*}{ColorMNIST}
& \multirow{3}{*}{IMQ--Euler}
& Observed arm & $9.30\!\pm\!0.30$ & $80.04\!\pm\!1.14$ & $30.644\!\pm\!0.809$ \\
& & Moment-matched & $17.03\!\pm\!0.57$ & $187.01\!\pm\!14.33$ & $50.020\!\pm\!0.726$ \\
& & Standard normal & $21.73\!\pm\!0.93$ & $165.85\!\pm\!6.21$ & $63.884\!\pm\!0.934$ \\
\midrule
\multirow{4}{*}{CelebA}
& \multirow{4}{*}{Riesz--AdamW}
& Observed arm & $26.56\!\pm\!1.55$ & $51.29\!\pm\!1.36$ & $11.567\!\pm\!0.085$ \\
& & Moment $\mathcal N(\mu,\Sigma)$ & $37.80\!\pm\!0.40$ & $51.31\!\pm\!1.53$ & $211.398\!\pm\!1.631$ \\
& & Normal (standardized) & $86.18\!\pm\!1.34$ & $54.17\!\pm\!1.58$ & $292.896\!\pm\!1.772$ \\
& & Normal (raw encoder) & $127.53\!\pm\!1.44$ & $70.28\!\pm\!1.12$ & $314.482\!\pm\!1.147$ \\
\midrule
\multirow{3}{*}{dSprites}
& \multirow{3}{*}{Riesz--AdamW}
& Observed marginal $P_Y$ & $8.41\!\pm\!0.08$ & $39.06\!\pm\!0.47$ & $36.488\!\pm\!1.648$ \\
& & Moment-matched & $8.40\!\pm\!0.06$ & $39.04\!\pm\!0.45$ & $36.273\!\pm\!1.617$ \\
& & Standard normal & $8.40\!\pm\!0.04$ & $39.04\!\pm\!0.40$ & $36.310\!\pm\!1.661$ \\
\bottomrule
\end{tabular}
\end{adjustbox}
\end{table}

Observed-outcome initialization gives the strongest results on ColorMNIST and CelebA, indicating that starting on the learned image manifold can materially facilitate particle optimization. On CelebA, the full moment-matched Gaussian attains almost the same latent \(\mathrm{SW}_2\) as the observed initialization, but substantially worse outcome-space \(\mathrm{SW}_2\) and FID. Thus, matching low-order latent moments does not by itself ensure realistic decoded samples.

In the two-dimensional benchmark, the preferred initialization depends on the particle update, while dSprites is nearly insensitive to the three initializations considered. Overall, observational initialization is most beneficial for the more complex image distributions, rather than being uniformly optimal.

\paragraph{Full ridge versus projected Nystr\"om.}
We compare the full-ridge and projected-Nystr\"om constructions of the conditional mean embedding on ColorMNIST. The paired runs use the same data seeds, intervention arms, and particle-flow settings. The projected estimator uses \(m_L= 1024\) uniformly sampled landmarks. Table~\ref{tab:ablation-solver-runtime} reports the resulting outcome-space discrepancy and the wall-clock time required to solve the conditional-mean-embedding regression.

\begin{table}[H]
\centering
\caption{ColorMNIST full-ridge and projected-Nystr\"om comparison ($M=2{,}000$). CME solve times are wall-clock seconds for one intervention arm on an L40S GPU, measured with both solvers run back to back in a single process on identical inputs, after a warm-up pass and with the solver order rotated across repeats; mean $\pm$ s.d.\ over three data seeds $\times$ three repeats. The timing covers the ridge solve alone.}
\label{tab:ablation-solver-runtime}
\setlength{\tabcolsep}{4pt}
\renewcommand{\arraystretch}{1.1}
\begin{adjustbox}{max width=\textwidth}
\begin{tabular}{@{}rcccc@{}}
\toprule
$n$
& \multicolumn{2}{c}{$10^3\!\times\!\mathrm{SW}_2$}
& \multicolumn{2}{c}{CME solve time (s)} \\
\cmidrule(lr){2-3}\cmidrule(lr){4-5}
& Full ridge & Projected Nystr\"om
& Full ridge & Projected Nystr\"om \\
\midrule
$5{,}000$
& $10.69\!\pm\!1.28$ & $10.57\!\pm\!1.03$
& $0.20\!\pm\!0.00$ & $0.03\!\pm\!0.00$ \\
$10{,}000$
& $10.19\!\pm\!0.37$ & $10.29\!\pm\!0.61$
& $1.26\!\pm\!0.00$ & $0.03\!\pm\!0.00$ \\
$20{,}000$
& $9.52\!\pm\!0.37$ & $9.35\!\pm\!0.41$
& $9.09\!\pm\!0.00$ & $0.04\!\pm\!0.00$ \\
\bottomrule
\end{tabular}
\end{adjustbox}
\end{table}

The full-ridge solve scales cubically in $n$ ($0.20 \to 1.26 \to 9.09$ seconds), while the projected Nyström solve is flat at $0.03$–$0.04$ seconds, a $6.7\times$ to $227\times$ reduction over this range, at no cost in outcome-space discrepancy.

\paragraph{Effect of confounding strength.}
For ColorMNIST, the parameter \(w\) controls the strength of the association between treatment and the context determining foreground colour. Increasing \(w\) makes the treatment assignment more deterministic and weakens effective overlap. We vary \(w\in\{1,\ldots,10\}\) while holding \(n=20{,}000\), \(M=2{,}000\), and all remaining settings fixed. 

\begin{table}[H]
\centering
\caption{
ColorMNIST ablation over confounding strength for DR--IMMD.
``Initial'' denotes the observational-arm particles before optimization,
and ``flow'' denotes the particles returned by IMMD-Flow.
Lower is better. Entries are means \(\pm\) sample standard deviations
across three data seeds.
}
\label{tab:ablation-confounding}
\small
\setlength{\tabcolsep}{10pt}
\renewcommand{\arraystretch}{1.10}
\begin{tabular}{@{}rcc@{}}
\toprule
&
\multicolumn{2}{c}{\(10^3\!\times\!\mathrm{SW}_2\)}
\\
\cmidrule(lr){2-3}
\(w\)
& initial
& flow
\\
\midrule
\(1\)
& \(10.89\!\pm\!0.53\)
& \(7.60\!\pm\!0.26\)
\\
\(2\)
& \(17.69\!\pm\!0.30\)
& \(7.64\!\pm\!0.27\)
\\
\(3\)
& \(22.61\!\pm\!0.96\)
& \(8.20\!\pm\!0.77\)
\\
\(4\)
& \(27.43\!\pm\!0.82\)
& \(8.09\!\pm\!0.42\)
\\
\(5\)
& \(31.62\!\pm\!0.42\)
& \(8.30\!\pm\!0.51\)
\\
\(6\)
& \(34.28\!\pm\!0.58\)
& \(8.44\!\pm\!0.34\)
\\
\(7\)
& \(37.83\!\pm\!0.06\)
& \(8.43\!\pm\!0.30\)
\\
\(8\)
& \(40.33\!\pm\!0.95\)
& \(8.68\!\pm\!0.62\)
\\
\(9\)
& \(42.24\!\pm\!0.58\)
& \(8.84\!\pm\!0.38\)
\\
\(10\)
& \(43.32\!\pm\!0.06\)
& \(9.46\!\pm\!0.53\)
\\
\bottomrule
\end{tabular}
\end{table}

As \(w\) increases, the observational-arm initialization moves progressively farther from the interventional target, with its \(\mathrm{SW}_2\) increasing from \(10.89\) to \(43.32\). In contrast, the final discrepancy remains comparatively stable, increasing only from \(7.60\) to \(9.46\). Thus, DR--IMMD corrects most of the confounding-induced distribution shift throughout the sweep, with only gradual degradation at the strongest assignment levels.

\subsection{Compute resources}
\label{sec:ComputeResources}

All experiments were run on an internal SLURM-managed cluster of Linux x86\_64 nodes, using Python 3.12.12 with PyTorch 2.10.0 built against CUDA 12.6 (NVIDIA driver 580.95.05). Each run was submitted as a single-node, single-task job requesting one GPU, two to eight CPU cores, between 12\,GB and 64\,GB of system memory, and a wall-clock limit of four to twelve hours; parameter sweeps were executed as SLURM job arrays with one configuration per array task. The ColorMNIST and dSprites experiments used a single NVIDIA L40S GPU with approximately 46\,GB of memory, and the CelebA experiments a single NVIDIA A100-SXM4-40GB with 40\,GB; the two-dimensional synthetic design contains no GPU code and was run on CPU. Every wall-clock comparison we report was pinned to one GPU model, so no timing is compared across hardware.

\subsection{Existing assets, licences, and sensitive content}
\label{sec:assets}
All datasets used in this paper are existing research benchmarks and are used solely for research evaluation. ColorMNIST is derived from MNIST, which is made available under the Creative Commons Attribution--ShareAlike 3.0 Unported (CC BY-SA 3.0) licence. dSprites is obtained from the official Google DeepMind repository\footnote{\url{https://github.com/google-deepmind/dsprites-dataset}}, which is distributed under the Apache License 2.0. CelebA is used under its dataset agreement for non-commercial research purposes only. The supplementary code does not redistribute CelebA; users must obtain the dataset from its official source and comply with its terms of use. For CelebA, we load the pretrained \texttt{stabilityai/sd-vae-ft-mse} checkpoint through the Hugging Face Diffusers implementation of \texttt{AutoencoderKL}.\footnote{ Diffusers: \url{https://github.com/huggingface/diffusers}; model checkpoint: \url{https://huggingface.co/stabilityai/sd-vae-ft-mse}.} The Diffusers library is distributed under the Apache License 2.0, whereas the model repository lists an MIT licence. We use the model only as a fixed encoder and decoder and do not fine-tune or redistribute its weights. We do not redistribute external datasets or pretrained model weights, and provide only code and instructions for obtaining these assets from their original sources.

CelebA contains identifiable photographs of public figures together with binary appearance annotations, including the dataset's sex and hair-colour labels. We use CelebA only as a controlled benchmark for population-level distributional rebalancing: we do not perform identity recognition or make person-specific causal claims. The results should not be interpreted as establishing that the generated distributions are demographically representative, fair, or appropriate for real-world use. No new human annotations or participant data were collected.

\end{document}